\documentclass{article}
\usepackage{amsfonts}
\usepackage{dsfont}
\usepackage{geometry}
\usepackage{soul}
\usepackage{booktabs}
\usepackage[round]{natbib}
\usepackage[utf8]{inputenc}
\usepackage{enumitem}
\usepackage{amsmath}
\usepackage{amsthm}
\usepackage{amssymb}
\usepackage{bbm}
\usepackage{comment}

\newtheorem{theorem}{Theorem}
\newtheorem{corollary}[theorem]{Corollary}
\newtheorem{proposition}[theorem]{Proposition}
\newtheorem{lemma}[theorem]{Lemma}

\theoremstyle{remark}
\newtheorem{remark}{Remark}[theorem]

\usepackage{graphicx}
\usepackage{xcolor}
\usepackage{hyperref}
\hypersetup{
	colorlinks=true,
	linktoc=all,
	linkcolor=blue, 
	urlcolor=blue,   
	citecolor=blue,  
	filecolor=blue,  
	linktocpage=true 
}
\usepackage{cleveref}

\begin{document}
	
	\title{Computational-Statistical Trade-off in Kernel Two-Sample Testing with Random Fourier Features}
	
	\author{Ikjun Choi \thanks{Department of Statistics and Data Sciences, The University of Texas at Austin.} 
		\and 
		Ilmun Kim \thanks{Department of Mathematical Sciences, KAIST.}
	}
	
	\maketitle	
	
	\begin{abstract}	
		Recent years have seen a surge in methods for two-sample testing, among which the Maximum Mean Discrepancy (MMD) test has emerged as an effective tool for handling complex and high-dimensional data. Despite its success and widespread adoption, the primary limitation of the MMD test has been its quadratic-time complexity, which poses challenges for large-scale analysis. While various approaches have been proposed to expedite the procedure, it has been unclear whether it is possible to attain the same power guarantee as the MMD test at sub-quadratic time cost. To fill this gap, we revisit the approximated MMD test using random Fourier features, and investigate its computational-statistical trade-off. We start by revealing that the approximated MMD test is pointwise consistent in power only when the number of random features approaches infinity. We then consider the uniform power of the test and study the time-power trade-off under a minimax testing framework. Our result shows that, by carefully choosing the number of random features, it is possible to attain the same minimax separation rates as the MMD test within sub-quadratic time. We demonstrate this point under different distributional assumptions such as densities in a Sobolev ball. Our theoretical findings are corroborated by simulation studies.
	\end{abstract}
	
	\medskip
	
	
	{
		\setcounter{tocdepth}{2}
		\tableofcontents
	}
	
	\section{Introduction}
	
	The problem of two-sample testing stands as a fundamental topic in statistics, concerned with comparing two distributions to determine their equivalence. Classical techniques, such as the $t$-test and Wilcoxon rank-sum test, have been widely used to tackle this problem, and their theoretical and empirical properties have been well-investigated. However, these classical approaches often require parametric or strong moment assumptions to fully ensure their soundness, and their power is limited to specific directions of alternative hypotheses, such as location shifts. While these classical approaches are effective in well-structured and simple scenarios, their limitations in handling the increasing complexity of modern statistical problems have consistently prompted the need for new developments~\citep[][for a recent review]{stolte2023review}. Among various advancements made to address this issue, the kernel two-sample test based on the maximum mean discrepancy \citep[MMD,][]{gretton2012kernel} has garnered significant attention over the years, due to its nonparametric nature and flexibility. It can be applied in diverse scenarios without requiring distributional assumptions and offers robust theoretical underpinnings. With its empirical success and popularity, various research endeavors have been dedicated to enhancing their performance and deepening our understanding of their theoretical properties. 
	
	Broadly, there are two main branches of research regarding the kernel test: (i) kernel selection and (ii) computational time-power trade-off. Regarding kernel selection, significant advancements have been made in the last decade, aiming to identify the kernel that best captures the difference between two distributions. A common approach involves sample splitting where one-half of the data is used for kernel selection and the other half of the data is used for the actual test~\citep[e.g.,][]{gretton2012b,sutherland2017generative,liu2020deep}. However, an inefficient use of the data from sample splitting often results in a loss of power, which has been the main criticism. Another approach for kernel selection involves aggregating multiple kernels, which avoids sample splitting but requires a careful selection of kernels in advance~\citep[e.g.,][]{schrab2022efficient,schrab2023mmdaggregated,biggs2023fuse,Chatterjee2023}.  
	
	Regarding the time-power trade-off, much effort has concentrated on constructing a time-efficient test statistic with competitive power. The standard estimator of MMD via U-statistics or V-statistics demands quadratic-time complexity, which hinders the use of kernel tests for large-scale analyses. To mitigate this computational challenge, various methods have been proposed by using linear-time statistics~\citep{gretton2012kernel,gretton2012b}, block-based statistics~\citep{zaremba2013b} and more generally incomplete U-statistics~\citep{Yamada2019,schrab2022efficient}. However, these methods typically sacrifice statistical power for computational efficiency.  Another approach that aims to balance this time-power trade-off is based on random Fourier features~\citep[RFF,][]{rahimi2007random}. The idea is to approximate a kernel function using a finite dimensional random feature mapping, which can be computed efficiently. The use of RFF in a kernel test was initially considered by \citet{zhao2015fastmmd} and explored further by follow-up studies \citep[e.g.,][]{cevid2022distributional,mukherjee2025minimax}. It is intuitively clear that the performance of an RFF-MMD test crucially depends on the number of random features. While there is a line of work studying theoretical aspects of RFFs~\citep[][for a survey]{liu2022random}, their focus is mainly on the approximation quality of RFFs, and the optimal choice of the number of random features that balances between computational costs and statistical power remains largely unexplored.
	
	Motivated by this gap, we consider kernel two-sample tests using random Fourier features and aim to establish theoretical foundations for their power properties. 
	Our tests are based on a permutation procedure, which is practically relevant but introduces additional technical challenges. As mentioned earlier, both the quality and the computational complexity of the RFF-MMD test heavily depend on the number of random features. Our primary focus therefore is to determine the number of random features that strikes an optimal balance. It is worth highlighting that the challenge in our analysis lies in managing the interplay of three distinct randomness sources: the data itself, the random Fourier features, and the permutations employed in our approach. All of these random sources are intertwined within the testing process, which makes our analysis non-trivial and unique.  To effectively manage this complexity, we systematically decompose and analyze each layer of randomness in the test procedure, transitioning them into forms that are more amenable for analysis. This approach allows us to build on existing results from the literature that specifically address each of the three aspects of randomness.

	In the next subsection, we present a brief review of prior work that is most relevant to our paper.

	\subsection{Related work} \label{Section: Related work}
	In recent years, there has been a growing body of literature aimed at investigating the power of MMD-based tests and enhancing their performance. For example, the work of \citet{bala2021optimality,li2019optimality} demonstrated that MMD tests equipped with a fine-tuned kernel can achieve minimax optimality with respect to the $L_2$ separation in an asymptotic sense. To establish a similar but non-asymptotic guarantee, \citet{schrab2023mmdaggregated} introduced a MMD aggregated test calibrated by using either permutations or a wild bootstrap. It is also worth noting that the minimax optimality of MMD two-sample tests has been established for separations other than the $L_2$ distance, such as MMD 
	distance \citep{kim2023diff}, and Hellinger distance \citep{hagrass2022spectral}. In addition to these works, several other MMD-based minimax tests have been proposed using techniques such as aggregation \citep{Fromont2013,biggs2023fuse,Chatterjee2023} and studentization \citep{Shekhar2023permu,kim2024dim}. 
	Despite significant recent advancements made in this field, the quadratic time complexity of these methods remains a barrier in large-scale applications, which highlights the need for more efficient yet powerful testing approaches.

	To address the computational concern of quadratic-time MMD tests, several time-efficient approaches have emerged, which leverage subsampled estimation techniques, such as linear-time statistics~\citep{gretton2012kernel,gretton2012b}, block-based statistics~\citep{zaremba2013b} and incomplete U-statistics~\citep{Yamada2019,schrab2022efficient}. However, in terms of power, these methods are either sub-optimal or ultimately require quadratic time complexity to achieve optimality~\citep[][Proposition 2]{Carles2023Compress}. Other advancements in accelerating two-sample tests have involved techniques, such as Nystr\"{o}m approximations \citep{chatalic2022nystrom}, analytic mean embeddings and smoothed characteristic functions \citep{chwialkowski2015fast, jitkrittum2016inter}, deep linear kernels \citep{kirchler2020}, as well as random Fourier features~\citep{zhao2015fastmmd}. These tests can also run in sub-quadratic time, while their theoretical guarantees on power remain largely unknown. We also mention the recent method using kernel thinning~\citep{dwivedi2021kernel, Carles2023Compress}, which achieves the same MMD separation rate as the quadratic-time test but with sub-quadratic running time. However, this guarantee is valid under specific distributional assumptions that differ from those we consider. Moreover, their result focuses solely on alternatives that deviate from the null in terms of the MMD metric.

	With this context in mind, we revisit the RFF-MMD test~\citep{zhao2015fastmmd} and delve into its time-power trade-off concerning the number of random features. Despite an extensive body of literature on random features for kernel approximation, prior work has mainly focused on the estimation quality of kernel approximation~\citep{rahimi2007random, zhao2015fastmmd, bharath2015optimal, sutherland2015error, yao2023error}, and a theoretical guarantee on the power of the RFF-MMD test has not been explored. In this work, we seek to bridge this gap by thoroughly analyzing the trade-off between computation time and statistical power in the context of the RFF-MMD test. Following our initial arXiv preprint, \citet{mukherjee2025minimax} extended this perspective and studied an analogous computational-statistical trade-off for random-feature approximations of the cubic-time spectral-regularized MMD test \citep{hagrass2022spectral}, which is designed for more general (possibly non-Euclidean) domain settings.

	\subsection{Our contributions}
	Having reviewed the prior work, we now summarize the key contributions of this paper. 
	\begin{itemize}
		\item \textbf{Inconsistency result for RFF-MMD~(\Cref{Section: Lack of Consistency}).} We first investigate the setting where the number of random Fourier features is fixed, and demonstrate that the RFF-MMD test fails to achieve pointwise consistency~(\Cref{negthm} and Corollary~\ref{negcor}). Concretely, we prove that there exist infinitely many pairs of distinct distributions for which the power of the RFF-MMD test using a fixed number of random Fourier features is almost equal to the size even asymptotically. 
		\item \textbf{Sufficient conditions for consistency~(\Cref{Section: Optimal rate of the number of random features}).} Our previous negative result clearly indicates that increasing the number of random Fourier features is necessary to achieve pointwise consistency. In \Cref{conthm}, we show that it is indeed sufficient to increase the number of Fourier features to infinity to achieve pointwise consistency, even at an arbitrarily slow rate.
		\item \textbf{Time-power trade-off~(\Cref{Section: Optimal rate of the number of random features}).} As mentioned before, there exists a clear trade-off between computational efficiency and statistical power in terms of the number of random Fourier features. To balance this trade-off, we adopt the non-asymptotic minimax testing framework and analyze how changes in the number of random Fourier features impact both computational efficiency and separation rates in terms of the $L_2$ metric (\Cref{L2thm}) and the MMD metric~(\Cref{MMDthm}).
		\item \textbf{Achieving optimality in sub-quadratic time~(\Cref{Section: Optimal rate of the number of random features}).}  We firmly demonstrate in \Cref{L2thm} that it is possible to achieve the minimax separation rate against $L_2$ alternatives in sub-quadratic time when the underlying distributions are sufficiently smooth. Similarly, we establish in Proposition~\ref{MMDproposition} that a parametric separation rate against MMD alternatives can be achieved in linear time for certain classes of distributions including Gaussian distributions.
	\end{itemize}
	Our theoretical results are validated through simulation studies under various scenarios and the code that reproduces our numerical results can be found at \url{https://github.com/ikjunchoi/rff-mmd}.

	\paragraph{Organization.} The remainder of this paper is organized as follows. We set up the problem and present relevant background information in \Cref{Section: Background}. \Cref{Section: Lack of Consistency} provides an inconsistency result of the RFF-MMD test and highlights the important role of the number of random features in the power performance. Moving forward to \Cref{Section: Optimal rate of the number of random features}, we investigate the time-power trade-off in terms of the number of random features, denoted as $R$, and discuss an optimal choice of $R$ under minimax frameworks. We present simulation results in \Cref{Section: Numerical studies} that confirm our theoretical findings. Finally, in \Cref{Section: Discussion}, we discuss the implications of our findings and suggest directions for future research. All technical proofs are collected in the appendix.

	\section{Background} \label{Section: Background}
	In this section, we set up the problem and lay out some background for this work. Specifically, \Cref{section.twosample} explains the two-sample problem that we tackle, and specifies the desired error guarantees. We then present a brief overview of the MMD in \Cref{MMDsection} and its estimators using random Fourier features in \Cref{Section: Random Fourier Features}. Lastly, in \Cref{Section: Permutation test}, we review the permutation method for evaluating the significance of a two-sample test statistic.

	\subsection{Two-sample problem}\label{section.twosample}
	Let $\mathcal{X}_{n_1}:=\{X_i\}^{n_1}_{i=1}$ be ${n_1}$ i.i.d.~random samples from the distribution $P_X$, and $\mathcal{Y}_{n_2}:=\{Y_j\}^{n_2}_{j=1}$ be ${n_2}$ i.i.d.~random samples from the distribution $P_Y$ where ${n_1},{n_2} \geq 2$. Based on these mutually independent samples, the problem of two-sample testing is concerned with determining whether $P_X$ and $P_Y$ agree or not. More formally, let $\mathcal P$ be a class of all possible pairs of distributions on some generic space $\mathbb{S}$, and consider two disjoint subsets in $\mathcal P$, namely $\mathcal{P}_0:=\left\{(P_X,P_Y)\in\mathcal P\,|\,P_X=P_Y\right\}$ and $ \mathcal{P}_1:=\left\{(P_X,P_Y)\in\mathcal P\,|\,P_X\neq P_Y\right\}$. Then, the \textit{null} hypothesis $H_0$ and the \textit{alternative} hypothesis $H_1$ of two-sample testing can be formulated as follows:
	$$
	H_0:(P_X,P_Y) \in \mathcal P_0\quad \text{vs.}\quad H_1:(P_X,P_Y) \in \mathcal P_1.
	$$
	In order to decide whether to reject $H_0$ or not, we devise a test function $\Delta_{{n_1},{n_2}} : (\mathbb S^{n_1}, \, \mathbb S^{{n_2}})\rightarrow \{0,1\}$, and reject the null hypothesis if and only if $\Delta_{{n_1},{n_2}}(\mathcal{X}_{n_1},\mathcal{Y}_{n_2})=1$. This decision-making process naturally leads to two types of errors, which we would like to minimize. The first error, called the \textit{type I error}, occurs by rejecting the null hypothesis despite being true. Conversely, the second error, called the \textit{type II error}, arises when the null hypothesis is accepted despite being false. One common approach to design an ideal test is to first bound the probability of the \textit{type I error} uniformly over $\mathcal{P}_0$ as
	$$
	\sup_{(P_X,P_Y)\in \mathcal P_0}\mathbb{P}_{X \times Y}(\Delta_{{n_1,n_2}}(\mathcal{X}_{n_1},\mathcal{Y}_{n_2})=1) \leq \alpha, \quad \text{for a given \textit{level} $\alpha \in (0,1),$}
	$$
	where $\mathbb{P}_{X \times Y}$ denotes the probability operator over $\mathcal{X}_{n_1} \overset{\mathrm{i.i.d.}}{\sim} P_X$ and $\mathcal{Y}_{n_2}  \overset{\mathrm{i.i.d.}}{\sim} P_Y$.
	We say that such a test is a \textit{level}-$\alpha$ test. Next, our focus shifts to controlling the \textit{type II error.}  Given a fixed pair $(P_X,P_Y)$ in $\mathcal{P}_1$ and a \textit{level}-$\alpha$ test $\Delta^\alpha_{{n_1,n_2}}$, suppose that the probability of the \textit{type II error} is upper bounded by some constant $\beta \in (0,1)$. Equivalently, the probability of correctly rejecting the null, referred to as the \textit{power}, is lower bounded by $1-\beta$.  Ideally, we expect that the power of the test $\Delta^\alpha_{{n_1,n_2}}$ against any fixed alternative $(P_X,P_Y) \in \mathcal P_1$ converges to one as we increase the data size ${n_1}$ and ${n_2}$. More formally, we desire a test $\Delta^\alpha_{{n_1,n_2}}$ to be \textit{pointwise consistent}, satisfying
	\begin{align} \label{Eq: pointwise consistency}
		\lim_{{n_1,n_2}\rightarrow\infty}\mathbb{P}_{X\times Y}\bigl(\Delta^\alpha_{{n_1,n_2}}(\mathcal{X}_{n_1},\mathcal{Y}_{n_2})=1\bigr)=1, \quad \text{for any fixed $(P_X,P_Y) \in \mathcal P_1$.}
	\end{align}
	A stronger notion of the power is \textit{uniform consistency}, guaranteeing that the power converges to one uniformly over a class of alternative distributions. See \Cref{Section: Optimal rate of the number of random features} for a discussion. For simplicity, in the rest of this paper, we consider $\mathbb{S}$ to be the $d$-dimensional Euclidean space denoted as $\mathbb{R}^d$.
	
	\subsection{Maximum Mean Discrepancy}\label{MMDsection}
	As an example of integral probability metrics, the MMD measures the discrepancy between two distributions in a nonparametric manner. Specifically, given a reproducing kernel Hilbert space (RKHS) $\mathcal H_k$ equipped with a positive definite kernel $k$, the MMD between $P_X$ and $P_Y$ is defined as
	\begin{align*}
		\mathrm{MMD}(P_X,P_Y;\mathcal H_k):=\sup_{f \in \mathcal H_k : \|f\|_{\mathcal H_k}\leq 1} \bigl|\mathbb E_{X}\left[f(X)\right]-\mathbb E_{Y}\left[f(Y)\right] \bigr|.
	\end{align*}
	It can also be represented as the RKHS distance between two mean embeddings of $P_X$ and $P_Y$, i.e.,
	$
	\mathrm{MMD}(P_X,P_Y;\mathcal H_k)=\|\mu_X-\mu_Y\|_{\mathcal H_k}
	$
	where $\mu_X(\cdot):=\mathbb E_{X}[k(X,\cdot)]$ and $\mu_Y(\cdot):=\mathbb E_{Y}[k(Y,\cdot)]$. For a characteristic kernel $k$, the mean embedding of the kernel is injective \citep{sriperumbudur2010hilbert}, which means that $\mathrm{MMD}(P_X,P_Y;\mathcal{H}_k)=0$ if and only if $P_X=P_Y$. Among several ways to estimate the MMD, one straightforward way is to substitute the population mean embeddings $\mu_X$ and $\mu_Y$ with the empirical counterparts $\hat{\mu}_{X}(\cdot)=\frac{1}{n_1}\sum_{i=1}^{n_1}k(X_i,\cdot)$ and $\hat{\mu}_{Y}(\cdot)=\frac1{{n_2}}\sum_{i=1}^{n_2}k(Y_i,\cdot)$. This plug-in approach results in a biased quadratic-time estimator of the squared MMD, also referred to as the V-statistic, given as 
	\begin{equation}   \label{mmdb2}
		\begin{aligned}
			\widehat {\mathrm{MMD}}_b^2( \mathcal{X}_{n_1}, \mathcal{Y}_{n_2};\mathcal H_k)&=\left\|\frac{1}{n_1}\sum_{i=1}^{n_1}k(X_i,\cdot)-\frac1{{n_2}}\sum_{i=1}^{n_2}k(Y_i,\cdot) \right\|_{\mathcal H_k}^2  \\ 
			&=  \frac{1}{{n_1^2}} \displaystyle\sum_{i=1}^{n_1} \sum_{j=1}^{n_1} k\left(X_i, X_j\right)+\frac{1}{{n_2^2}} \sum_{i=1}^{n_2} \sum_{j=1}^{n_2} k\left(Y_i, Y_j\right)  -\frac{2}{{n_1} {n_2}} \sum_{i=1}^{n_1} \sum_{j=1}^{n_2} k\left(X_i, Y_j\right).
		\end{aligned}
	\end{equation}
	Denoting ${N}:={n_1}+{n_2}$, this plug-in estimator requires a quadratic-time cost of $O({N}^2d)$ in terms of the sample size $N$ as it involves evaluating pairwise kernel similarities between samples. Another common approach to estimate $\mathrm{MMD}^2(P_X,P_Y;\mathcal H_k)$ is using the U-statistic \citep[e.g.,][Lemma 6]{gretton2012kernel}, which is given as
	\begin{equation*}
		\begin{aligned}
			\widehat {\mathrm{MMD}}_u^2( \mathcal{X}_{n_1}, \mathcal{Y}_{n_2};\mathcal H_k) \,=\, & \frac{1}{{n_1}({n_1}-1)} \sum_{1\leq i\neq j \leq {n_1}} k\left(X_i, X_j\right)  + \frac{1}{{n_2}({n_2}-1)} \sum_{1\leq i\neq j \leq {n_2}}  k\left(Y_i, Y_j\right) \\
			- & \frac{2}{{n_1} {n_2}} \sum_{i=1}^{n_1} \sum_{j=1}^{n_2} k\left(X_i, Y_j\right).
		\end{aligned}
	\end{equation*}
	This estimator is an unbiased estimator of the squared MMD and also requires quadratic-time computational costs. 
	
	\subsection{Random Fourier features} \label{Section: Random Fourier Features} 
	Numerous approaches have been introduced to mitigate the computational cost of quadratic-time statistics often at the cost of sacrificing power performance. As reviewed in \Cref{Section: Related work}, some notable approaches include incomplete U-statistics~\citep{gretton2012kernel,zaremba2013b,schrab2022efficient}, Nystr\"{o}m approximations~\citep{chatalic2022nystrom}, kernel thinning~\citep{dwivedi2021kernel,Carles2023Compress} 
	and random Fourier features~\citep{rahimi2007random,zhao2015fastmmd,mukherjee2025minimax}. This work focuses on the method utilizing random Fourier features and investigates the effect of the number of random features on the power of a test. At the heart of this method is Bochner's theorem (Lemma \ref{bochner}), which offers a means to approximate the kernel using a low-dimensional feature mapping $\psi_{\omega}$, satisfying $k(x,y)\approx \langle\psi_{\omega}(x),\psi_{\omega}(y)\rangle$. If a bounded continuous positive definite kernel $k$ is translation invariant on $\mathbb R^d$, that is, $k(x,y)=\kappa(x-y)$, Bochner's theorem guarantees the existence of a nonnegative Borel measure $\Lambda$. It can be shown that $\Lambda$ is the inverse Fourier transform of $\kappa$ and satisfies
	$$
	k(x,y)=\int_{\mathbb R^d} e^{\sqrt{-1}\omega^\top(x-y)}d\Lambda(\omega)\stackrel{(\dagger)}{=}\int_{\mathbb R^d} \cos\bigl({\omega^\top(x-y)}\bigr)d\Lambda(\omega),
	$$
	where the equality $(\dagger)$ is obtained by the fact that $\kappa$ is both real and symmetric. Without loss of generality, we assume that $\Lambda$ is a probability measure, allowing the last integral to be expressed as $\mathbb E_{\omega\sim\Lambda}[\langle\psi_{\omega}(x),\psi_{\omega}(y)\rangle]$ with $\psi_{\omega}(x):=[\cos (\omega^\top x),\sin(\omega^\top x)]^\top$. If not, we instead work with the scaled versions of $\Lambda$ and $\psi_{\omega}$, given as $\Lambda' := \kappa(0)^{-1} \Lambda$ and $\psi_{\omega}'(\cdot):=[\sqrt{\kappa(0)}\cos (\omega^\top \cdot),\sqrt{\kappa(0)}\sin(\omega^\top \cdot)]^\top$. In this case, $k(x,y)$ can be represented as $\mathbb E_{\omega\sim\Lambda'}[\langle\psi_{\omega}'(x),\psi_{\omega}'(y)\rangle]$. 
	
	Now, by drawing a sequence of i.i.d.~$R$ random frequencies $\boldsymbol{\omega}_R := \{\omega_r\}^R_{r=1}$ from $\Lambda$, we construct an unbiased estimator of $k(x,y)$ defined as an inner product of random feature maps:
	\begin{equation}\label{hatk}
		\hat k(x,y):= \frac1R \sum\limits_{r=1}^R\langle \psi_{\omega_r}(x),\psi_{\omega_r}(y)\rangle
		=\langle \boldsymbol{\psi}_{\boldsymbol{\omega}_R}(x),\boldsymbol{\psi}_{\boldsymbol{\omega}_R}(y)\rangle,
	\end{equation}
	where $\psi_{\omega_r}(x)=[\cos (\omega_r^\top x),\sin(\omega_r^\top x)]^\top$ and $\boldsymbol{\psi}_{\boldsymbol{\omega}_R}(x)=\frac{1}{\sqrt R}[\psi_{\omega_1}(x)^\top,~\ldots~,\psi_{\omega_R}(x)^\top]^\top \in \mathbb R^{2R}$. Let us define the vector in $\mathbb{R}^{2R}$ representing the difference in sample means of random feature maps as follows: 
	\begin{equation*}
		T(\mathcal{X}_{n_1}, \mathcal{Y}_{n_2};\boldsymbol{\omega}_R):=  \frac{1}{n_1}\sum_{i=1}^{n_1}\boldsymbol{\psi}_{\boldsymbol{\omega}_R}(X_i)-\frac1{{n_2}}\sum_{j=1}^{n_2}\boldsymbol{\psi}_{\boldsymbol{\omega}_R}(Y_j).
	\end{equation*}
	Also, denote the quadratic form of $T:=T(\mathcal{X}_{n_1}, \mathcal{Y}_{n_2};\boldsymbol{\omega}_R)$ as $V:=T^\top T$. When we replace the kernel $k$ in Equation \eqref{mmdb2} with the estimated $\hat k$, we obtain a RFF-MMD estimator of $\mathrm{MMD}^2$ that can run with a time complexity of $O({N}Rd)$:
	\begin{equation}\label{rMMDb}
		\text{r}\widehat {\mathrm{MMD}}_b^2( \mathcal{X}_{n_1}, \mathcal{Y}_{n_2};\boldsymbol{\omega}_R):= V(\mathcal{X}_{n_1}, \mathcal{Y}_{n_2};\boldsymbol{\omega}_R)=\Bigg\|\frac{1}{n_1}\sum_{i=1}^{n_1}\boldsymbol{\psi}_{\boldsymbol{\omega}_R}(X_i)-\frac1{{n_2}}\sum_{j=1}^{n_2}\boldsymbol{\psi}_{\boldsymbol{\omega}_R}(Y_j) \Bigg\|_{\mathbb R^{2R}}^2.
	\end{equation}
	Notably, this estimator can be computed in linear time in terms of the pooled sample size ${N}$, and this computational benefit has motivated the prior work, such as \citet{zhao2015fastmmd}, \citet{sutherland2015error} and \citet{cevid2022distributional}, that consider RFF-MMD statistics. 
	
	One may also consider an unbiased RFF-MMD statistic, given as
	\begin{equation}
		\begin{aligned} \label{rMMDu}
			\text{r}\widehat {\mathrm{MMD}}_u^2( \mathcal{X}_{n_1}, \mathcal{Y}_{n_2};\boldsymbol{\omega}_R)&:= \frac{1}{{n_1}({n_1}-1)} \sum_{1\leq i\neq j \leq {n_1}} \langle \boldsymbol{\psi}_{\boldsymbol{\omega}_R}(X_i),\boldsymbol{\psi}_{\boldsymbol{\omega}_R}(X_j)\rangle\\
   &\quad +\frac{1}{{n_2}({n_2}-1)} \sum_{1\leq i\neq j \leq {n_2}}  \langle \boldsymbol{\psi}_{\boldsymbol{\omega}_R}(Y_i),\boldsymbol{\psi}_{\boldsymbol{\omega}_R}(Y_j)\rangle\\
   &\quad-\frac{2}{{n_1} {n_2}} \sum_{i=1}^{n_1} \sum_{j=1}^{n_2} \langle \boldsymbol{\psi}_{\boldsymbol{\omega}_R}(X_i),\boldsymbol{\psi}_{\boldsymbol{\omega}_R}(Y_j)\rangle,\\ 
		\end{aligned}
	\end{equation}
	which also involves $O({N}Rd)$ computational time~\citep[Appendix A.1]{zhao2015fastmmd}. In this work, we consider both $\text{r}\widehat {\mathrm{MMD}}_b^2$ and $\text{r}\widehat {\mathrm{MMD}}_u^2$ to demonstrate statistical and computational trade-offs in RFF-based two-sample testing.
	
	\subsection{Permutation tests} \label{Section: Permutation test}
	There have been several methods proposed for determining the threshold for MMD tests, which ensures (asymptotic or non-asymptotic) type I error control. These methods include those using limiting distributions or concentration inequalities, Gamma approximations, and bootstrap/permutation methods~\citep[e.g.,][]{gretton2012kernel,schrab2023mmdaggregated}. Among these, permutation tests stand out for their unique strength: they maintain level $\alpha$ for any finite sample size and often achieve optimal power~\citep[e.g.,][]{Kim2022}. This advantage has made permutation tests a popular choice in real-world applications despite extra computational costs. Given their practical relevance, this work focuses on permutation-based MMD tests and establishes their theoretical guarantees. 
	
	To explain the procedure, let us write the pooled sample as $\mathcal Z_{N}:=\left\{Z_1,\ldots,Z_{N}\right\}=\left\{\mathcal{X}_{n_1}, \mathcal{Y}_{n_2}\right\}$, and denote the collection of all possible permutations of $(1,2,\ldots,{N})$ as $\Pi_{N}$. Given a permutation $\pi:=(\pi(1),\ldots,\pi({N}))\in \Pi_{N},$ we denote the permuted pooled samples as $\mathcal Z^\pi_{N}:=\left\{Z_{\pi(1)},\ldots,Z_{\pi({N})}\right\}.$ Then, for a generic test statistic $T_{{n_1},{n_2}},$ the permutation distribution of $T_{{n_1},{n_2}}$ is defined as
	$$
	F^\pi_{T_{{n_1},{n_2}}}(t):=\frac{1}{{N}!}\sum_{\pi \in \Pi_{{N}}} \mathds 1 \{T_{{n_1},{n_2}}(\mathcal Z^\pi_{N})\leq t\}.
	$$
	The permutation test rejects the null hypothesis when $T_{{n_1},{n_2}}(\mathcal Z_{N})> q_{{n_1},{n_2},1-\alpha}$ where $q_{{n_1},{n_2},1-\alpha}$ denotes the $1-\alpha$ quantile of $F^\pi_{T_{{n_1},{n_2}}}$ given as $$
	q_{{n_1},{n_2},1-\alpha}:=\inf\big\{t:F_{T_{{n_1},{n_2}}}^{\pi}(t)\geq1-\alpha\big\}.
	$$ 
	It is well-known that the resulting permutation test maintains non-asymptotic type I error control under the exchangeability of random vectors~\citep[e.g.,][Theorem 1]{hemerik2018exact}. This exchangeability condition is satisfied under the null hypothesis of two-sample testing where $\mathcal{Z}_{N}$ are assumed to be i.i.d.~random vectors. 
	
	A more computationally efficient permutation test is defined through Monte Carlo simulations. Let $\pi_1,\ldots,\pi_B$ be permutation vectors randomly drawn from $\Pi_{N}$ with replacement. We let $T_{{n_1},{n_2}}^{(1)},\ldots,T_{{n_1},{n_2}}^{(B)}$ denote the test statistics computed based on $\mathcal{Z}_{N}^{\pi_1},\dots,\mathcal{Z}_{N}^{\pi_B}$. Let $\hat{q}_{{n_1},{n_2},1-\alpha}$ be the $1-\alpha$ quantile of the empirical distribution of $\{T_{{n_1},{n_2}}, T_{{n_1},{n_2}}^{(1)},\ldots,T_{{n_1},{n_2}}^{(B)} \}$, and reject the null when $T_{{n_1},{n_2}} > \hat{q}_{{n_1},{n_2},1-\alpha}$. The resulting Monte Carlo-based test is also valid in finite samples~\citep[][Theorem 2]{hemerik2018exact} and has almost equivalent power behavior as the full permutation test for sufficiently large $B$.

	\section{Lack of consistency}\label{Section: Lack of Consistency}
	In this section, we show that the RFF-MMD test, employing a finite number of random Fourier features, lacks pointwise consistency --- i.e., it fails to fulfill the guarantee in Equation~\eqref{Eq: pointwise consistency} --- even when the underlying kernel is characteristic. We establish this inconsistency result by focusing on a permutation test based on the test statistic in Equation~\eqref{rMMDb} or that in Equation~\eqref{rMMDu}, while our main idea is not limited to these specific tests. We start by explaining the intuition behind this negative result in \Cref{Section: intuition} and then present the main results in \Cref{Section: Main results}.

	\subsection{Preliminaries and intuition} \label{Section: intuition}
	An alternative formulation of $\text{r}\widehat {\mathrm{MMD}}_b^2$ in Equation~\eqref{rMMDb} is in terms of the characteristic functions of $P_X$ and $P_Y$. This reformulation provides a key insight into our negative result in \Cref{Section: Main results}. To fix ideas, the squared MMD with a translation-invariant kernel $k$ can be represented as $\mathrm{MMD}^2(P_X,P_Y;\mathcal H_k)=\int_{\mathbb{R}^d} |\phi_X(\omega)-\phi_Y(\omega)|^2d\Lambda(\omega)$ where $\phi_X$ and $\phi_Y$ are the characteristic functions of $P_X$ and $P_Y$, respectively~\citep[e.g.,][Corollary 4]{sriperumbudur2010hilbert}. Letting $\hat\phi_X(\omega):= \frac{1}{n_1} \sum_{i=1}^{n_1}e^{\sqrt{-1}\omega^\top X_i}$ and $\hat\phi_Y(\omega):= \frac{1}{n_2} \sum_{j=1}^{n_2}e^{\sqrt{-1}\omega^\top Y_j}$, we may represent the plug-in estimator in Equation~\eqref{mmdb2} as
	\begin{align*}
		\widehat {\mathrm{MMD}}_b^2( \mathcal{X}_{n_1}, \mathcal{Y}_{n_2};\mathcal H_k) = \int_{\mathbb{R}^d} |\hat\phi_X(\omega)-\hat\phi_Y(\omega)|^2d\Lambda(\omega).
	\end{align*}
	With this identity in place, the RFF-MMD statistic $\text{r}\widehat {\mathrm{MMD}}_b^2$ can be regarded as an approximation of the above plug-in estimator via Monte Carlo simulations with $R$ random frequencies $\{\omega_r\}_{r=1}^R \overset{\mathrm{i.i.d.}}{\sim} \Lambda$. Specifically, the RFF-MMD statistic can be written in terms of the empirical characteristic functions as:
	\begin{align*}
		\text{r}\widehat {\mathrm{MMD}}_b^2( \mathcal{X}_{n_1}, \mathcal{Y}_{n_2};\boldsymbol{\omega}_R) = \frac{1}{R} \sum^R_{r=1} |\hat\phi_X(\omega_r)-\hat\phi_Y(\omega_r)|^2.
	\end{align*}
	As is well-known, the characteristic function uniquely determines the distribution of a random vector. Therefore, when the support of $\Lambda$ is the entire Euclidean space, the population MMD becomes zero if and only if $P_X$ and $P_Y$ coincide. However, the empirical MMD evaluated on a finite number of random points is unable to capture an arbitrary difference between $P_X$ and $P_Y$, even asymptotically. At a high-level, this happens due to a combination of two factors. First of all, it is possible that two distinct characteristic functions can be equal in an interval~\citep[e.g.,][page 74]{romano1986}. Moreover, if random evaluation points $\{\omega_r\}_{r=1}^R$ fall within such interval with high probability, then the RFF-MMD statistic would behave similarly to the null case, resulting in a test that is inconsistent with a fixed number of random features. This observation was partly made in \citet[Proposition 1]{chwialkowski2015fast}, which we generalize to $\mathbb{R}^d$ as below. 
	\begin{lemma}[\citealt{chwialkowski2015fast}] \label{lemma1}
		Let $R \in \mathbb N$ be a fixed number and let $\boldsymbol{\omega}_R=\{\omega_r\}^R_{r=1}$ be a sequence of real-valued i.i.d.~random vectors from a probability distribution on $\mathbb R^d$ which is absolutely continuous with respect to the Lebesgue measure. For arbitrary $ \epsilon \in (0,1)$, there exists an uncountable set $\mathcal A_\epsilon$ of mutually distinct probability distributions on $\mathbb R^d$ such that for any distinct pair $P_X,P_Y \in \mathcal A_\epsilon$ and their corresponding random vectors $X$ and $Y$, it holds that $\mathbb{P}_{\boldsymbol{\omega}_R}(\frac{1}{R}\sum^R_{r=1}|\phi_X(\omega_r)-\phi_Y(\omega_r)|^2=0)\geq1-\epsilon$.
	\end{lemma}
	The above lemma implies that there exists a certain pair of $(P_X,P_Y)$ under the alternative such that the expectation of the RFF-MMD statistic~\eqref{rMMDb} is approximately zero with high probability. Given that the same test statistic has an expectation approximately equal to zero under the null, one may argue that the power of an RFF-MMD test would be strictly less than one against that specific alternative. However, this argument is insufficient to correctly claim the lack of consistency. An instructive example would be the case where a test statistic $W$ is either $0$ or $1/n$ with probability $1-\alpha$ and $\alpha$, respectively, under the null, whereas it takes the value $\alpha/n$ with probability one. In this case, it is clear to see that the expectation of $W$ remains the same under $H_0$ and $H_1$, converging to zero as $n \rightarrow \infty$. Nevertheless, if we reject the null when $W > 0$, the resulting test has size $\alpha$ and power one for any value of $n \geq 1$. This toy example suggests that Lemma~\ref{lemma1} is insufficient to formally prove the inconsistency result and we indeed need a distribution-level understanding of the RFF-MMD statistic. Moreover, when the critical value is determined via the permutation procedure (\Cref{Section: Permutation test}), we further need to take care of random sources arising from permutations, which adds an additional layer of technical challenges. With this context in place, we next develop inconsistency results by carefully studying the limiting distribution of the RFF-MMD statistic and its permuted counterpart.
	
	\subsection{Main results} \label{Section: Main results}
	Consider a permutation test based on the test statistic in~Equation~\eqref{rMMDb} defined as follows:
	\begin{equation}\label{pertest}
		\Delta_{{n_1,n_2},R}^{\alpha}(\mathcal{X}_{n_1}, \mathcal{Y}_{n_2};\boldsymbol{\omega}_R) :=\Delta_{{n_1,n_2},R}^{\alpha}:= \mathds{1}\big\{V(\mathcal{X}_{n_1}, \mathcal{Y}_{n_2};\boldsymbol{\omega}_R) > q_{{n_1,n_2},1-\alpha}\big\},
	\end{equation}
	where $q_{{n_1},{n_2},1-\alpha}:=\inf\{t:F_{V}^{\pi}(t)\geq1-\alpha\}$ and $F_{V}^{\pi}(t) = \frac{1}{{N}!}\sum_{\pi \in \Pi_{{N}}} \mathds 1 \{V(Z_{\pi(1)},\dots,Z_{\pi({N})};\boldsymbol{\omega}_R)\leq t\}$. Building on the intuition laid out in \Cref{Section: intuition}, we aim to prove that the asymptotic power of the test $\Delta_{{n_1},{n_2},R}^{\alpha}$ is strictly less than one with a fixed number of $R$ against certain fixed alternatives. To formally achieve this, let $\boldsymbol{\psi}_{\boldsymbol{x}}$ be defined similarly as $\boldsymbol{\psi}_{\boldsymbol{\omega}_R}$ by replacing $\boldsymbol{\omega}_R$ with $\boldsymbol{x} \in \mathbb{R}^{d \times R}$. Based on Euler's formula, the event $\frac1R \sum^R_{r=1}|\phi_X(\omega_r)-\phi_Y(\omega_r)|^2=0$ is equivalent to $\boldsymbol{\omega}_R \in \mathcal{E} := \mathcal{E}(X,Y)$ where
	\begin{equation}\label{meaneq}
		\mathcal{E}(X,Y) := \bigl\{ \boldsymbol{x} \in \mathbb{R}^{d \times R}:  \mathbb E_{X}[\boldsymbol{\psi}_{\boldsymbol{x}} (X)]=\mathbb E_{Y}[\boldsymbol{\psi}_{\boldsymbol{x}}(Y)]\bigr\}.
	\end{equation} 
	We refer to the event $\boldsymbol{\omega}_R \in \mathcal{E}$ as the \textit{first moment equivalence (1-ME) condition}, which holds with high probability, say $1-\epsilon$, for some fixed $(P_X,P_Y)$ according to Lemma~\ref{lemma1}. As mentioned earlier, the 1-ME condition alone is insufficient to formally prove the inconsistency result, which prompts an extension of the 1-ME condition to include higher-order moments. We consider a subset $\mathcal{E}_k\subseteq \mathcal{E}$ where $\boldsymbol{\omega}_R \in \mathcal{E}_k$
	implies equivalence of all mixed moments up to order $k$. Specifically, for
	a multi-index $\boldsymbol i=(i_1,\ldots,i_{2R})\in(\mathbb N\cup\{0\})^{2R}$, write
	$|\boldsymbol i|:=\sum_{\ell=1}^{2R}i_\ell.$ Define
		\begin{equation*}
	\mathcal{E}_k:=\bigl\{\boldsymbol{x} \in \mathbb{R}^{d \times R}\,:\, \mathbb E_{X}[\boldsymbol{\psi}_{\boldsymbol{x}} (X)^{\boldsymbol i}]=\mathbb E_{Y}[\boldsymbol{\psi}_{\boldsymbol{x}}(Y)^{\boldsymbol i}],~ \forall \boldsymbol i\in(\mathbb N\cup\{0\})^{2R}~\text{such that}~ |\boldsymbol i|\leq k
	\bigr\},
	\end{equation*} 
	where $\boldsymbol{\psi}_{\boldsymbol{x}}(X)^{\boldsymbol i}:=\prod_{\ell=1}^{2R}\boldsymbol{\psi}_{\boldsymbol{x}}(X)_\ell^{i_\ell}$ and $\boldsymbol{\psi}_{\boldsymbol{x}}(Y)^{\boldsymbol i}:=\prod_{\ell=1}^{2R}\boldsymbol{\psi}_{\boldsymbol{x}}(Y)_\ell^{i_\ell}$.	
	We refer to the event $\boldsymbol{\omega}_R\in \mathcal{E}_k$ as the \textit{first $k$ moments equivalence (k-ME) condition}. In the following proposition, we prove a generalized version of Lemma~\ref{lemma1} demonstrating that the $k$-ME condition holds with high probability for some fixed $(P_X,P_Y)$. The proof of this result can be found in \Cref{Section: proof of proposition.kmoment}.
	
	\begin{proposition}\label{proposition.kmoment}
		Let $k,R \in \mathbb N$ be fixed numbers and let $\boldsymbol{\omega}_R=\{\omega_r\}^R_{r=1}$ be a sequence of real-valued i.i.d.~random vectors from a probability distribution on $\mathbb R^d$ which is absolutely continuous with respect to the Lebesgue measure. For arbitrary $ \epsilon \in (0,1)$, there exists an uncountable set $\mathcal A_{k,\epsilon}$ of mutually distinct probability distributions on $\mathbb R^d$ such that for any distinct pair $P_X,P_Y \in \mathcal A_{k,\epsilon}$ and their corresponding random vectors $X$ and $Y$, it holds that $\mathbb{P}_{\boldsymbol{\omega}_R}(\boldsymbol{\omega}_R\in \mathcal{E}_k)\geq1-\epsilon$.
	\end{proposition}
	Suppose that $P_X,P_Y \in \mathcal A_{k,\epsilon}$, specified in Proposition~\ref{proposition.kmoment}. The power of the considered test against this specific alternative is then upper bounded as 
	\begin{equation}\label{upper bound for the power}
		\begin{aligned}
			\mathbb{P}(\Delta_{{n_1},{n_2},R}^{\alpha} = 1) & = \int_{\mathcal{E}_k}\mathbb{P}(\Delta_{{n_1},{n_2},R}^{\alpha} = 1 \,|\, \boldsymbol{\omega}_R = \boldsymbol{\omega}) f_{\boldsymbol{\omega}_R}(\boldsymbol{\omega}) d\boldsymbol{\omega} + \int_{\mathcal{E}^c_k}\mathbb{P}(\Delta_{{n_1},{n_2},R}^{\alpha} = 1 \,|\, \boldsymbol{\omega}_R = \boldsymbol{\omega}) f_{\boldsymbol{\omega}_R}(\boldsymbol{\omega}) d\boldsymbol{\omega} \\
			& \leq \int_{\mathcal{E}_k}\mathbb{P}(\Delta_{{n_1},{n_2},R}^{\alpha} = 1 \,|\, \boldsymbol{\omega}_R = \boldsymbol{\omega}) f_{\boldsymbol{\omega}_R}(\boldsymbol{\omega}) d\boldsymbol{\omega} + \epsilon.
		\end{aligned}
	\end{equation}
	Given this bound, our proof for inconsistency revolves around showing that $\int_{\mathcal{E}_k}\mathbb{P}(\Delta_{{n_1},{n_2},R}^{\alpha} = 1 \,|\, \boldsymbol{\omega}_R = \boldsymbol{\omega}) f_{\boldsymbol{\omega}_R}(\boldsymbol{\omega}) d\boldsymbol{\omega}$ is sufficiently small. This in turn requires understanding the limiting behavior of the test statistic $V(\mathcal{X}_{n_1}, \mathcal{Y}_{n_2};\boldsymbol{\omega}_R)$ and the permutation critical value $q_{{n_1},{n_2},1-\alpha}$ under the $k$-ME condition. On the one hand, the limiting distribution of the test statistic can be derived using the standard asymptotic tools such as the central limit theorem. On the other hand, we leverage asymptotic results for permutation distributions in \cite{chung2016multivariate} to show that the critical value $q_{{n_1},{n_2},1-\alpha}$ converges to the $1-\alpha$ quantile of a continuous distribution. We point out that both limiting and permutation distributions of $V(\mathcal{X}_{n_1}, \mathcal{Y}_{n_2};\boldsymbol{\omega}_R)$ are determined by the first two moments of $\boldsymbol{\psi}_{\boldsymbol{\omega}_R} (X)$ and $\boldsymbol{\psi}_{\boldsymbol{\omega}_R} (Y)$. Furthermore, both distributions become asymptotically identical when those moments are the same, implying the coincidence of both distributions under the 2-ME condition. Consequently, the power of the test $\Delta_{{n_1},{n_2},R}^{\alpha}$ under the 2-ME condition remains small even asymptotically, which together with inequality~\eqref{upper bound for the power}, leads to the inconsistency result. This negative result is formally stated in the following theorem and the proof can be found in \Cref{Section: proof of negthm}.
	\begin{theorem}\label{negthm}
		Let $k(x,y)=\kappa(x-y)$ be a bounded continuous positive definite kernel whose inverse Fourier transform is absolutely continuous with respect to the Lebesgue measure. Then, given any $\epsilon>0$, for the test $\Delta_{{n_1},{n_2},R}^{\alpha}$ defined in Equation \eqref{pertest} with a fixed number $R \geq 1$ and the limiting sample-ratio $p:=\lim_{{n_1},{n_2}\rightarrow\infty}\frac{{n_1}}{{n_1}+{n_2}} \in (0,1)$, there exist uncountably many pairs of distinct probability distributions $P_X$ and $P_Y$ on $\mathbb R^d$ that satisfy 
		$$
		\limsup_{{n_1},{n_2}\rightarrow\infty}\mathbb P_{X\times Y\times \omega}\big(\Delta_{{n_1},{n_2},R}^{\alpha}(\mathcal{X}_{n_1}, \mathcal{Y}_{n_2};\boldsymbol{\omega}_R)=1\big)\leq  \alpha + \epsilon.
		$$
	\end{theorem}
	
	The underlying idea of the proof for \Cref{negthm} can be applied to the unbiased RFF-MMD statistic in~Equation~\eqref{rMMDu} as well. In particular, consider a permutation test
	\begin{equation}\label{upertest}
		\Delta_{{n_1},{n_2},R}^{\alpha,u}(\mathcal{X}_{n_1}, \mathcal{Y}_{n_2};\boldsymbol{\omega}_R) := \mathds{1}\big\{{U(\mathcal{X}_{n_1}, \mathcal{Y}_{n_2};\boldsymbol{\omega}_R)} > q_{{n_1},{n_2},1-\alpha}^u\big\},
	\end{equation}
	where $U(\mathcal{X}_{n_1}, \mathcal{Y}_{n_2};\boldsymbol{\omega}_R):=\text{r}\widehat {\mathrm{MMD}}_u^2( \mathcal{X}_{n_1}, \mathcal{Y}_{n_2};\boldsymbol{\omega}_R)$ and $q_{{n_1},{n_2},1-\alpha}^u:=\inf\{t:F_{U}^{\pi}(t)\geq1-\alpha\}$ is the corresponding critical value. Building on the observation that the difference between $U(\mathcal{X}_{n_1}, \mathcal{Y}_{n_2};\boldsymbol{\omega}_R)$ and $V(\mathcal{X}_{n_1}, \mathcal{Y}_{n_2};\boldsymbol{\omega}_R)$ is asymptotically negligible, we derive a result analogous to Theorem \ref{negthm}, demonstrating that $\Delta_{{n_1},{n_2},R}^{\alpha,u}$ fails to be pointwise consistent. 
	
	\begin{corollary}\label{negcor}
		Consider the same setting in \Cref{negthm}. Given any $\epsilon >0$, for the test $\Delta_{{n_1},{n_2},R}^{\alpha,u}$ defined in Equation \eqref{upertest} with a fixed number $R$ and the limiting sample-ratio $p$, there exist uncountably many pairs of distinct probability distributions $P_X$ and $P_Y$ on $\mathbb R^d$ that satisfy 
		$$
		\limsup_{{n_1},{n_2}\rightarrow\infty}\mathbb P_{X\times Y\times \omega}\big(\Delta_{{n_1},{n_2},R}^{\alpha,u}(\mathcal{X}_{n_1}, \mathcal{Y}_{n_2};\boldsymbol{\omega}_R)=1\big)\leq  \alpha + \epsilon.
		$$
	\end{corollary}
	
	The proof of Corollary \ref{negcor} can be found in \Cref{Section: proof of negcor}. Our findings so far indicate that RFF-MMD tests with a fixed number of random features fail to be pointwise consistent. To address this issue, we naturally consider increasing $R$ with the sample size and show that the tests then become pointwise consistent. Moreover, in some cases, the RFF-MMD test can attain comparable power to the quadratic-time MMD test but in strictly less than quadratic time. These are the topics of the next section.
	
	\section{Optimal choice of the number of random features}  \label{Section: Optimal rate of the number of random features}
	We now turn to scenarios where the number of random Fourier features grows with the sample size, and examine computational and statistical trade-offs in selecting these random features. The first result of this section complements the previous inconsistency results, indicating that the RFF-MMD tests are pointwise consistent as long as the number of random Fourier features increases to infinity even at an arbitrarily slow rate. 
	
	\begin{theorem}\label{conthm}
		Consider an arbitrary sequence $\{R_{n}\}_{{n} \geq 1}$ that increases as $\lim_{{n}\rightarrow\infty}R_{n}=\infty$ and assume that the kernel $k(\cdot,\cdot)$ in \Cref{negthm} is characteristic.
		Then, against any fixed alternative $(P_X,P_Y)\in \mathcal P_1$, the permutation test $\Delta_{{n_1},{n_2},R}^{\alpha}$ defined in Equation \eqref{pertest} with $R=R_{n}$ and $n := \min \{n_1, n_2\}$ satisfies
		$$
		\lim_{{n_1},{n_2}\rightarrow\infty}\mathbb{P}_{X\times Y\times \omega}\big(\Delta_{{n_1},{n_2},R}^{\alpha}(\mathcal{X}_{n_1}, \mathcal{Y}_{n_2};\boldsymbol{\omega}_R)=1\big)=1.
		$$
		This result also holds for the permutation test $\Delta_{{n_1},{n_2},R}^{\alpha,u}$ defined in Equation \eqref{upertest}.
	\end{theorem}
	
	The proof of Theorem \ref{conthm} is given in \Cref{Section: proof of conthm}. It is worth noting that increasing the number of random features comes with an increase in computational cost. On the other hand, using a small number of random features may lead to suboptimal power performance compared to the quadratic-time MMD test. Therefore, achieving a balance between computational costs and statistical power is crucial from a practical standpoint. To determine the number of random features that balance this time-power trade-off, we adopt the minimax testing framework pioneered by \citet{Ingster1987,ingster1993} explained below.

	\paragraph{Minimax two-sample testing framework.} 
	While pointwise consistency in Equation~\eqref{Eq: pointwise consistency} is an important property, it only provides a guarantee against a fixed pair of alternative distributions, which may be regarded as a weak property. Given some constant $\beta \in (0,1)$, one might instead aim to build a test that also uniformly bounds the probability of \textit{type II error} in a non-asymptotic sense:
	$$
	\sup_{(P_X,P_Y)\in \mathcal P_1}\mathbb{P}_{X\times Y}\big(\Delta^\alpha_{{n_1,n_2}}(\mathcal{X}_{n_1},\mathcal{Y}_{n_2})=0\big)\leq\beta.
	$$
	In general, however, achieving this uniform guarantee is not feasible unless the two classes $\mathcal{P}_0$ and $\mathcal{P}_1$ are sufficiently distant. Therefore it is common to introduce a gap between $\mathcal P_0$ and $\mathcal P_1$, and analyze the minimum gap for which the testing error is uniformly controlled. 
	In detail, we define a class of alternative pairs $\mathcal P_1(\mathcal C, \delta,\epsilon):=\{(P_X,P_Y)\in\mathcal C\,|\,\delta(P_X,P_Y)\geq\epsilon\}$ where $\delta$ is a metric of interest, $\mathcal C \subseteq \mathcal P$ is a predefined class of distribution pairs (if not stated otherwise, $\mathcal C =\mathcal P$), and $\epsilon>0$ is a separation parameter. Then the \textit{uniform separation rate} that measures the performance of the test $\Delta$ is defined \citep[e.g.,][]{baraud2002,schrab2023mmdaggregated} as
	\begin{equation*}
		\rho\left(\Delta,\, \beta,\, \mathcal C,\, \delta\right):=\inf \Big\{\epsilon>0:  \sup_{(P_X,P_Y)\in \mathcal P_1(\mathcal C, \delta,\epsilon)}\mathbb{P}_{X\times Y}\big(\Delta(\mathcal{X}_{n_1},\mathcal{Y}_{n_2})=0\big)\leq\beta\Big\}.
	\end{equation*}
	Among all possible \textit{level}-$\alpha$ tests, it is reasonable to consider a test that achieves the smallest \textit{uniform separation} as an optimal test. More formally, we define the \textit{minimax separation} as
	\begin{equation*}
		\rho^\star\left(\alpha,\,\beta, \,\mathcal C, \,\delta\right):=\inf_{\Delta_\alpha} \rho\left(\Delta_\alpha,\, \beta,\,\mathcal C,\, \delta\right),
	\end{equation*}
	where the infimum is taken over all \textit{level}-$\alpha$ tests, and we refer to the test $\Delta$ satisfying $\rho\left(\Delta, \beta, \mathcal C, \delta\right) = \rho^\star\left(\alpha,\beta, \mathcal C, \delta\right)$ as a minimax optimal test. However, except for a few parametric problems, it is generally infeasible to devise an optimal test that precisely achieves the minimax separation. As a compromise, it is now conventional to seek a \textit{minimax rate optimal} test, which achieves the minimax separation, up to a constant. It has been shown that the quadratic-time MMD test is minimax rate optimal against the $L_2$ metric~\citep{schrab2023mmdaggregated,li2019optimality} and against the MMD metric~\citep{Kim2021,kim2023diff}. Our aim is to determine the minimum number of random features $R$ for which the RFF-MMD test attains the same optimality property as the quadratic-time MMD.

	\subsection{Uniform consistency in {$L_2$} metric} 
	We start by examining the uniform separation rate of the RFF-MMD test over the Sobolev ball with respect to the $L_2$ distance. Let us denote $\mathcal S_d^{s}(M_1)$ as the $s$th order Sobolev ball in $\mathbb R^d$ with radius $M_1>0,$ that is
	$$
	\mathcal S_d^{s}(M_1):=\Big\{f \in L^1\left(\mathbb{R}^d\right) \cap L^2\left(\mathbb{R}^d\right): \int_{\mathbb{R}^d}\|\omega\|_2^{2 s}|\widehat{f}(\omega)|^2 \mathrm{~d} \omega \leq(2 \pi)^d M_1^2\Big\},
	$$
	where $s>0$ is the smoothness parameter, and $\hat f(\omega)=\frac{1}{(2\pi)^{d/2}}\int_{\mathbb R^d}f(x)e^{-i\langle x,\omega\rangle}dx$ is the Fourier transform of $f$. Furthermore, each of $L^1(\mathbb{R}^d)$ and $L^2(\mathbb{R}^d)$ denotes a set of functions that are integrable in absolute value and square integrable, respectively. Let $\mathcal{P}_{\mathrm{conti}}$ be the collection of paired distributions on $\mathbb{R}^d$ where each pair of distributions $(P_X,P_Y) \in \mathcal{P}_{\mathrm{conti}}$ admits the probability density functions $(p_X,p_Y)$ with respect to the Lebesgue measure. Defining the class of distribution pairs with some constant $M_2>0$ given as
	\begin{align*}
		\widetilde{\mathcal{C}}_{L_2}:=\big\{(P_X,P_Y)\in \mathcal P_{\mathrm{conti}} \,\big|\,p_X-p_Y\in\mathcal S_d^{s}(M_1),~\max\{\|p_X\|_\infty,\|p_Y\|_\infty\}\leq M_2\big\},
	\end{align*}
	\citet{schrab2023mmdaggregated} demonstrated that the \textit{minimax rate} in terms of the $L_2$ distance is $\rho^\star(\alpha,\beta,\widetilde{\mathcal{C}}_{L_2},\delta_{L_2}) \asymp {n}^{-2s/(4s+d)}$, where $a_n \asymp b_n$ indicates $c \leq |a_n/b_n| \leq C$ for some positive constants $c,C$. They further showed that the MMD test using a translation-invariant kernel is \textit{minimax rate optimal} in a non-asymptotic sense. A similar but asymptotic result was obtained by \citet{li2019optimality}, focusing specifically on the Gaussian kernel.\footnote{Both \citet{schrab2023mmdaggregated} and \citet{li2019optimality} assume that $n_1 \asymp n_2$ under which the minimax rate against the $L_2$ alternative is given as $(n_1+n_2)^{-2s/(4s+d)}$. Without this balanced sample size assumption, however, the minimax rate is dominated by the minimum sample size i.e., $n^{-2s/(4s+d)}$.} It is intuitively clear that when the number of random features $R$ is sufficiently large, the RFF-MMD test will also attain the same minimax optimality as the law of large numbers guarantees that the approximated kernel $\hat{k}$ converges to the underlying kernel $k$ almost surely. Our next question is then to ask how rapidly $R$ should be increased to ensure the same minimax guarantee and whether it is possible to attain the same optimality in sub-quadratic time. We answer these questions in the affirmative. 
	
	Similarly to \cite{schrab2023mmdaggregated}, our analysis assumes that the kernel $k$ can be represented as a product of $d$ one-dimensional translation-invariant characteristic kernels with a given bandwidth $\lambda$. More specifically, we assume that the kernel $k$ can be decomposed as   
	$$
	k(x,y)=k_\lambda(x,y):=\prod^d_{i=1}\frac1{\lambda_i}\kappa_i\left(\frac{x_i-y_i}{\lambda_i}\right)
	$$
	for $\lambda=(\lambda_1,\ldots,\lambda_d)^\top\in(0,\infty)^d,$ where  $\kappa_i:\mathbb R \rightarrow \mathbb R$ are some non-negative functions in $L^1\left(\mathbb{R}\right) \cap L^2\left(\mathbb{R}\right)$ satisfying $\int_\mathbb R \kappa_i(x)dx=1$ for $i = 1,\ldots,d$. We note that $k_\lambda$ is indeed a characteristic kernel on $\mathbb R^d$, and we treat the bandwidth $\lambda$ as a tuning parameter that varies with the sample size. In order to highlight the dependence on $\lambda$, we let $\Delta_{{n_1},{n_2},R}^{\alpha,\lambda}$ (resp.~$\Delta_{{n_1},{n_2},R}^{\alpha,u,\lambda}$) denote the test $\Delta_{{n_1},{n_2},R}^{\alpha}$~(resp.~$\Delta_{{n_1},{n_2},R}^{\alpha,u}$) equipped with the kernel $k_\lambda$. Given a constant $M_3>0$, let us now consider a subset of $\widetilde{\mathcal{C}}_{L_2}$ where the support of individual distributions lies within the $d$-dimensional hypercube $[-M_3,M_3]^d$. In other words, we define 
	\begin{equation*}
		\mathcal{C}_{L_2} := \bigl\{(P_X,P_Y) \in \widetilde{\mathcal{C}}_{L_2} \,\big|\, \mathrm{support}(P_X),\, \mathrm{support}(P_Y) \subset [-M_3,M_3]^d \bigr\}.
	\end{equation*}
	Recalling $N=n_1 + n_2$ and $n=\min\{n_1,n_2\}$, the following theorem discusses the choice of $R$ and $\lambda$ that allows $\Delta_{{n_1},{n_2},R}^{\alpha,\lambda}$ and $\Delta_{{n_1},{n_2},R}^{\alpha,u,\lambda}$ to achieve the \textit{minimax separation rate} against the class of alternatives defined on $\mathcal{C}_{L_2}$. 
	\begin{theorem}\label{L2thm}
		 Consider the test $\Delta_{{n_1},{n_2},R}^{\alpha,\lambda}$ and $\Delta_{{n_1},{n_2},R}^{\alpha,u,\lambda}$ with $\lambda_i = {n}^{-2/(4s+d)}$, $i=1,\ldots,d$. Then, there exist some positive constants $C_R(M_3,\beta,d)$ and $C_{L_2}(M_1,M_2, M_3, \alpha,\beta,d,s)$ such that the uniform separation of $\Delta_{{n_1},{n_2},R}^{\alpha,\lambda}$ with the choice of $R \geq C_R(M_3,\beta,d){n}^{2d/(4s+d)}$ satisfies
		$$
		\rho\big(\Delta_{{n_1},{n_2},R}^{\alpha,\lambda},\,\beta,\,\mathcal C_{L_2},\, \delta_{L_2}\big)\leq C_{L_2}(M_1,M_2,M_3,\alpha,\beta,d,s){n}^{-2s/(4s+d)}.
		$$
		The same guarantee also holds for $\Delta_{{n_1},{n_2},R}^{\alpha,u,\lambda}$. Moreover, the computational cost of the corresponding test statistics $\text{r}\widehat {\mathrm{MMD}}_b^2$ and $\text{r}\widehat {\mathrm{MMD}}_u^2$ is $O(Nn^{\frac{2d}{4s+d}}d)$.
	\end{theorem}
	\Cref{L2thm}, proven in \Cref{Section: proof of L2thm}, has several interesting aspects worth highlighting. First of all, it indicates that the RFF-MMD tests can achieve the optimal separation rate ${n}^{-2s/(4s+d)}$ when $R$ is larger than ${n}^{2d/(4s+d)}$. This in turn suggests that this optimality can be attained in sub-quadratic time when the underlying distributions are sufficiently smooth (i.e., $s > d/4$). Indeed, the computational time becomes linear in $N$ as $d/s \rightarrow 0$. On the other hand, the computational complexity may need to exceed quadratic-time to achieve the minimax separation rate in non-smooth cases. 
	
	An astute reader may have realized that \Cref{L2thm} is established for distributions on the bounded domain, which differs from the unbounded setting in the prior work~\citep{schrab2023mmdaggregated}. We impose this additional constraint for analytical tractability, and in fact, the bounded domain is frequently assumed in minimax analysis~\citep[e.g.,][]{Ingster1987,ingster1993,arias2018remember}. Nevertheless, it is important to point out that the worst-case instance used for deriving the minimax lower bound is defined on a bounded domain, say $[0,1]^d$. Therefore the minimax rate remains unchanged for the bounded distributions that we consider.

	\subsection{Uniform consistency in MMD metric} \label{Section: Uniform consistency in MMD metric}
	In the previous subsection, we demonstrated that the RFF-MMD tests can achieve the minimax separation rate in sub-quadratic time complexity. It is worth pointing out that this result is presented against the Sobolev smooth $L_2$ alternatives, and optimal choices of the bandwidth $\lambda$, which parameterizes the kernel $k_\lambda$, and $R$ that balances between computational and statistical trade-offs may vary depending on classes of alternatives. To illustrate this point, we now turn to studying the uniform separation rate of the RFF-MMD test with respect to the MMD metric equipped with a generic kernel $k$, and discuss the choice of $R$ that strikes the aforementioned trade-offs. Given a kernel $k$, consider the alternative $\mathcal P_1(\mathcal C,\delta,\epsilon)$ with a class of distribution pairs, $\mathcal C$, and a MMD metric,
	$
	\delta_{\text{MMD}}(P_X,P_Y)=\mathrm{MMD}(P_X,P_Y;\mathcal H_k).
	$
	As formally shown in \citet{kim2023diff}, the minimax rate of testing against the MMD metric satisfies $\rho^\star\asymp n^{-1/2}$ where $n = \min\{{n_1,n_2}\}$. The next theorem demonstrates that the number of random features $R$ required to achieve the minimax separation rate in terms of the MMD metric is of order $N$ where recall $N = n_1 + n_2$; thereby the overall runtime becomes $Nn$ in the sample size. The proof can be found in \Cref{Section: proof of MMDthm}.
	\begin{theorem}\label{MMDthm}
		Consider the tests $\Delta_{{n_1},{n_2},R}^{\alpha}$ and
		$\Delta_{{n_1},{n_2},R}^{\alpha,u}$ equipped with a characteristic kernel $k$ satisfying the assumptions of \Cref{negthm}, and suppose further that
		$0\leq k(x,y)\leq K$ for all $x,y\in\mathbb R^d$. Then, the test $\Delta_{{n_1},{n_2},R}^{\alpha}$ with $R={n} = \min\{{n_1,n_2}\}$ achieves the minimax separation rate, satisfying
		$$
		\rho\big(\Delta_{{n_1},{n_2},R}^{\alpha},\,\beta,\,\mathcal C,\,\delta_{\mathrm{MMD}} \big)\leq C_{\mathrm{MMD}}(\alpha,\beta,K){n}^{-1/2}
		$$
		for some positive constant $C_{\mathrm{MMD}}(\alpha,\beta,K).$ The same guarantee also holds for $\Delta_{{n_1},{n_2},R}^{\alpha,u}$. Moreover, the computational cost of the corresponding test statistics $\text{r}\widehat {\mathrm{MMD}}_b^2$ and $\text{r}\widehat {\mathrm{MMD}}_u^2$ is $O(N{n}d).$
	\end{theorem}
    As pointed out in \citet{Carles2023Compress}, existing analyses of the RFF-MMD test yield only a cubic-time complexity bound when attempting to match the power of the standard MMD test. In contrast, \Cref{MMDthm} shows that the RFF-MMD test can achieve the same minimax separation rate with only quadratic-time complexity. Indeed, we can further improve this point:~when properly carving out the distributions of interest, it becomes possible to achieve the same separation rate of $n^{-1/2}$ in sub-quadratic or even linear-time complexity. To demonstrate this point, denote the U-statistic in Equation~\eqref{rMMDu} with a single random Fourier feature (i.e., $R=1$) as $U_1$. One of the crucial steps in the proof of \Cref{MMDthm} involves finding an upper bound for the expectation $\mathbb E_{\omega}[(\mathbb E_{X\times Y}[U_1\,|\, \omega ])^2]$. Since the kernel is uniformly bounded and $\mathbb{E}[U_1] = \mathrm{MMD}^2(P_X,P_Y;\mathcal H_{k})$, the previous expectation is bounded above by $\mathrm{MMD}^2(P_X,P_Y;\mathcal H_{k})$, up to a constant. Our analysis utilizes this somewhat crude, but not universally improvable, upper bound, which is where the quadratic-time complexity arises.

	Now let us consider a subclass of distribution pairs $\mathcal C' \subseteq\mathcal C$. Suppose that there exist some constants $c \in (1,2]$ and $C > 0$ such that the following inequality
	\begin{equation}\label{momentbound}
		\begin{aligned}
			\mathbb E_{\omega}\big[\big(\mathbb E_{X\times Y}[U_1\,|\, \omega ]\big)^2\big] &  \leq C \big(\mathbb E_{\omega}\big[\mathbb E_{X\times Y}[U_1\,|\, \omega]\big]\big)^c = C \big( \mathrm{MMD}^2(P_X,P_Y;\mathcal H_{k})\big)^c
		\end{aligned}
	\end{equation}
	holds for all $(P_X,P_Y)\in \mathcal C'$ (see Remark~\ref{Remark: range of c} of the appendix for a discussion on the range of $c$). Against this class of alternatives $\mathcal{C}'$, our proof shows that the RFF-MMD test achieves ${n}^{-1/2}$-separation rate within sub-quadratic time. Specifically, the time complexity depends on the value of $c$ in Equation~\eqref{momentbound} with a precise computational cost of $O(N{n}^{2-c}d)$ for $c \in (1,2]$. As a concrete example, consider the class of pairs of Gaussian distributions with a common fixed covariance $\Sigma\in \mathbb R^{d\times d}$, denoted as
	\begin{align}\label{normalclass}
		\mathcal C_{N,\Sigma}:=\big\{(P_X,P_Y)\in \mathcal{P}_{\mathrm{conti}} \,\big|\, P_X = N(\mu_X,\Sigma),~ P_Y = N(\mu_Y,\Sigma) ~ \text{where $\mu_X, \mu_Y\in \mathbb{R}^d$}  \big\},
	\end{align}
	and set $\mathcal C' = \mathcal C_{N,\Sigma}$. For this Gaussian subclass and a generic Gaussian kernel given as 
	$$
	k_\lambda(x,y)=\prod^{d}_{i=1}\frac{1}{\sqrt{\pi}\lambda_i}e^{-\frac{(x_i-y_i)^2}{\lambda_i^2}}
	$$ with bandwidth $\lambda=(\lambda_1,\ldots,\lambda_d)^\top \in (0,\infty)^d$, we prove that the inequality in Equation~\eqref{momentbound} holds with the constant $c=2$. This main building block allows us to show the following proposition, indicating that the RFF-MMD test achieves the uniform separation rate of ${n}^{-1/2}$ in linear-time complexity.

	\begin{proposition}\label{MMDproposition}
		For the class of distribution pairs $\mathcal C_{N,\Sigma}$ and the Gaussian kernel $k_\lambda(x,y)$ with any fixed bandwidth $\lambda=(\lambda_1,\ldots,\lambda_d)^\top \in (0,\infty)^d$, there exist some positive constants $C_1(\beta,d,\lambda,
		\Sigma)$ and $C_2(\alpha,\beta,d,
		\lambda)$ such that $\Delta_{{n_1,n_2},R}^{\alpha,\lambda}$ with the choice of $R \geq C_1(\beta,d,\lambda,
		\Sigma)$ satisfies
		$$
		\rho\big(\Delta_{{n_1,n_2},R}^{\alpha,\lambda},\,\beta,\,\mathcal C_{N,\Sigma},\,\delta_{\mathrm{MMD}}\big)\leq C_2(\alpha,\beta,d,
		\lambda){n}^{-1/2},
		$$
		and the computational cost of the corresponding estimator $\text{r}\widehat {\mathrm{MMD}}_b^2$ is $O(Nd).$ This result also holds for the test $\Delta_{{n_1,n_2},R}^{\alpha,u,\lambda}$ with the same choice of $R$ and its corresponding estimator $\text{r}\widehat {\mathrm{MMD}}_u^2.$
	\end{proposition}
	Proposition \ref{MMDproposition}, proven in \Cref{Section: proof of MMDproposition}, states that the RFF-MMD test requires only a fixed number of random features to match the uniform separation rate of the original MMD test. At first glance, this appears to contradict Theorem \ref{negthm}, which demonstrates the pointwise inconsistency of the test when the number of random features $R$ is fixed. However, this is not a contradiction as Proposition \ref{MMDproposition} assumes a smaller, specific class of distributions, whereas Theorem \ref{negthm} considers all possible distributions. Notably, the distributions that lead to the inconsistency demonstrated in Theorem \ref{negthm} do not fall within the class $\mathcal{C}_{N,\Sigma}$. 
	
	While we focus on the class of Gaussian distributions for technical tractability, we believe that Proposition \ref{MMDproposition} holds for a broader class of distributions as evidenced by our empirical studies. It would be of great interest to further explore classes of distributions for which the RFF-MMD test offers significant computational gains over the original MMD test, while maintaining nearly the same power. We leave this topic for future work.

	\section{Numerical studies} \label{Section: Numerical studies}
	In this section, we compare the empirical power and computational time of RFF-MMD tests with other computationally efficient methods such as linear-time statistics~\citep[lMMD;][]{gretton2012kernel,gretton2012b}, block-based statistics~\citep[bMMD;][]{zaremba2013b}, incomplete U-statistics~\citep[incMMD;][]{Yamada2019,schrab2022efficient} under several different scenarios. Within each scenario, we run RFF-MMD tests with varying numbers of random features $R \in\{10, 200, 1000\}$, and also run the quadratic time MMD test~\citep{gretton2012b} as a benchmark for comparison. In our simulations, all kernel tests employ a Gaussian kernel with the bandwidth selected using the median heuristic. The significance level is set at $\alpha=0.05$ and the critical value of each test is determined by using permutation or bootstrap methods with $B=199$ Monte Carlo iterations. The power of each test is approximated by averaging the results over $2000$ repetitions. 
	
	The specific scenarios that we consider in our simulation studies are described as follows. 
	\begin{itemize}
		\item \textbf{Scenario 1: Univariate Gaussians.} Our first experiment is concerned with comparing two Gaussian distributions on $\mathbb R$ with a mean difference or a variance difference. Specifically, we first evaluate the performance of the methods in distinguishing $P_X=N(0, 1)$ from $P_Y=N(\mu,1)$ by (i) varying $\mu$ from $0$ to $0.3$ and (ii) varying the sample sizes $n_1 = n_2$ with a fixed mean difference of $\mu=0.15$. We conducted a similar experiment to evaluate the performance of the methods in distinguishing $P_X = N(0, 1)$ from $P_Y = N(0, \sigma^2)$ by (i) varying $\sigma$ from $0.5$ to $2$ and (ii) varying the sample sizes $n_1 = n_2$ with fixed variance $\sigma=1.3$.		
		\item \textbf{Scenario 2: High-dimensional Gaussians.} We also compared the power of the tests for distinguishing two Gaussian distributions with different mean vectors or covariance matrices in high-dimensional settings. For location alternatives, we let $\boldsymbol{\mu}_{0.1,20}\in \mathbb R^d$ be a vector whose first $20$ coordinates are $0.1,$ and the others are 0. We set $P_X=N(\boldsymbol 0_d, I_{d\times d})$ and $P_Y=N(\boldsymbol{\mu}_{0.1,20}, I_{d\times d}),$ and also report the test powers by varying $d$ from $20$ to $2000$ or the sample sizes $n_1 = n_2$ with fixed $d=1000$. For scale alternatives, we set $P_X=N(\boldsymbol 0_d, I_{d\times d})$ and $P_Y=N(\boldsymbol 0_d, \sigma^2 I_{d\times d}),$ and vary $\sigma$ from 0.95 to 1.1 or the sample sizes $n_1 = n_2$ while fixing $\sigma = 1.03$.
		
		\item \textbf{Scenario 3: Perturbed uniforms.} Motivated by the experiments conducted in \citet{schrab2022efficient,schrab2023mmdaggregated,biggs2023fuse}, we investigate the test powers for capturing perturbations in uniform distributions on $\mathbb R$ or $\mathbb R^2$. Specifically, for $t\in \mathbb R^d,$ we set the density of the null distribution as $f_X(t)=\mathds 1_{[0,1]^d}(t)$ and that of the alternative as $f_Y(t) = \mathds 1_{[0,1]^d}(t) + \alpha E_{d,p}(t)$ where $\alpha \in [0,p]$ is perturbation amplitude and $E_{d,p}(t)$ is the $d$-dimensional perturbation function of size $p$, defined as $E_{d,p}(t):=p^{-1}e^d\sum_{u\in\{1,\ldots,p\}^d}\theta_u\prod^d_{i=1} G(pt_i-u_i)$ with $\{\theta_u\}_{u\in\{1,\ldots,p\}^d}\in\{-1,1\}^{p^d}.$ The perturbation shape function $G(t)$ is given by: 
		$$
		G(t):=\exp \left(-\frac{1}{1-(4 t+3)^2}\right) \mathds{1}_{\left(-1,-\frac{1}{2}\right)}(t)-\exp \left(-\frac{1}{1-(4 t+1)^2}\right) \mathds{1}_{\left(-\frac{1}{2}, 0\right)}(t), \quad t \in \mathbb{R}.
		$$
		We set $E_{1,2}(t)$ as a one-dimensional alternative and $E_{2,1}(t)$ as a two-dimensional alternative. In this case, the perturbation amplitude $\alpha=0$ implies the null hypothesis and we consider different scenarios by varying $\alpha$ from $0$ to $0.9$.
		Additionally, we fix the perturbation amplitude at $\alpha=0.6$ for $E_{1,2}(t)$ and $\alpha=0.45$ for $E_{2,1}(t),$ and vary the sample sizes $n_1 = n_2$.
		
		\item \textbf{Scenario 4: MNIST.} To evaluate the performance of the methods in real-world settings, we consider a task of distinguishing between the distribution of even-number images and the distribution of odd-number images in the MNIST dataset. Each data point $z$ is an image with dimension $d=28 \times 28 =784$ (without downsampling) or $d=7 \times 7 =49$ (with downsampling), with labels $L_z\in\{0,1,\ldots ,9\}$. We collect the images of even numbers to define a distribution $ P_{\text{even}}:=\{z : L_z\in\{0,2,4,6,8\}\}$ and collect the images of odd numbers to define another distribution $P_{\text{odd}}:=\{z : L_z\in\{1,3,5,7,9\}\}$. Given a mixing rate $\gamma \in[0,1]$, we set $P_X = P_{\text{even}}$ and $P_Y = (1-\gamma) P_{\text{even}} + \gamma P_{\text{odd}}$. Accordingly, we regard the case $\gamma=0$ as the null hypothesis and vary $\gamma$ from $0$ to $0.3$ to evaluate the power performance. When we vary the sample sizes $n_1 = n_2$, we fix the mixing rate at $\gamma=0.1$.

	\end{itemize}
	
	\begin{figure}[t]
		
		\includegraphics[width=\textwidth]{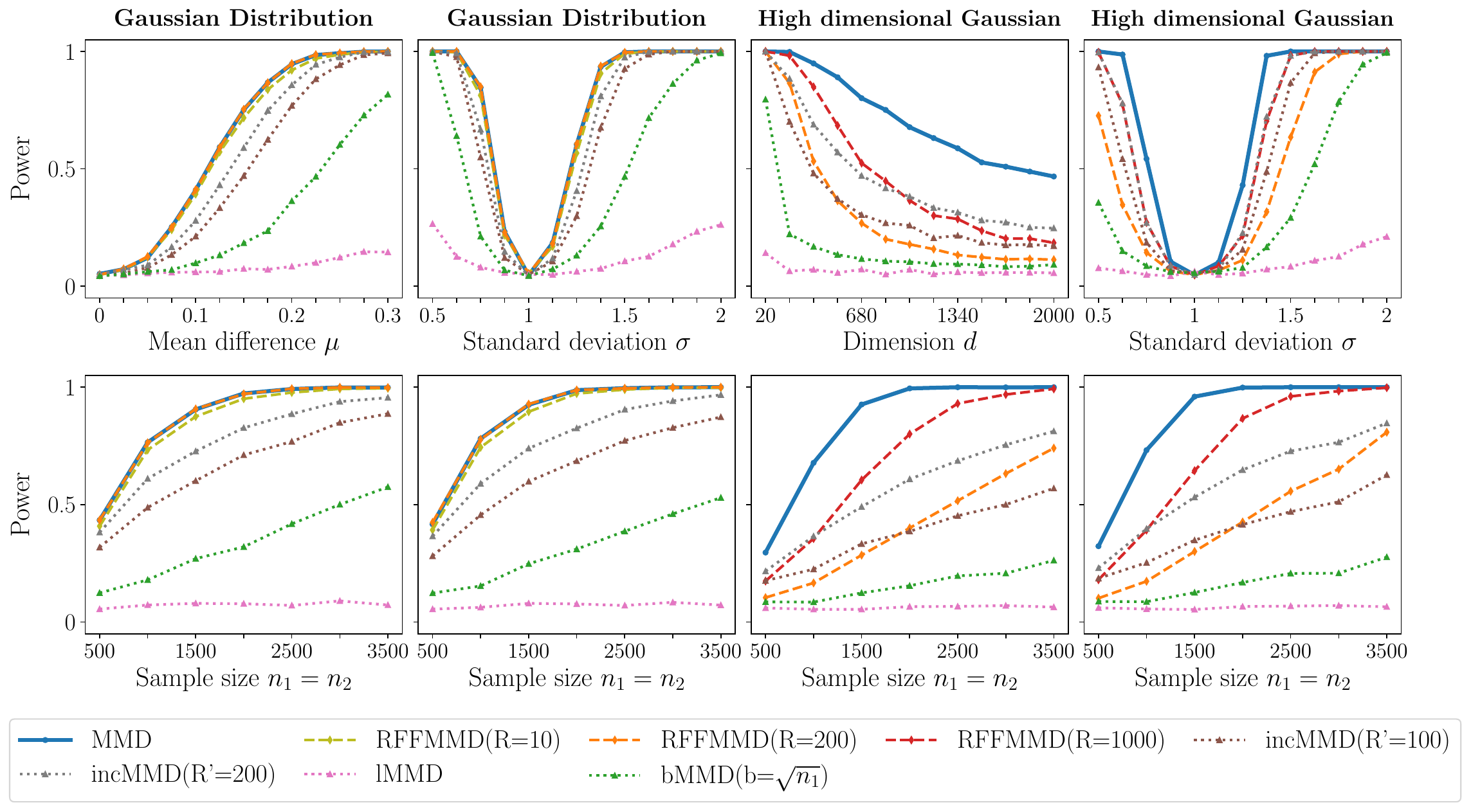}
		\caption{Power experiments with two different settings: (i) univariate Gaussian distribution, (ii) high-dimensional Gaussian distribution. The sample sizes are set to ${n_1}={n_2}=1000$ for the first row of graphs. For the second row of graphs, parameters are set to $\mu=0.15$ in the first column, $\sigma=1.3$ in the second column, $d=1000$ in the third column, and $\sigma=1.03$ in the fourth column.} \label{simulation: first two}
	\end{figure}
	
\begin{figure}[t]
		\begin{center}
			\includegraphics[width=1\textwidth]{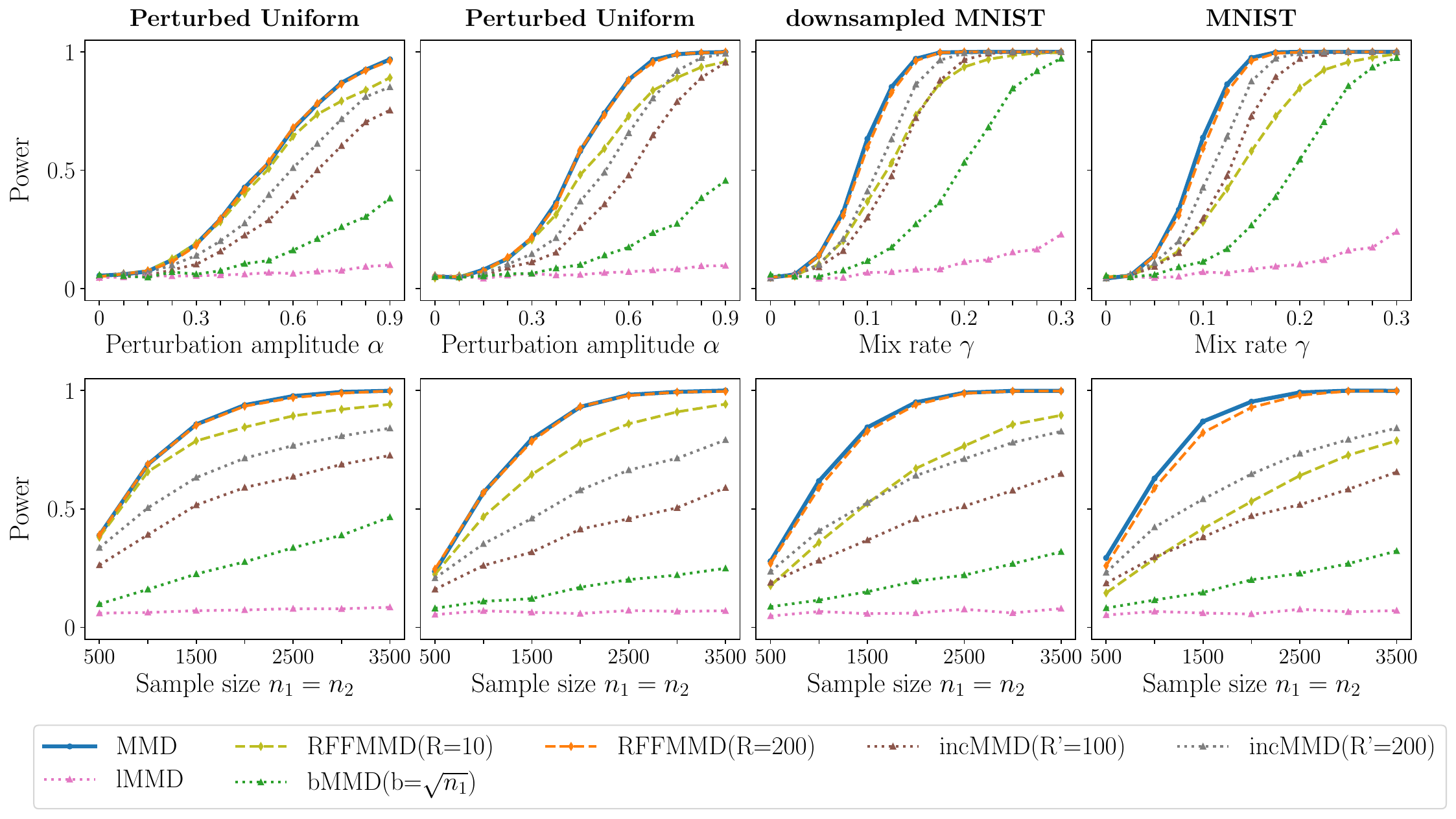}
			\caption{Power experiments with two different settings: (i) perturbed uniform distribution, (ii) MNIST. The sample sizes are set to ${n_1}={n_2}=1000$ for the first row of graphs. For the second row of graphs, parameters are set to $\alpha=0.6$ in the first column, $\alpha=0.45$ in the second column, and $\gamma=0.1$ in the third and last column.} \label{simulation: last}
		\end{center}
	\end{figure}

	The simulation results for the first two scenarios are displayed in Figure~\ref{simulation: first two}, whereas the simulation results for the last two scenarios can be found in Figure~\ref{simulation: last}. We first note that the test power of RFF-MMD test increases monotonically with $R$, converging to the power of the quadratic-time MMD test. This empirically illustrates that the RFF-MMD test approximates the quadratic time MMD test as $R$ increases. Also, the empirical results demonstrate that different values of $R$ are required depending on the underlying distribution to match the power of the RFF-MMD test with that of the quadratic-time MMD test. Specifically, the RFF-MMD test matched the power of the original MMD test in all cases when $R=200$, except in Scenario 2, where it matched when $R=1000$. It is also worth noting that the RFF-MMD test outperforms other efficient methods in Scenarios 1 and 3, even with $R$ as small as 10. 
	
	In Scenario 2, which involves a high-dimensional Gaussian setting, we observed that the power of the RFF-MMD test drops more sharply than that of the incMMD test when the sample size is fixed and the dimension increases. Conversely, when the dimension is fixed and the sample size increases, the power of the RFF-MMD test converges to that of the quadratic time MMD test more quickly than the incMMD test. A similar phenomenon was observed in Scenario 4: as the dimension increases from downsampled MNIST to MNIST data, the power curve of the RFF-MMD test shifts downward, while the power curves of other methods show little variation for the same mixing rate. However, when fixing the mixing rate and varying the sample size, the power of the RFF-MMD test increases faster than that of the incMMD test. This can be explained by the fact that the RFF-MMD test involves kernel approximation. As the dimension increases while the number of random features remains fixed, the accuracy of the kernel approximation decreases, leading to a relatively faster decline in power compared to the incMMD test. Conversely, when the dimension is fixed and the sample size varies, the incMMD test considers only a subset of the samples for computing the test statistic, resulting in a relatively slower increase in power compared to the RFF-MMD test.

We empirically measured the computational time of the considered methods under Scenario 1, as recorded in Table \ref{simulation: speed}. In the experiments, we varied the sample size from $250$ to $8000$, with a mean difference of $\mu=0.15$. To ensure the efficiency of the experiments, we measured the time taken to compute the test statistic once, rather than the time taken to perform the permutation test. The results were approximated by averaging over $1000$ repetitions. From Table \ref{simulation: speed}, we experimentally confirmed that while the computational time of the conventional MMD increases quadratically with the sample size, the computational times of RFF-MMD and incMMD increase linearly. Additionally, the last row of Table \ref{simulation: speed} demonstrates that the time increases linearly with the number of features, which aligns with the theoretical computational time of $O(NRd)$ for RFF-MMD. We also note that similar patterns were observed in other simulation scenarios.

	\begin{table}[h!]
		\caption{Computational time (in seconds) comparisons of the considered methods under Scenario 1.} \label{simulation: speed}
		\vspace{-0.1cm}
		\begin{center}
			\begin{small}
				\begin{tabular}{ccccccccc}
					\toprule
					\begin{tabular}{@{}c@{}}Sample \\ size\end{tabular} 
					& MMD
					& \begin{tabular}{@{}c@{}}RFF-MMD \\ $R=10$\end{tabular} 
					& \begin{tabular}{@{}c@{}}RFF-MMD \\ $R=200$\end{tabular}   
					& \begin{tabular}{@{}c@{}}RFF-MMD \\ $R=1000$\end{tabular}  
					& \begin{tabular}{@{}c@{}}incMMD \\ $R'=100$\end{tabular}
					& \begin{tabular}{@{}c@{}}incMMD \\ $R'=200$\end{tabular}
					& lMMD
					& \begin{tabular}{@{}c@{}}bMMD \\ $b=n^{1/2}$\end{tabular} 
					\\
					\midrule
					250 & 0.0088 & 0.0002 & 0.0009 & 0.0070 & 0.0057 & 0.0084 & 0.0001 & 0.0006 \\
					500 & 0.0411 & 0.0003 & 0.0017 & 0.0130 & 0.0140 & 0.0251 & 0.0001 & 0.0019\\
					1000 & 0.1946 & 0.0004 & 0.0051 & 0.0254 & 0.0325 & 0.0681 & 0.0002 & 0.0053\\
					2000 & 0.7983 & 0.0006 & 0.0097 & 0.0485 & 0.0744 & 0.1497 & 0.0004 & 0.0155\\
					4000 & 3.2662 & 0.0010 & 0.0192 & 0.0966 & 0.1567 & 0.3128 & 0.0007 & 0.0439\\
					8000 & 13.247 & 0.0020 & 0.0371 & 0.1933 & 0.3189 & 0.6391 & 0.0015 & 0.1426\\
					\bottomrule
				\end{tabular}
			\end{small}
		\end{center}
	\end{table}

	\section{Discussion} \label{Section: Discussion}
	In this work, we laid the theoretical foundations for kernel MMD tests using random Fourier features. Firstly, we proved that pointwise consistency is attainable if and only if the number of random Fourier features tends to infinity with the sample size. This observation naturally motivates an investigation into the optimal choice of the number of random Fourier features that strikes a balance between computational efficiency and statistical power.  We explored this time-power trade-off under the minimax testing framework, and showed that it is possible to attain minimax separation rates within sub-quadratic time under certain distributional assumptions. We also validated these theoretical findings through numerical studies.
	
	Our work opens up several promising avenues for future work. A natural extension of our work is to adapt our techniques to other kernel-based inference methods, such as the Hilbert--Schmidt independence criterion, and investigate fundamental time-power trade-offs in different applications. From a technical standpoint, it remains open whether a similar result to \Cref{L2thm} can be obtained for distributions with unbounded supports. Future work can also attempt to extend our results in \Cref{Section: Optimal rate of the number of random features} to other metrics such as the Hellinger distance~\citep[e.g.,][]{hagrass2022spectral} and explore further improvements under other smoothness conditions. Finally, it would be of interest to consider deterministic Fourier features, which are shown to better approximate a kernel than random Fourier features~\citep[e.g.,][]{wesel2021large}, and apply those in our application. We leave all these intriguing yet challenging problems to future work.

    \paragraph{Acknowledgments} We acknowledge support from the Basic Science Research Program through the National Research Foundation of Korea (NRF) funded by the Ministry of Education (2022R1A4A1033384), and the Korea government (MSIT) RS-2023-00211073. We are grateful to Yongho Jeon and Gyumin Lee for their careful proofreading and helpful discussion. 
 
	\bibliographystyle{apalike}
	\bibliography{ref}
	
	\appendix
	
	\section{Technical lemmas}
	In this section, we collect technical lemmas used in the main proofs of our results. 
	
	\begin{lemma}[{\citealp[Bochner's theorem,][]{bochner1933}}]\label{bochner}
		A translation-invariant bounded continuous kernel $k(x,y)=\kappa(x-y)$ on $\mathbb R^d$ is positive definite if and only if there exists a finite non-negative Borel measure $\Lambda$ on $\mathbb R^d$ such that
		$$
		k(x,y)=\int_{\mathbb R^d} e^{\sqrt{-1}\omega^\top(x-y)}d\Lambda(\omega).
		$$
		
	\end{lemma}

	The following result is commonly known as Young's convolution inequality. 
	\begin{lemma}[{\citealp[][Theorem 3.9.4]{Bogachev2007}}]\label{young}
		Let $p,q,r\in[1,\infty]$ satisfy $1/p + 1/q = 1+1/r.$ Then, for any functions $f \in L_p(\mathbb R^d)$ and $g \in L_q(\mathbb R^d)$,
		$$
		\|f \ast g\|_r \leq \|f\|_p \|g\|_q.
		$$
		
	\end{lemma}

    We next collect useful asymptotic tools from \cite{chung2016multivariate} to analyze the limiting behavior of permutation distributions.
	\begin{lemma}[{\citealt[Lemma A.2]{chung2016multivariate}}]\label{chung.lemmaA2}
		Suppose $X^n=(X_1,\ldots,X_n)$ has distribution $P_n$ in $\mathcal X_n,$ and $\boldsymbol{G}_n$ is a finite group of transformations $g$ of $\mathcal X_n$ onto itself. Also, let $G_n$ be a random variable that is uniform on $\boldsymbol{G}_n$. Assume $X^n$ and $G_n$ are mutually independent. For a $d$-dimensional test statistic $B_n=B_n(X^n),$ let $\hat R_n^B$ denote the randomization distributions of a $d$-dimensional random vector $B_n,$ defined by
		\begin{align}\label{lemmaA2.eq1}
			\hat R^B_n(t)= \frac 1 {|\boldsymbol{G}_n|}\sum_{g\in \boldsymbol{G}_n} \mathds 1\{B_n(gX^n)\leq t\}.
		\end{align}
		Suppose, under $P_n,$
		\begin{align}\label{lemmaA2.eq2}
			B_n(G_nX^n)\xrightarrow[]{p}b
		\end{align}
		for a constant $b\in \mathbb R^d.$ Then under $P_n,$
		$$
		\hat R^B_n(t)= \frac 1 {|\boldsymbol{G}_n|}\sum_{g\in \boldsymbol{G}_n} \mathds 1\{B_n(gX^n)\leq t\} \xrightarrow[]{p}\delta_b(t)\quad\text{if}~t\neq b,
		$$
		where $\delta_c$ denotes the distribution function corresponding to the point mass function at $c\in \mathbb R^d.$
	\end{lemma}

	\begin{lemma}[{\citealt[Lemma A.3]{chung2016multivariate}}]\label{chung.lemmaA3}
		Let $B_n$ and $T_n$ be sequences of $d$-dimensional random variables satisfying Equation \eqref{lemmaA2.eq2} and
		$$
		(T_n(G_nX^n),T_n(G'_nX^n))\xrightarrow{d}(T,T'),
		$$
		where $T$ and $T'$ are independent, each with common $d$-variate cumulative distribution function $R^T(\cdot).$ Let $\hat R^{T+B}_n(t)$ denote the randomization distribution of $T_n + B_n,$ defined in Equation \eqref{lemmaA2.eq1} with $B$ replaced by $T+B.$ Then, $\hat R^{T+B}_n(t)$ converges to the cumulative distribution function of $T+b$  in probability. In other words, 
		$$
		\hat R^{T+B}_n(t) = \frac 1 {|\boldsymbol{G}_n|}\sum_{g\in \boldsymbol{G}_n} \mathds 1\{T_n(gX^n)+B_n(gX^n)\leq t\}\xrightarrow{p}R^{T+b}(t),
		$$
		if $R^{T+b}$ is continuous at $t\in \mathbb R^d,$ where $R^{T+b}(\cdot)$ denotes the corresponding $d$-variate cumulative distribution function of $T+b.$
	\end{lemma}

	\begin{lemma}[{\citealt[Lemma A.6]{chung2016multivariate}}]\label{chung.lemmaA6}
		Suppose the randomization distribution of a test statistic $T_n$ converges to $T$ in probability. In other words,
		$$
		\hat R^{T}_n(t) = \frac 1 {|\boldsymbol{G}_n|}\sum_{g\in \boldsymbol{G}_n} \mathds 1\{T_n(gX^n)\leq t\}\xrightarrow{p}R^{T}(t),
		$$
		if $R^T$ is continuous at $t\in \mathbb R^d,$ where $R^T(\cdot)$ denotes the corresponding cumulative distribution function of $T$. Let $h$ be a measurable map from $\mathbb R^d$ to $\mathbb R^s.$ Let $C$ be the set of points in $\mathbb R^d$ for which $h$ is continuous. If $\mathbb P(T\in C)=1,$ then the randomization distribution of $h(T_n)$ converges to $h(T)$ in probability.
	\end{lemma}

	\begin{lemma}[{\citealt[Theorem 2.1]{chung2016multivariate}}]\label{lemma2}
		Suppose that $X_1,\ldots,X_{n_1}$ are ${n_1}$ i.i.d.~random samples from the $d$-dimensional distribution $P_X$, where $X_i=(X_{i,1},\ldots,X_{i,d})^\top$ for $i=1,\ldots,n_1$ with mean vector $\mu$ and covariance matrix $\Sigma_X$ , and independently, $Y_1,\ldots,Y_{n_2}$ be ${n_2}$ i.i.d.~random samples from the $d$-dimensional distribution $P_Y$, where $Y_i=(Y_{i,1},\ldots,Y_{i,d})^\top$ for $i=1,\ldots,n_2$ with the common mean vector $\mu$ and covariance matrix $\Sigma_Y.$ Let $N={n_1}+{n_2}$ and write $Z=(Z_1,\ldots,Z_N)=(X_1,\ldots,X_{n_1},Y_1,\ldots,Y_{n_2}).$ Consider a test statistic $T_{{n_1},{n_2}}(Z_1,\ldots,Z_N)={n_1}^{-1/2}\big[\sum_{i=1}^{n_1} X_i-\frac {n_1} {n_2} \sum_{j=1}^{n_2} Y_j\big]$ and its permutation distribution
		$$
		\hat R^T_{{n_1},{n_2}}(t)=\frac{1}{N!}\sum_{\pi\in \boldsymbol{G}_N} \mathds 1\{T_{{n_1},{n_2}}(Z_{\pi(1)},\ldots,Z_{\pi(N)})\leq t\}
		$$
		where $\boldsymbol{G}_N$ denotes the $N!$ permutations of $\{1,2,\ldots,N\}$ and $t\in\mathbb R^{d}$. Assume $0<\operatorname{Var}(X_{i,k})<\infty$ and $0<\operatorname{Var}(Y_{j,k})<\infty$ for all $i=1,\ldots,{n_1},$ $j=1,\ldots,{n_2},$ and $k=1,\ldots,d.$ Let ${n_1},{n_2}\rightarrow\infty,$ $p=\lim_{{n_1},{n_2}\rightarrow\infty} \frac{{n_1}}{{n_1}+{n_2}}$ and assume that $\bar \Sigma = \frac p {1-p}\Sigma_X+\Sigma_Y$ is positive definite. Then,
		$$
		\sup_{t\in \mathbb R^{d}}\big|\hat R^T_{{n_1},{n_2}}(t)-G(t)\big|\xrightarrow{p}0,
		$$
		where $G$ denotes the $d$-variate normal distribution with mean $\boldsymbol{0}$ and variance $\bar \Sigma.$
		
	\end{lemma}

    The following result is a slight modification of \citet[][Proposition 1]{ramdas2015decreasing} tailored to our kernel setting.
        \begin{lemma}[{\citealt[Proposition 1]{ramdas2015decreasing}}]\label{gaussian MMD lemma}
            Suppose $P_X=N(\mu_X,\Sigma)$ and $P_Y=N(\mu_Y,\Sigma)$. The squared MMD between $P_X$ and $P_Y$ using a Gaussian kernel $\kappa_\lambda(x-y)=\prod^{d}_{i=1}\frac{1}{\sqrt{\pi}\lambda_i}\exp{\bigl\{-\frac{(x_i-y_i)^2}{\lambda_i^2}\bigr\}}$ has the following explicit form:
            $$
            \mathrm{MMD}^2(P_X,P_Y;\mathcal H_{k_\lambda}) = 2\left(\frac{1}{4\pi}\right)^{d / 2} \frac{1-\exp \big\{-(\mu_X-\mu_Y)^{\top}\left(\Sigma+D(\lambda^2/4)\right)^{-1} (\mu_X-\mu_Y) / 4\big\}}{\left|\Sigma+D(\lambda^2/4)\right|^{1 / 2}},
            $$
            where $D(\lambda^2/4)=\mathrm{diag}(\lambda_1^2/4,\ldots,\lambda_d^2/4).$
        \end{lemma}
 		
 		The next lemma facilitates the calculation of $\mathbb E_{\omega}\big[\big(\mathbb E_{X\times Y}[U_1\,|\, \omega ]\big)^2\big]$. The proof can be found in Appendix~\ref{Section: Proof of lemma: mmdcalculate}.
 		
 		\begin{lemma}\label{mmdcalculate} Let $X',X''$ and $Y',Y''$ be the independent copies of $X$ and $Y$, respectively. Then, the following two equations hold:
 			\begin{align*}
 				\mathbb E_{\omega}\big[\big(\mathbb E_{X\times Y}[U_1\,|\, \omega ]\big)^2\big]&=2\kappa(0)\mathrm{MMD}^2(P_{X-X'},P_{X''-Y};\mathcal H_{k})+2\kappa(0)\mathrm{MMD}^2(P_{Y-Y'},P_{X-Y''};\mathcal H_{k})\\
 				&\quad-\kappa(0)\mathrm{MMD}^2(P_{X+X'},P_{Y+Y'};\mathcal H_{k}) \quad \text{and} \\[.5em]
 				\mathbb E_{\omega}\big[\big(\mathbb E_{X\times Y}[U_1\,|\, \omega ]\big)^2\big]&=2\kappa(0)\mathrm{MMD}^2(P_{X+X'},P_{X''+Y};\mathcal H_{k})+2\kappa(0)\mathrm{MMD}^2(P_{Y+Y'},P_{X+Y''};\mathcal H_{k})\\
 				&\quad-\kappa(0)\mathrm{MMD}^2(P_{X+X'},P_{Y+Y'};\mathcal H_{k}).
 			\end{align*}
 		\end{lemma}
 		
 		The next lemma serves as a main building block in the proof of Propositions~\ref{MMDproposition}. The proof can be found in Appendix~\ref{Section: Proof of lemma: moment bound for Gaussian}.

 		\begin{lemma} \label{lemma: moment bound for Gaussian}
 			For the class of distribution pairs, $\mathcal C_{N,\Sigma}$, in Equation \eqref{normalclass} and the Gaussian kernel $k_\lambda(x,y)$ with any fixed bandwidth $\lambda=(\lambda_1,\ldots,\lambda_d)^\top \in (0,\infty)^d$, the inequality in Equation \eqref{momentbound} holds with $c= 2$. Specifically, there exists a constant  $C=C(d,\lambda,\Sigma)>0$ such that 
 			\begin{align*}
 				\mathbb E_{\omega}\big[\big(\mathbb E_{X\times Y}[U_1\,|\, \omega ]\big)^2\big] &  \leq C \big(\mathbb E_{\omega}\big[\mathbb E_{X\times Y}[U_1\,|\, \omega]\big]\big)^c = C(d,\lambda,\Sigma) \big( \mathrm{MMD}^2(P_X,P_Y;\mathcal H_{k_\lambda})\big)^2.
 			\end{align*}
 		\end{lemma}
 		\medskip
 		
 		\begin{remark} \noindent \label{Remark: range of c}
 			Regarding the range of $c\in(1,2]$ in Equation \eqref{momentbound}, note that when $c > 2$, the inequality in Equation~\eqref{momentbound} yields $\mathrm{MMD} \gtrsim 1$ and this is not of our interest (in fact, this condition becomes vacuous since $\mathrm{MMD}$ using a bounded kernel is bounded above by a constant). More specifically, by Jensen's inequality, we have $\mathrm{MMD}^4\leq\mathbb E_{\omega}\big[\big(\mathbb E_{X\times Y}[U_1\,|\, \omega ]\big)^2\big]$ and then the inequality in Equation~\eqref{momentbound} implies
 			\begin{align*}
 				& \mathrm{MMD}^4(P_X,P_Y;\mathcal H_{k_\lambda}) \leq C \mathrm{MMD}^{2c}(P_X,P_Y;\mathcal H_{k_\lambda}) \\
 				\equiv ~& \mathrm{MMD}^{2(2-c)}(P_X,P_Y;\mathcal H_{k_\lambda}) \leq C \\
 				\equiv ~& \frac{1}{C^{\frac{1}{2(c-2)}}}  = C' \leq \mathrm{MMD}(P_X,P_Y;\mathcal H_{k_\lambda}).
 			\end{align*}
 			Therefore we may see some computational gain when $c>2$ but it is  only possible when the minimum separation is of constant order as $\mathrm{MMD} \gtrsim 1$.
 		\end{remark}

	\section{Proofs}
	
	\paragraph{Notation and terminology.} We start by organizing the notation and the terminology we use throughout this appendix. Unless explicitly stated otherwise, the symbol $\mathbb P(\cdot)$ denotes the probability measure that takes into account all inherent uncertainties. In addition, we represent constants as $C_1,C_2,\ldots$, which may depend on ``fixed'' parameters such as $M_1,M_2,M_3,\alpha,\beta,d,s$ that do not vary with the sample sizes ${n_1}$ and ${n_2}$. The specific values of these constants may vary in different places. We use the notation $A_n\xrightarrow{p}A$ to denote that the sequence of the random variables $A_n$ converges in probability to a random variable $A.$ We also introduce a terminology for the convergence of permutation distributions. For a given generic test statistic $T_{{n_1},{n_2}}$ and a continuous random variable $G$, denote the permutation distribution of $T_{{n_1},{n_2}}$ as $F^\pi_{T_{{n_1},{n_2}}}(\cdot):=\frac{1}{N!}\sum_{\pi \in \Pi_{N}} \mathds 1 \{T_{{n_1},{n_2}}(Z_{\pi(1)},\dots,Z_{\pi(N)})\leq \cdot\}$ and the cumulative distribution function (CDF) of $G$ as $F_G(\cdot) := \mathbb{P}(G \leq \cdot)$. Suppose that 
	$$
	\sup_{t\in \mathbb R^d}\big|F^\pi_{T_{{n_1},{n_2}}}(t)-F_G(t) \big|\xrightarrow{p}0,
	$$
	or equivalently, for any given $\epsilon>0,$
	$$
	\lim_{{n_1},{n_2}\rightarrow\infty}\mathbb P \bigg(\sup_{t\in \mathbb R^d}\big|F^\pi_{T_{{n_1},{n_2}}}(t)-F_G(t) \big|>\epsilon\bigg)=0.
	$$
	In this case, we say that $F^\pi_{T_{{n_1},{n_2}}}$ \textit{converges weakly in probability} to $G$, as in \citet{chung2016multivariate}. Also, if a sequence of random variables $\{H_n\}_{n=1}^\infty$ converges in distribution to a continuous random variable $H$, we use the expression that aligns with the above:
	$$
	\sup_{t\in \mathbb R^d}\big|F_{H_n}(t)-F_H(t) \big|\rightarrow0,
	$$
	instead of $H_n\xrightarrow{d} H.$ Note that P\'{o}lya's theorem can be generalized into the multivariate case \citep[][Lemma C.7]{guo2024ranktransformed} and thus guarantees the equivalence between those two expressions under the assumption that $H$ is continuous.

	\subsection{Proof of Proposition \ref{proposition.kmoment}} \label{Section: proof of proposition.kmoment}
	For simplicity, we consider the case where $d=1$, as the scenario with $d \geq 2$ can be extended naturally by taking the Cartesian product of the one-dimensional cases. Also, we consider the case $k=2$, since the case $k=1$ is identical to Lemma \ref{lemma1}, and the logic used for $k=2$ can be extended to prove the cases for $k \geq 3$.
	
	Now, suppose that $k=2$. Given frequencies $\boldsymbol{\omega}_R=\{\omega_1,\ldots,\omega_R\},$ recall that the feature mapping is defined as
	$$
	\boldsymbol{\psi}_{\boldsymbol{\omega}_R}(x)=\frac{1}{\sqrt R}[\cos (\omega_1^\top x),\sin(\omega_1^\top x),~\ldots~,\cos (\omega_R^\top x),\sin(\omega_R^\top x)]^\top \in \mathbb R^{2R},
	$$
	and $\mathcal E_2$ can be written as 
	\begin{equation*}
		\mathcal E_2:= \bigl\{ \boldsymbol{x} \in \mathbb{R}^{d \times R}:  \mathbb E_{X}[\boldsymbol{\psi}_{\boldsymbol{x}} (X)]=\mathbb E_{Y}[\boldsymbol{\psi}_{\boldsymbol{x}}(Y)],~\mathbb E_{X}[\boldsymbol{\psi}_{\boldsymbol{x}} (X)\boldsymbol{\psi}_{\boldsymbol{x}} (X)^\top ]=\mathbb E_{Y}[\boldsymbol{\psi}_{\boldsymbol{x}}(Y)\boldsymbol{\psi}_{\boldsymbol{x}}(Y)^\top]\bigr\}.
	\end{equation*}
	Then, the components of the second moment matrix of $\boldsymbol{\psi}_{\boldsymbol{\omega}_R}(X)$ should be one of the following terms: 
	\begin{align*}
		\mathbb E_{X}[\cos (\omega_i^\top X)\cos (\omega_j^\top X)],&\quad i,j=1,\ldots,R, \quad \text{or}\\
		\mathbb E_{X}[\cos (\omega_i^\top X)\sin (\omega_j^\top X)],&\quad i,j=1,\ldots,R, \quad \text{or}\\
		\mathbb E_{X}[\sin (\omega_i^\top X)\sin (\omega_j^\top X)],&\quad i,j=1,\ldots,R.
	\end{align*}
	Similarly, the same argument holds true for $\boldsymbol{\psi}_{\boldsymbol{\omega}_R}(Y)$. Hence, to show that $\mathbb E_{X}[\boldsymbol{\psi}_{\boldsymbol{\omega}_R}(X)\boldsymbol{\psi}_{\boldsymbol{\omega}_R}(X)^\top]=\mathbb E_{Y}[\boldsymbol{\psi}_{\boldsymbol{x}}(Y)\boldsymbol{\psi}_{\boldsymbol{x}}(Y)^\top]$ holds, it is enough to show that the following identities are satisfied simultaneously:
	\begin{align*}
		\mathbb E_{X}[\cos (\omega_i^\top X)\cos (\omega_j^\top X)]&=\mathbb E_{Y}[\cos (\omega_i^\top Y)\cos (\omega_j^\top Y)],\\
		\mathbb E_{X}[\cos (\omega_i^\top X)\sin (\omega_j^\top X)]&=\mathbb E_{Y}[\cos (\omega_i^\top Y)\sin (\omega_j^\top Y)],\\
		\mathbb E_{X}[\sin (\omega_i^\top X)\sin (\omega_j^\top X)]&=\mathbb E_{Y}[\sin (\omega_i^\top Y)\sin (\omega_j^\top Y)],
	\end{align*}
	for all $i,j=1,\ldots,R.$
	By trigonometric identities, these identities are equivalent to
	\begin{align*}
		\mathbb E_{X}\big[\cos \big((\omega_i-\omega_j)^\top X\big)+\cos \big((\omega_i+\omega_j)^\top X\big)\big]&=\mathbb E_{Y}\big[ \cos \big((\omega_i-\omega_j)^\top Y\big)+\cos \big((\omega_i+\omega_j)^\top Y\big)\big],\\
		\mathbb E_{X}\big[\sin \big((\omega_i+\omega_j)^\top X\big)-\sin \big((\omega_i-\omega_j)^\top X\big)\big]&=\mathbb E_{Y}\big[\sin \big((\omega_i+\omega_j)^\top Y\big)-\sin \big((\omega_i-\omega_j)^\top Y\big)\big],\\
		\mathbb E_{X}\big[\cos \big((\omega_i-\omega_j)^\top X\big)-\cos \big((\omega_i+\omega_j)^\top X\big)\big]&=\mathbb E_{Y}\big[\cos \big((\omega_i-\omega_j)^\top Y\big)-\cos \big((\omega_i+\omega_j)^\top Y\big)\big],
	\end{align*}
	for all $i,j=1,\ldots,R.$ Hence, if we show that the following identities
	\begin{align*}
		\mathbb E_{X}\big[\cos \big((\omega_i+\omega_j)^\top X\big)\big]&=\mathbb E_{Y}\big[\cos \big((\omega_i+\omega_j)^\top Y\big)\big],\\
		\mathbb E_{X}\big[\sin \big((\omega_i+\omega_j)^\top X\big)\big]&=\mathbb E_{Y}\big[\sin \big((\omega_i+\omega_j)^\top Y\big)\big],\\
		\mathbb E_{X}\big[\cos \big((\omega_i-\omega_j)^\top X\big)\big]&=\mathbb E_{Y}\big[\cos \big((\omega_i-\omega_j)^\top Y\big)\big],\\
		\mathbb E_{X}\big[\sin \big((\omega_i-\omega_j)^\top X\big)\big]&=\mathbb E_{Y}\big[\sin \big((\omega_i-\omega_j)^\top Y\big)\big]
	\end{align*}
	hold for all $i,j=1,\ldots,R$ given $\boldsymbol{\omega}_R,$ the second moment matrices of $\boldsymbol{\psi}_{\boldsymbol{\omega}_R}(X)$ and $\boldsymbol{\psi}_{\boldsymbol{\omega}_R}(Y)$ become identical. Also, note that satisfying the first two identities is equivalent to the coincidence of characteristic functions $\phi_X(\omega)$ and $\phi_Y(\omega)$ at the point $\omega_i+\omega_j,$ and the last two identities imply the coincidence of $\phi_X(\omega)$ and $\phi_Y(\omega)$ at the point $\omega_i-\omega_j.$
	Let us denote the collection of frequencies that need to be controlled by
	$$\mathcal W_R:=\{\omega_i:1\le i\le R\}\cup\{\omega_i+\omega_j:1\le i,j\le R\}\cup\{\omega_i-\omega_j:1\le i\neq j\le R\},
	$$
	and then, a sufficient condition for $\boldsymbol{\omega}_R\in \mathcal E_2$ is $\phi_X(W_r)=\phi_Y(W_r)$ for all $W_r\in \mathcal W_R.$ Note that the cardinality of $\mathcal W_R$ is $|\mathcal W_R|\leq 2R^2.$ The cases $\omega_i-\omega_i=0$ are omitted since
$\phi_X(0)=\phi_Y(0)=1$ holds automatically. Now, observe that all random variables $W_r\in \mathcal W_R$ have continuous probability distributions, and thus, for any $\epsilon>0$, we can find $I=I(\epsilon)>0$ satisfying $\mathbb P(W_r \in [-I,I])\leq (2R^2)^{-1}\epsilon$ for $W_r\in \mathcal W_R.$ Then,
	\begin{align*}
    \mathbb P\big(W\in[-I,I]^c\ \text{for all }W\in\mathcal W_R\big)
    &\geq 1-\sum_{W\in\mathcal W_R}\mathbb P(W\in[-I,I])\\
    &\geq 1-\epsilon.
    \end{align*}
	Here, according to P\'{o}lya's criterion \citep[][Theorem 1]{polya1949remarks}, we can find uncountably many characteristic functions that vanish outside the interval $[-I,I].$ Let $\mathcal A_{k,\epsilon}$ denote the corresponding family of
    probability distributions. A representative example of these characteristic functions~\citep[see e.g.,][Proposition 1]{chwialkowski2015fast} is a set $\{f_\delta\}_{\delta>I^{-1}}$ where
	\begin{align*}
		f_\delta(\omega)=\left\{\begin{array}{lll}
			1-\delta|\omega| & \text { when } & |\omega| \leq \frac1\delta, \\[.5em]
			0 & \text { when } & |\omega| \geq \frac1\delta.
		\end{array}\right.
	\end{align*}
	Then, for any distribution pair $P_X,P_Y\in\mathcal A_{k,\epsilon},$ we have
	\begin{align*}
		\mathbb{P}_{\boldsymbol{\omega}_R}(\boldsymbol{\omega}_R\in \mathcal E_2)&\geq 1-\sum_{W\in\mathcal W_R}\mathbb P(W\in[-I,I])\\
    &\geq 1-\epsilon,
	\end{align*}
	and this completes the proof.

	\subsection{Proof of Theorem \ref{negthm}} \label{Section: proof of negthm}
	
	Recall that we use a permutation test defined as follows:
	\begin{equation*}
		\Delta_{{n_1},{n_2},R}^{\alpha} := \mathds{1}(V > q_{{n_1},{n_2},1-\alpha}).
	\end{equation*}
	We first note that the permutation test is invariant under multiplying a positive constant ${n_1}$ to the test statistic. Therefore, throughout the proof of Theorem \ref{negthm}, we consider ${n_1}V$ as a test statistic and its permutation quantile ${n_1}q_{{n_1},{n_2},1-\alpha}$ instead of $V$ and $q_{{n_1},{n_2},1-\alpha}$, respectively.
	Now, note that both the test statistic ${n_1}V$ and the permutation quantile ${n_1}q_{{n_1},{n_2},1-\alpha}$ are random variables. Our strategy to prove Theorem \ref{negthm} is to first assume the ME condition~\eqref{meaneq}, $\boldsymbol{\omega}_R \in \mathcal{E}$, which holds for any distribution pair $P_X,P_Y \in \mathcal A_\epsilon$ with high probability by Lemma~\ref{lemma1}. 
	For such fixed $\boldsymbol{\omega}_R$, we analyze the asymptotic behavior of ${n_1}V$ and show that the test power is strictly smaller than one for an uncountable number of pairs of distributions $P_X$ and $P_Y$. 
	
	\subsubsection*{Asymptotic behavior of the unconditional distribution}
	Let us start by investigating the test statistic ${n_1}V$ with a fixed $\boldsymbol{\omega}_R \in \mathcal{E}$. For the sake of notation, we denote the covariances of $\boldsymbol{\psi}_{\boldsymbol{\omega}_R}(X)$ and $\boldsymbol{\psi}_{\boldsymbol{\omega}_R}(Y)$ as $\Sigma_{\boldsymbol{\psi}_{\boldsymbol{\omega}_R}(X)}$ and $\Sigma_{\boldsymbol{\psi}_{\boldsymbol{\omega}_R}(Y)}$, respectively. Note that $\boldsymbol{\psi}_{\boldsymbol{\omega}_R}(X)$ and $\boldsymbol{\psi}_{\boldsymbol{\omega}_R}(Y)$ are trigonometric functions, having finite variance. Hence the central limit theorem guarantees that the unconditional distribution of ${n_1}^{1/2}T$ converges in distribution to $N(0,\tilde \Sigma)$ where $\tilde \Sigma=\Sigma_{\boldsymbol{\psi}_{\boldsymbol{\omega}_R}(X)}+\frac {p}{1-p}\Sigma_{\boldsymbol{\psi}_{\boldsymbol{\omega}_R}(Y)} \in \mathbb R^{2R\times 2R}$. Letting $\tilde k=\operatorname{rank}(\tilde \Sigma)$, consider an eigendecomposition of $\tilde{\Sigma}$:
	$$
	\tilde \Sigma= \tilde Q\tilde D \tilde Q^\top.
	$$
    Here, $\tilde D=\mathrm{diag}(\tilde \lambda_1,\ldots,\tilde \lambda_{\tilde k}) \in \mathbb R^{\tilde k\times \tilde k}$ is a diagonal matrix formed from the non-zero eigenvalues of $\tilde \Sigma$, and $\tilde Q \in \mathbb R^{2R\times \tilde k}$ is an orthogonal matrix with columns corresponding to the eigenvectors of $\tilde\Sigma$. Then, a Gaussian random vector $A\sim N(0,\tilde \Sigma)$ can be decomposed as $A = \tilde Q\tilde D^{\frac12}G$ where $G=(G_1,\ldots,G_{\tilde k})^\top\sim N(0,I_{\tilde k \times \tilde k})$. Therefore, the distribution of $\|A\|^2_{\mathbb R^{2R}}$ can be derived as follows:
	\begin{align*}
		\|A\|^2_{\mathbb R^{2R}}=A^\top A&=G^\top\tilde D^{\frac12}\tilde Q^\top\tilde Q\tilde D^{\frac12}G =G^\top\tilde D^{\frac12}\tilde D^{\frac12}G\\
		&=\sum^{\tilde k}_{i=1}\tilde \lambda_i G^2_i.
	\end{align*}
	Based on the fact that ${n_1}^{1/2}T$ converges in distribution to $A$, the continuous mapping theorem guarantees
	\begin{equation}\label{true.ex}
		\sup_{t\in \mathbb R}\big|F_{{n_1}V}(t)-F_{\sum\tilde\lambda_iG_i^2}(t) \big|\rightarrow0
	\end{equation}
	for fixed $\boldsymbol{\omega}_R \in \mathcal{E}$, where $F_{\sum\tilde\lambda_iG_i^2}$ is the CDF of $\sum^{\tilde k}_{i=1}\tilde\lambda_iG_i^2.$
	If $\tilde k=2R$, then the limiting distribution becomes $\sum^{2R}_{i=1}\tilde\lambda_iG_i^2$. Even when $\tilde k$ is strictly less than $2R$, we note that the distribution $F_{\sum\tilde\lambda_iG_i^2}$ can also be regarded as the distribution of $\sum^{2R}_{i=1}\tilde\lambda_iG_i^2$ instead of $\sum^{\tilde k}_{i=1}\tilde\lambda_iG_i^2,$ since we can extend the eigenvalue set $\{\tilde\lambda_i\}_{i=1}^{\tilde k}$ to $\{\tilde\lambda_i\}_{i=1}^{2R}$ by including zero eigenvalues.
	
	\subsubsection*{Asymptotic behavior of the permutation distribution}
	
	We now examine the asymptotic behavior of the permutation distribution of ${n_1}V$ and ${n_1}q_{{n_1},{n_2},1-\alpha}$ for fixed $\boldsymbol{\omega}_R \in \mathcal{E}$. First, let $\bar k=\operatorname{rank}(\bar \Sigma)$ where $\bar \Sigma = \frac {p}{1-p}\Sigma_{\boldsymbol{\psi}_{\boldsymbol{\omega}_R}(X)}+\Sigma_{\boldsymbol{\psi}_{\boldsymbol{\omega}_R}(Y)}$, and consider an eigendecomposition
	\begin{align*}
		\bar \Sigma&= Q D Q^\top=\left[\bar Q \,|\, Q_0\right]
		\left[
		\begin{array}{c|c}
			\bar D & \mathbf 0 \\
			\hline
			\mathbf 0 & \mathbf 0
		\end{array}
		\right] \left[\bar Q \,|\, Q_0\right]^\top\\
		&=\bar Q \bar D \bar Q^\top,
	\end{align*}
	where $D \in \mathbb R^{2R\times 2R}$ is a diagonal matrix composed of $\bar D$ and zeros, $\bar D=\text{diag}(\bar \lambda_1,\ldots,\bar \lambda_{\bar k} )\in \mathbb R^{\bar k\times \bar k}$ is a diagonal matrix formed from the non-zero eigenvalues of $\bar \Sigma$. In addition $\bar Q \in \mathbb R^{2R\times \bar k}$ denotes an orthogonal matrix whose columns are the eigenvectors of $\bar\Sigma$ and $Q_0 \in \mathbb R^{2R\times (2R-\bar k)}$ denotes an orthogonal matrix that makes $Q=\left[\bar Q \,|\, Q_0 \right] \in \mathbb R^{2R\times 2R}$ also orthogonal. Note that ${n_1}V$ can be decomposed as 
	\begin{equation}\label{Tdecompose}
		\begin{aligned}
			{n_1}V&={n_1}T^\top T\\
			&={n_1}T^\top QQ^\top T\\
			&={n_1}T^{\top}\bar Q\bar Q^{\top}T+{n_1}T^{\top}Q_0Q_0^{\top}T.
		\end{aligned}
	\end{equation}

	For the first term, note that $\bar Q^{\top}T=\frac{1}{n_1}\sum_{i=1}^{n_1}\bar Q^{\top}\boldsymbol{\psi}_{\boldsymbol{\omega}_R}(X_i)-\frac1{{n_2}}\sum_{j=1}^{n_2}\bar Q^{\top}\boldsymbol{\psi}_{\boldsymbol{\omega}_R}(Y_j)$ and $\mathbb E_{X}[\bar Q^{\top}\boldsymbol{\psi}_{\boldsymbol{\omega}_R} (X)]=\mathbb E_{Y}[\bar Q^{\top}\boldsymbol{\psi}_{\boldsymbol{\omega}_R} (Y)]$ for $\boldsymbol{\omega}_R \in \mathcal{E}$. Furthermore, we observe that
	\begin{align*}
		\frac {p}{1-p}\Sigma_{\bar Q^\top\boldsymbol{\psi}_{\boldsymbol{\omega}_R}(X)}+\Sigma_{\bar Q^\top\boldsymbol{\psi}_{\boldsymbol{\omega}_R}(Y)}&=\frac {p}{1-p}\bar Q^{\top}\Sigma_{\boldsymbol{\psi}_{\boldsymbol{\omega}_R}(X)}\bar Q+\bar Q^{\top}\Sigma_{\boldsymbol{\psi}_{\boldsymbol{\omega}_R}(Y)}\bar Q\\
		&=\bar Q^{\top}\Big(\frac {p}{1-p}\Sigma_{\boldsymbol{\psi}_{\boldsymbol{\omega}_R}(X)}+\Sigma_{\boldsymbol{\psi}_{\boldsymbol{\omega}_R}(Y)}\Big)\bar Q\\
		&=\bar Q^{\top}\bar \Sigma\bar Q\\
		&=\bar D.
	\end{align*}
	Since $\bar D$ is positive definite, we apply Lemma \ref{lemma2} to ${n_1}^{1/2}\bar Q^{\top}T$ and conclude that the permutation distribution of ${n_1}^{1/2}\bar Q^{\top}T$ converges weakly in probability to $N(0,\bar D)$ for fixed $\boldsymbol{\omega}_R \in \mathcal{E}$. Therefore, the continuous mapping theorem for permutation distribution \citep[Lemma A.6]{chung2016multivariate} guarantees that the permutation distribution of ${n_1}T^{\top}\bar Q\bar Q^{\top}T$ converges weakly in probability to $\sum^{\bar k}_{i=1}\bar\lambda_iG_i^2$ for $\boldsymbol{\omega}_R \in \mathcal{E}$. More formally, if we denote the permutation distribution of ${n_1}T^{\top}\bar Q\bar Q^{\top}T$ as $F^\pi_{({n_1}^{1/2}\bar Q^{\top}T)^2}(t)$, for any $\epsilon>0,$ we have
	\begin{equation*}
		\lim_{{n_1},{n_2}\rightarrow\infty}\mathbb P\Big(\sup_{t\in \mathbb R}\big|F^\pi_{({n_1}^{1/2}\bar Q^\top T)^2}(t)-F_{\sum\bar\lambda_iG_i^2}(t) \big|>\epsilon \,\Big|\,\boldsymbol{\omega}_R=\boldsymbol{\omega}\Big)=0
	\end{equation*}
	for each fixed $\boldsymbol{\omega} \in \mathcal{E}$, where $F_{\sum\bar\lambda_iG_i^2}$ denotes the CDF of $ \sum^{\bar k}_{i=1}\bar\lambda_iG_i^2.$

	For the second term in Equation \eqref{Tdecompose}, we note that the null space of the sum of two positive semidefinite matrices is the intersection of the null spaces of each of them. Since $Q_0$ spans the null space of $\bar \Sigma=\frac {p}{1-p}\Sigma_{\boldsymbol{\psi}_{\boldsymbol{\omega}_R}(X)}+\Sigma_{\boldsymbol{\psi}_{\boldsymbol{\omega}_R}(Y)}$ and $\bar \Sigma$ is positive semidefinite, the columns of $Q_0$ are in the null space of $\Sigma_{\boldsymbol{\psi}_{\boldsymbol{\omega}_R}(X)}$ and $\Sigma_{\boldsymbol{\psi}_{\boldsymbol{\omega}_R}(Y)}$. This implies that $Q_0^{\top}\boldsymbol{\psi}_{\boldsymbol{\omega}_R}(X)=Q_0^{\top}\mathbb E\big[\boldsymbol{\psi}_{\boldsymbol{\omega}_R}(X)\big]$ and $Q_0^{\top}\boldsymbol{\psi}_{\boldsymbol{\omega}_R}(Y)=Q_0^{\top}\mathbb E\big[\boldsymbol{\psi}_{\boldsymbol{\omega}_R}(Y)\big]$ almost surely. Hence, for any permutation $\pi \in \Pi_N$, we have
	$$
	Q_0^{\top} T(Z_{\pi(1)},\ldots,Z_{\pi(N)})=\frac{1}{n_1}\sum_{i=1}^{n_1}Q_0^{\top}\boldsymbol{\psi}_{\boldsymbol{\omega}_R}(Z_{\pi(i)})-\frac1{{n_2}}\sum_{j={n_1}+1}^NQ_0^{\top}\boldsymbol{\psi}_{\boldsymbol{\omega}_R}(Z_{\pi(j)})=\mathbf 0,
	$$
	from ME condition, and thus we conclude that the permutation distribution of ${n_1}T^{\top}Q_0Q_0^{\top}T$ is degenerate at zero.
	Combining the results, Equation \eqref{Tdecompose} implies $F^\pi_{{n_1}V}(t)=F^\pi_{({n_1}^{1/2}\bar Q^\top T)^2}(t)$, thus for any $\epsilon>0,$
	\begin{equation*}
		\lim_{{n_1},{n_2}\rightarrow\infty}\mathbb P\Big(\sup_{t\in \mathbb R}\big|F^\pi_{{n_1}V}(t)-F_{\sum\bar\lambda_iG_i^2}(t) \big|>\epsilon \,\Big|\,\boldsymbol{\omega}_R=\boldsymbol{\omega}\Big)=0,
	\end{equation*}
	for each fixed $\boldsymbol{\omega} \in \mathcal{E}$.
	Similar to Equation \eqref{true.ex}, we can include zero eigenvalues and $F_{\sum\bar\lambda_iG_i^2}$ can be seen as the distribution of $ \sum^{2R}_{i=1}\bar\lambda_iG_i^2$, instead of $ \sum^{\bar k}_{i=1}\bar\lambda_iG_i^2.$
	
	When a sequence of ``random'' distribution functions converges weakly in probability to a fixed distribution function, it ensures convergence in its quantile \citep[][Lemma 11.2.1 (ii)]{lehmann2006testing}. Hence, the critical value ${n_1}q_{{n_1},{n_2},1-\alpha}$ of ${n_1}V$ converges in probability to $ q_{R,1-\alpha}$, where $q_{R,1-\alpha}$ is the $(1-\alpha)$-quantile of $\sum^{2R}_{i=1}\bar\lambda_iG_i^2$ under the ME condition. In other words, for any $\epsilon>0,$
	\begin{equation}\label{cdist.appendix}
		\lim_{{n_1},{n_2}\rightarrow\infty}\mathbb P\big(|{n_1}q_{{n_1},{n_2},1-\alpha}-q_{R,1-\alpha} |>\epsilon \,\big|\,\boldsymbol{\omega}_R=\boldsymbol{\omega}\big)=0
	\end{equation}
	for each fixed $\boldsymbol{\omega} \in \mathcal{E}$.

	\subsubsection*{Constructing $P_X$ and $P_Y$}
	We start by summarizing the analysis we have done so far. For the permutation test $\Delta_{{n_1},{n_2},R}^{\alpha} := \mathds{1}({n_1}V > {n_1}q_{{n_1},{n_2},1-\alpha})$ with $\boldsymbol{\omega}_R \in {\mathcal E}$, which implies that the 1-ME condition holds, we have shown that the unconditional distribution of the test statistic ${n_1}V$, denoted as $F_{{n_1}V}$, converges in distribution to $\sum^{2R}_{i=1}\tilde\lambda_iG_i^2$ and the critical value ${n_1}q_{{n_1},{n_2},1-\alpha}$ of permutation distribution $F^\pi_{{n_1}V}$ converges in probability to $ q_{R,1-\alpha}$, that is, the $(1-\alpha)$-quantile of $\sum^{2R}_{i=1}\bar\lambda_iG_i^2$. Since $\{\tilde\lambda_i\}_{i=1}^{2R}$ and $\{\bar\lambda_i\}_{i=1}^{2R}$ are the eigenvalues of $\tilde \Sigma$ and $\bar \Sigma$, the asymptotic power depends on the difference between these matrices. Note that they are given as 
	$$
	\tilde \Sigma=\Sigma_{\boldsymbol{\psi}_{\boldsymbol{\omega}_R}(X)}+\frac {p}{1-p}\Sigma_{\boldsymbol{\psi}_{\boldsymbol{\omega}_R}(Y)} \quad \text{and} \quad \bar \Sigma=\frac {p}{1-p}\Sigma_{\boldsymbol{\psi}_{\boldsymbol{\omega}_R}(X)}+\Sigma_{\boldsymbol{\psi}_{\boldsymbol{\omega}_R}(Y)},
	$$
	which involves the second moments of the feature mappings.
	
	Here, given $\boldsymbol{\omega}_R\in\mathcal E$, suppose that the matrices $\tilde \Sigma$ and $\bar \Sigma$ coincide for some distribution pair $P_X, P_Y$. This implies that the permutation distribution $F^\pi_{{n_1}V}$ and the unconditional distribution $F_{{n_1}V}$ become asymptotically identical for such fixed $\boldsymbol{\omega}_R$, and thus we can expect that the test fails to distinguish those distributions $P_X$ and $P_Y$ in this case. To formalize this scenario, consider an extension of the 1-ME condition to include second moments. Recall that the 1-ME condition is $\boldsymbol{\omega}_R \in {\mathcal E},$ meaning that the first moments of the feature mappings are identical. Now, consider the 2-ME condition, $\boldsymbol{\omega}_R \in {\mathcal E_2},$ which implies the coincidence up to the second moments of the feature mappings. i.e., 
	\begin{equation*}
		\mathcal E_2:= \bigl\{ \boldsymbol{x} \in \mathbb{R}^{d \times R}:  \mathbb E_{X}[\boldsymbol{\psi}_{\boldsymbol{x}} (X)]=\mathbb E_{Y}[\boldsymbol{\psi}_{\boldsymbol{x}}(Y)],~\mathbb E_{X}[\boldsymbol{\psi}_{\boldsymbol{x}} (X)\boldsymbol{\psi}_{\boldsymbol{x}} (X)^\top ]=\mathbb E_{Y}[\boldsymbol{\psi}_{\boldsymbol{x}}(Y)\boldsymbol{\psi}_{\boldsymbol{x}}(Y)^\top]\bigr\}.
	\end{equation*} 
	Then, note that Proposition \ref{proposition.kmoment} guarantees that the 2-ME condition holds for any distribution pair $P_X,P_Y \in \mathcal A_{k,\epsilon}$ with arbitrarily high probability $1-\epsilon$. Also, observe that if $\boldsymbol{\omega}_R \in {\mathcal E_2},$ then the covariance matrices of the feature mapping $\Sigma_{\boldsymbol{\psi}_{\boldsymbol{\omega}_R}(X)}$ and $\Sigma_{\boldsymbol{\psi}_{\boldsymbol{\omega}_R}(Y)}$ are the same, and this indicates that the matrices $\tilde \Sigma$ and $\bar \Sigma$ coincide. We again emphasize that $F^\pi_{{n_1}V}$ and $F_{{n_1}V}$ become asymptotically identical  in this case.  Therefore, for such $\boldsymbol{\omega}_R \in {\mathcal E_2},$ the critical value ${n_1}q_{{n_1},{n_2},1-\alpha}$ converges in probability to the $(1-\alpha)$-quantile of the unconditional distribution, and combining this fact with Equation \eqref{true.ex} and Equation \eqref{cdist.appendix}, Slutsky's theorem yields the convergence
	\begin{equation}\label{asconv}
		\lim_{{n_1},{n_2}\rightarrow\infty}\mathbb P({n_1}V\geq {n_1}q_{{n_1},{n_2},1-\alpha}\,|\, \boldsymbol{\omega}_R = \boldsymbol{\omega}) = \alpha,
	\end{equation}
	for fixed $\boldsymbol{\omega} \in {\mathcal E_2}$.
	For a given $\epsilon \in (0,1)$, let $P_X$ and $P_Y$ be distinct distributions in $\mathcal A_{k,\epsilon}$ defined in Proposition \ref{proposition.kmoment}. Then, we obtain the following result: 
	\begin{align*}
		\mathbb{P}(\Delta_{{n_1},{n_2},R}^{\alpha} = 1) & = \int_{\mathcal E_2}\mathbb{P}(\Delta_{{n_1},{n_2},R}^{\alpha} = 1 \,|\, \boldsymbol{\omega}_R = \boldsymbol{\omega}) f_{\boldsymbol{\omega}_R}(\boldsymbol{\omega}) d\boldsymbol{\omega} + \int_{\mathcal E_2^c}\mathbb{P}(\Delta_{{n_1},{n_2},R}^{\alpha} = 1 \,|\, \boldsymbol{\omega}_R = \boldsymbol{\omega}) f_{\boldsymbol{\omega}_R}(\boldsymbol{\omega}) d\boldsymbol{\omega} \\
		& \leq \int_{\mathcal E_2}\mathbb{P}(\Delta_{{n_1},{n_2},R}^{\alpha} = 1 \,|\, \boldsymbol{\omega}_R = \boldsymbol{\omega}) f_{\boldsymbol{\omega}_R}(\boldsymbol{\omega}) d\boldsymbol{\omega} + \epsilon
	\end{align*}
	Since the probability is bounded by 1, the dominated convergence theorem guarantees
	\begin{align*}
	\lim_{{n_1},{n_2} \rightarrow \infty} \int_{\mathcal E_2}\mathbb{P}(\Delta_{{n_1},{n_2},R}^{\alpha} = 1 \,|\, \boldsymbol{\omega}_R = \boldsymbol{\omega}) f_{\boldsymbol{\omega}_R}(\boldsymbol{\omega}) d\boldsymbol{\omega} &= \int_{\mathcal E_2} \lim_{{n_1},{n_2} \rightarrow \infty}  \mathbb{P}(\Delta_{{n_1},{n_2},R}^{\alpha} = 1 \,|\, \boldsymbol{\omega}_R = \boldsymbol{\omega}) f_{\boldsymbol{\omega}_R}(\boldsymbol{\omega}) d\boldsymbol{\omega}\\
	&\leq\alpha,
	\end{align*}
	where the last inequality follows from \Cref{asconv}.
	Therefore, we get the desired result:
	\begin{align*}
		\limsup_{{n_1},{n_2} \rightarrow \infty} \mathbb{P}(\Delta_{{n_1},{n_2},R}^{\alpha} = 1) & \leq \lim_{{n_1},{n_2} \rightarrow \infty} \int_{\mathcal E_2}\mathbb{P}(\Delta_{{n_1},{n_2},R}^{\alpha} = 1 \,|\, \boldsymbol{\omega}_R = \boldsymbol{\omega}) f_{\boldsymbol{\omega}_R}(\boldsymbol{\omega}) d\boldsymbol{\omega} + \epsilon\\
		& \leq \alpha + \epsilon.
	\end{align*}

	\subsection{Proof of Corollary \ref{negcor}} \label{Section: proof of negcor}
	As shown by \citet[Appendix A.1]{zhao2015fastmmd}, the unbiased estimator of MMD can be written as
	\begin{align*}
		\widehat {\mathrm{MMD}}_u^2( \mathcal{X}_{n_1}, \mathcal{Y}_{n_2};\mathcal H_k)&=\widehat {\mathrm{MMD}}_b^2( \mathcal{X}_{n_1}, \mathcal{Y}_{n_2};\mathcal H_k)+\frac{1}{{n_1}-1}\sum^{n_1}_{i=1}\sum^{n_1}_{j=1}\frac{1}{{n_1^2}}k(X_i,X_j)+\frac{1}{{n_2}-1}\sum^{n_2}_{i=1}\sum^{n_2}_{j=1}\frac{1}{{n_2^2}}k(Y_i,Y_j)\\&\quad-\kappa(0)\bigg(\frac1{{n_1}-1}+\frac1{{n_2}-1}\bigg),
	\end{align*}
	where $k(x,y)=\kappa(x-y)$. By replacing the kernel $k(x,y)$ with $\hat k(x,y)=\langle\boldsymbol{\psi}_{\boldsymbol{\omega}_R}(x),\boldsymbol{\psi}_{\boldsymbol{\omega}_R}(y) \rangle$, we get
	\begin{equation}    \label{corproof}
		\begin{aligned}
			\text{r}\widehat {\mathrm{MMD}}_u^2( \mathcal{X}_{n_1}, \mathcal{Y}_{n_2};\boldsymbol{\omega}_R)&=\text{r}\widehat {\mathrm{MMD}}_b^2( \mathcal{X}_{n_1}, \mathcal{Y}_{n_2};\boldsymbol{\omega}_R)+\frac{1}{{n_1}-1}\sum^{n_1}_{i=1}\sum^{n_1}_{j=1}\frac{1}{{n_1^2}}\hat k(X_i,X_j)\\&\quad+\frac{1}{{n_2}-1}\sum^{n_2}_{i=1}\sum^{n_2}_{j=1}\frac{1}{{n_2^2}}\hat k(Y_i,Y_j)-\kappa(0)\bigg(\frac1{{n_1}-1}+\frac1{{n_2}-1}\bigg)\\
			&=\text{r}\widehat {\mathrm{MMD}}_b^2( \mathcal{X}_{n_1}, \mathcal{Y}_{n_2};\boldsymbol{\omega}_R)+\frac{1}{{n_1}-1}\bigg\| \frac{1}{n_1}\sum^{n_1}_{i=1}\boldsymbol{\psi}_{\boldsymbol{\omega}_R}(X_i)\bigg\|_{\mathbb R^{2R}}^2\\&\quad+\frac{1}{{n_2}-1}\bigg\| \frac1{{n_2}}\sum^{n_2}_{i=1}\boldsymbol{\psi}_{\boldsymbol{\omega}_R}(Y_i)\bigg\|_{\mathbb R^{2R}}^2-\kappa(0)\bigg(\frac1{{n_1}-1}+\frac1{{n_2}-1}\bigg).
		\end{aligned}    
	\end{equation}
	
	Recall that we use the test statistics given as ${n_1}V={n_1}\cdot \text{r}\widehat {\mathrm{MMD}}_b^2( \mathcal{X}_{n_1}, \mathcal{Y}_{n_2};\boldsymbol{\omega}_R)$ and ${n_1}U={n_1}\cdot \text{r}\widehat {\mathrm{MMD}}_u^2( \mathcal{X}_{n_1}, \mathcal{Y}_{n_2};\boldsymbol{\omega}_R)$ in the permutation tests defined in Theorem \ref{negthm} and Corollary \ref{negcor}, respectively. Also, throughout the proof of Corollary \ref{negcor}, we assume $\kappa(0)=1$, which can be done without loss of generality as mentioned earlier in \Cref{Section: Random Fourier Features}. Then, multiplying ${n_1}$ on both sides of Equation \eqref{corproof}, we get
	\begin{equation}\label{umdiff}
		{n_1}U={n_1}V+\frac{{n_1}}{{n_1}-1}\bigg\| \frac{1}{n_1}\sum^{n_1}_{i=1}\boldsymbol{\psi}_{\boldsymbol{\omega}_R}(X_i)\bigg\|_{\mathbb R^{2R}}^2+\frac{{n_1}}{{n_2}-1}\bigg\| \frac1{{n_2}}\sum^{n_2}_{i=1}\boldsymbol{\psi}_{\boldsymbol{\omega}_R}(Y_i)\bigg\|_{\mathbb R^{2R}}^2-\frac {n_1}{{n_1}-1}-\frac{{n_1}}{{n_2}-1}.
	\end{equation}
	Our aim is to show that the unconditional distribution of the second and third terms and their permutation distribution are asymptotically the same under the ME condition. To start with, recall that the ME condition is $\boldsymbol{\omega}_R\in \mathcal E,$ implying $\mathbb E[\boldsymbol{\psi}_{\boldsymbol{\omega}_R}(X)]=\mathbb E[\boldsymbol{\psi}_{\boldsymbol{\omega}_R}(Y)].$ Let us denote the exact value of this expectation as $\mu_{\boldsymbol{\omega}_R}$ for such fixed $\boldsymbol{\omega}_R\in \mathcal E$. 
	Then, as $\frac{1}{n_1}\sum^{n_1}_{i=1}\boldsymbol{\psi}_{\boldsymbol{\omega}_R}(X_i)$ and $\frac1{{n_2}}\sum^{n_2}_{i=1}\boldsymbol{\psi}_{\boldsymbol{\omega}_R}(Y_i)$ are the sample means of $\boldsymbol{\psi}_{\boldsymbol{\omega}_R}(X_i)$ and $\boldsymbol{\psi}_{\boldsymbol{\omega}_R}(Y_i)$ with finite variance, the law of large numbers ensures that
	$$
	\frac{1}{n_1}\sum^{n_1}_{i=1}\boldsymbol{\psi}_{\boldsymbol{\omega}_R}(X_i)\xrightarrow{p} \mu_{\boldsymbol{\omega}_R} \quad \text{and} \quad  \frac1{{n_2}}\sum^{n_2}_{i=1}\boldsymbol{\psi}_{\boldsymbol{\omega}_R}(Y_i)\xrightarrow{p} \mu_{\boldsymbol{\omega}_R}.
	$$
	Therefore, by applying the continuous mapping theorem and Slutsky's theorem, it can be shown that 
	
	\begin{equation}\label{cor4.1}
		\sup_{t\in \mathbb R}\big|F_{{n_1}U}(t)-L_{{n_1}V+c}(t) \big|\rightarrow0,
	\end{equation}
	where $F_{{n_1}U}(t)$ denotes the unconditional distribution of ${n_1}U$ and $L_{{n_1}V+c}$ denotes the asymptotic unconditional distribution of ${n_1}V+c(p,\mu_{\boldsymbol{\omega}_R}):={n_1}V+\big(1+\frac{p}{1-p}\big)\|\mu_{\boldsymbol{\omega}_R}\|_{\mathbb R^{2R}}^2-1-\frac{p}{1-p}$. Note that we derived the asymptotic unconditional distribution of ${n_1}V$ in Equation \eqref{true.ex}.
	
	For the case of the permutation distribution, we reformulate the $2R$-dimensional vectors as follows:
	$$
	\Psi^{(i)}_N(Z_1,\dots,Z_N):= \sum^N_{i=1}\nu_i\boldsymbol{\psi}_{\boldsymbol{\omega}_R}(Z_i),\quad \Psi^{(ii)}_N(Z_1,\dots,Z_N):=\sum^N_{i=1}(1-\nu_i)\boldsymbol{\psi}_{\boldsymbol{\omega}_R}(Z_i),
	$$
	where $\nu_i = \mathds{1}(i \leq {n_1})$. Then, we observe that $\frac{1}{n_1}\sum^{n_1}_{i=1}\boldsymbol{\psi}_{\boldsymbol{\omega}_R}(X_i)$ and $\frac1{{n_2}}\sum^{n_2}_{i=1}\boldsymbol{\psi}_{\boldsymbol{\omega}_R}(Y_i)$ in Equation \eqref{umdiff} can be written as $\frac{1}{n_1}\Psi^{(i)}_N(Z_1,\dots,Z_N)$ and $\frac1{{n_2}}\Psi^{(ii)}_N(Z_1,\dots,Z_N).$ Let $F^\pi_{X}$ denote the permutation distribution of $\frac{1}{n_1}\Psi^{(i)}_N(Z_1,\dots,Z_N)$, defined by
	$$
	F^\pi_{X}(t):=\frac1{N!}\sum_{\pi\in \Pi_N} \mathds 1 \bigg\{\frac{1}{n_1}\Psi^{(i)}_N(Z_{\pi(1)},\dots,Z_{\pi(N)})\leq t \bigg\},
	$$
	and let $F^\pi_{Y}$ denote the permutation distribution of $\frac1{{n_2}}\Psi^{(ii)}_N(Z_1,\dots,Z_N)$ defined similarly. Let $G$ be a random variable that is uniformly distributed over $\Pi_N$.
	If we can show $\frac{1}{n_1}\Psi^{(i)}_N(Z_{G(1)},\dots,Z_{G(N)})\xrightarrow{p}\mu_{\boldsymbol{\omega}_R}$ and $\frac1{{n_2}}\Psi^{(ii)}_N(Z_{G(1)},\dots,Z_{G(N)})\xrightarrow{p}\mu_{\boldsymbol{\omega}_R}$, then the desired results $F^\pi_{X} \xrightarrow{p} \mu_{\boldsymbol{\omega}_R}$ and $F^\pi_{Y} \xrightarrow{p} \mu_{\boldsymbol{\omega}_R}$ are followed by Lemma \ref{chung.lemmaA2}. Since $R\in \mathbb N$ is a fixed number, it suffices to show that each component of $\frac{1}{n_1}\Psi^{(i)}_N(Z_{G(1)},\dots,Z_{G(N)})$ converges to the corresponding component of $\mu_{\boldsymbol{\omega}_R}$ in probability.
	
	{\allowdisplaybreaks For $1\leq k \leq 2R$, let us denote the $k$-th component of $\Psi^{(i)}_N(Z_{G(1)},\dots,Z_{G(N)})$, $\boldsymbol{\psi}_{\boldsymbol{\omega}_R}(Z_i)$ and $\mu_{\boldsymbol{\omega}_R}$ as $\Psi^{(i)}_N(Z,G)_k$, $\boldsymbol{\psi}_{\boldsymbol{\omega}_R}(Z_i)_k$ and $\mu_{R,k}$, respectively. Note that
	\begin{align*}
		\mathbb E \left[\frac{1}{n_1} \Psi^{(i)}_N(Z,G)_k\right]&=\frac{1}{n_1}\sum^N_{i=1}\,\mathbb E \left[\nu_{i}\boldsymbol{\psi}_{\boldsymbol{\omega}_R}(Z_{G(i)})_k\right]\\
		&=\frac{1}{n_1}\sum^{n_1}_{i=1}\,\mathbb E\left[\boldsymbol{\psi}_{\boldsymbol{\omega}_R}(Z_{G(i)})_k\right]\\
		&=\frac{1}{n_1}\sum^{n_1}_{i=1}\,\mathbb E_Z\Bigl[\mathbb E_G\left[\boldsymbol{\psi}_{\boldsymbol{\omega}_R}(Z_{G(i)})_k|Z_1,\ldots,Z_N\right]\Bigr]\\
		&\stackrel{(a)}{=}\frac1{n_1}\sum^{n_1}_{i=1}\mathbb E_Z\bigg[\frac1N\sum^N_{j=1}\boldsymbol{\psi}_{\boldsymbol{\omega}_R}(Z_{j})_k\bigg]\\
		&=\frac1{n_1}\sum^{n_1}_{i=1}\frac1N\big({n_1}\cdot \mathbb E[\boldsymbol{\psi}_{\boldsymbol{\omega}_R}(X)_k]+{n_2}\cdot \mathbb E[\boldsymbol{\psi}_{\boldsymbol{\omega}_R}(Y)_k] \big)\\
		&=\mu_{R,k},
	\end{align*}}
	where $(a)$ holds since $G$ is uniformly distributed over $\Pi_N$.
	
	Furthermore, we note that
	\begin{align*}
		\mathbb E\big[\boldsymbol{\psi}_{\boldsymbol{\omega}_R}(X)_k^2\big]\leq 1, \quad \mathbb E\left[\boldsymbol{\psi}_{\boldsymbol{\omega}_R}(Y)_k^2\right]\leq 1,
	\end{align*}
	and also
	\begin{align*}
		\mathbb E\big[\boldsymbol{\psi}_{\boldsymbol{\omega}_R}(Z_{G(1)})_k\boldsymbol{\psi}_{\boldsymbol{\omega}_R}(Z_{G(2)})_k\big]&=\frac{1}{N(N-1)}\sum_{1\leq i\neq j \leq N}\mathbb E\big[\boldsymbol{\psi}_{\boldsymbol{\omega}_R}(Z_i)_k\boldsymbol{\psi}_{\boldsymbol{\omega}_R}(Z_j)_k\big]\\
		&=\frac{1}{N(N-1)}\sum_{1\leq i\neq j \leq N}\mathbb E\big[\boldsymbol{\psi}_{\boldsymbol{\omega}_R}(Z_i)_k\big]\mathbb E\big[\boldsymbol{\psi}_{\boldsymbol{\omega}_R}(Z_j)_k\big]\\
		&=\frac{1}{N(N-1)}\cdot N(N-1)\mu_{R,k}^2\\&=\mu_{R,k}^2.
	\end{align*}
	Based on these observations, we have
	{\allowdisplaybreaks
	\begin{align*}
		\operatorname{Var}\bigg[\frac{1}{n_1} \Psi^{(i)}_N(Z,G)_k\bigg]&=\mathbb E\bigg[\frac{1}{n_1^2} \Psi^{(i)}_N(Z,G)_k^2\bigg]-\mu_{R,k}^2\\
		&=\frac{1}{{n_1^2}}\mathbb E\bigg[\sum^{N}_{i=1}\sum^{N}_{j=1}\nu_{i}\nu_{j}\boldsymbol{\psi}_{\boldsymbol{\omega}_R}(Z_{G(i)})_k\boldsymbol{\psi}_{\boldsymbol{\omega}_R}(Z_{G(j)})_k \bigg]-\mu_{R,k}^2\\
		&=\frac{1}{{n_1^2}}\sum^{n_1}_{i=1}\mathbb E\big[\boldsymbol{\psi}_{\boldsymbol{\omega}_R}(Z_{G(i)})_k^2\big]\\&\quad+\frac{1}{{n_1^2}}\sum_{1\leq i\neq j \leq n_1}\mathbb E\left[\boldsymbol{\psi}_{\boldsymbol{\omega}_R}(Z_{G(i)})_k\boldsymbol{\psi}_{\boldsymbol{\omega}_R}(Z_{G(j)})_k\right]-\mu_{R,k}^2\\
		&\leq\frac{1}{{n_1^2}}\cdot n_1+\frac{1}{{n_1^2}}\cdot{n_1}({n_1}-1)\cdot\mu_{R,k}^2-\mu_{R,k}^2\\
		&=\frac1{n_1}\big(1-\mu_{R,k}^2\big)\\
		&\rightarrow 0.
	\end{align*}
	}
	Therefore, we now have $\frac{1}{n_1} \Psi^{(i)}_N(Z,G)_k\xrightarrow{p}\mu_{R,k}$ for each $k$, and this implies
	$$
	\frac{1}{n_1}\Psi^{(i)}_N(Z_{G(1)},\dots,Z_{G(N)})\xrightarrow{p}\mu_{\boldsymbol{\omega}_R}.
	$$
	Similarly, we can get 
	$$
	\frac1{{n_2}}\Psi^{(ii)}_N(Z_{G(1)},\dots,Z_{G(N)})\xrightarrow{p}\mu_{\boldsymbol{\omega}_R}.
	$$
	
	For the final step, let $F_{{n_1}U}^\pi$ denote the permutation distribution function of ${n_1}U$. Then we apply the continuous mapping theorem for permutation distributions (Lemma \ref{chung.lemmaA6}) and Slutsky's theorem extended for permutation distributions (Lemma \ref{chung.lemmaA3}) to conclude that 
	\begin{equation*}
		\lim_{{n_1},{n_2}\rightarrow\infty}\mathbb P\Big(\sup_{t\in \mathbb R}\big|F^\pi_{{n_1}U}(t)-L^\pi_{{n_1}V+c}(t) \big|>\epsilon \Big)=0,
	\end{equation*}
	where $L^\pi_{{n_1}V+c}$ is the asymptotic permutation distribution of ${n_1}V+c(p,\mu_{\boldsymbol{\omega}_R})$ under the ME condition. Therefore, in the same manner as Equation \eqref{cdist.appendix}, we have 
	$$
	\lim_{{n_1},{n_2}\rightarrow\infty}\mathbb P\Big(\left|{n_1}q_{{n_1},{n_2},1-\alpha}^u-\big(q_{R,1-\alpha}+c(p,\mu_{\boldsymbol{\omega}_R})\big) \right|>\epsilon \,\Big|\,\boldsymbol{\omega}_R=\boldsymbol{\omega} \Big)=0
	$$
	for fixed $\boldsymbol{\omega}\in \mathcal E,$ where $q_{R,1-\alpha}$ denotes the $(1-\alpha)$-quantile of the distribution of $\sum^{2R}_{i=1}\bar\lambda_iG_i^2.$
	Combining the result with Equation \eqref{cor4.1}, Slutsky's theorem yields 
	\begin{align*}
		&\lim_{{n_1},{n_2}\rightarrow\infty}\mathbb P\big({n_1}U\geq {n_1}q_{{n_1},{n_2},1-\alpha}^u\,\big|\, \boldsymbol{\omega}_R = \boldsymbol{\omega}\big)
		\\=&\lim_{{n_1},{n_2}\rightarrow\infty}\mathbb P\big({n_1}V+c(p,\mu_{\boldsymbol{\omega}_R})\geq q_{R,1-\alpha}+c(p,\mu_{\boldsymbol{\omega}_R})\,\big|\, \boldsymbol{\omega}_R = \boldsymbol{\omega}\big)\\
		=&\lim_{{n_1},{n_2}\rightarrow\infty}\mathbb P\big({n_1}V+c(p,\mu_{\boldsymbol{\omega}_R})\geq {n_1}q_{{n_1},{n_2},1-\alpha}+c(p,\mu_{\boldsymbol{\omega}_R})\,\big|\, \boldsymbol{\omega}_R = \boldsymbol{\omega}\big)\\
		=&\lim_{{n_1},{n_2}\rightarrow\infty}\mathbb P\big({n_1}V\geq {n_1}q_{{n_1},{n_2},1-\alpha}\,\big|\, \boldsymbol{\omega}_R = \boldsymbol{\omega}\big)
	\end{align*}
	for fixed $\boldsymbol{\omega}\in \mathcal E.$
	Hence, the lack of consistency of the test $\Delta_{{n_1},{n_2},R}^{\alpha,u}$ follows from Theorem \ref{negthm}.
	
	\subsection{Proof of Theorem \ref{conthm}} \label{Section: proof of conthm}
	Recall that we use a permutation test defined as follows:
	\begin{align*}
		\Delta_{{n_1},{n_2},R}^{\alpha} := \mathds{1}(V > q_{{n_1},{n_2},1-\alpha}).
	\end{align*}
	For \textit{pointwise consistency}, our strategy is to find a sequence $\beta_{{n_1},{n_2},R}\rightarrow0$ such that 
	\begin{align*}
		\mathbb{P}_{X\times Y\times \omega}\big(\Delta_{{n_1},{n_2},R}^{\alpha}(\mathcal{X}_{n_1},\mathcal{Y}_{n_2})=0\big)\leq \beta_{{n_1},{n_2},R}
	\end{align*}
	is true for ${n_1},{n_2} \geq N_{(P_X,P_Y)}$ and $R\geq R_{(P_X,P_Y)}$, where $N_{(P_X,P_Y)}$ and $R_{(P_X,P_Y)}$ are constants depending on a given $(P_X,P_Y)$ with $P_X \neq P_Y$. Similarly, if we use the test $\Delta_{{n_1},{n_2},R}^{\alpha,u}$ instead of $\Delta_{{n_1},{n_2},R}^{\alpha}$, our goal is to show that $\mathbb{P}_{X\times Y\times \omega}(\Delta_{{n_1},{n_2},R}^{\alpha,u}(\mathcal{X}_{n_1},\mathcal{Y}_{n_2})=0)\leq \beta_{{n_1},{n_2},R}.$
	To achieve this goal, we use the approach that replaces a random permutation quantile with a deterministic quantity~\citep[see][]{Fromont2013,Kim2022,schrab2023mmdaggregated}.
	First, we start by integrating frameworks to analyze tests $\Delta_{{n_1},{n_2},R}^{\alpha}$ and $\Delta_{{n_1},{n_2},R}^{\alpha,u}$ in a unified manner. Let us define four events,
	\begin{align*}
		\mathcal A_V&:=\{V \leq q_{{n_1},{n_2},1-\alpha}\},\\
		\mathcal A_U&:=\{U \leq q^u_{{n_1},{n_2},1-\alpha}\},
	\end{align*}
	and
	\begin{align}
		\mathcal B_{V,\beta}:=\bigg\{\mathbb E\left[V\right] \geq \sqrt{\frac1\beta\operatorname{Var}\left[V\right]}+q_{{n_1},{n_2},1-\alpha}\bigg\},\label{eventt}\\ 
		\mathcal B_{U,\beta}:=\bigg\{\mathbb E\left[U\right] \geq \sqrt{\frac1\beta\operatorname{Var}\left[U\right]}+q^u_{{n_1},{n_2},1-\alpha}\bigg\}.\label{eventu}
	\end{align}
	Observe that $\mathbb P(\mathcal A_V)\leq\beta$ implies $\mathbb{P}_{X\times Y\times \omega}\big(\Delta_{{n_1},{n_2},R}^{\alpha}(\mathcal{X}_{n_1},\mathcal{Y}_{n_2})=0\big)\leq \beta,$
	and similarly $\mathbb P(\mathcal A_U)\leq\beta$ implies $\mathbb{P}_{X\times Y\times \omega}\big(\Delta_{{n_1},{n_2},R}^{\alpha,u}(\mathcal{X}_{n_1},\mathcal{Y}_{n_2})=0\big)\leq \beta.$ Then, for an event $\mathcal B_\beta \subseteq \mathcal B_{V,\beta}\cap\mathcal B_{U,\beta},$ we claim that $\mathbb P(\mathcal B_\beta)=1$ implies $\mathbb P(\mathcal A_V)\leq\beta$ and $\mathbb P(\mathcal A_U)\leq\beta.$ To see this, observe that Chebyshev's inequality yields
	\begin{align*}
		\mathbb P\left(\mathcal A_V,\mathcal B_\beta \right)&\leq\mathbb P\left(\mathcal A_V,\mathcal B_{V,\beta} \right)\\
		&\leq\mathbb P\bigg(V \leq \mathbb E\left[V\right]-\sqrt{\frac1\beta\operatorname{Var}\left[V\right]}\bigg) \\
		&=\mathbb P\bigg(\sqrt{\frac1\beta\operatorname{Var}\left[V\right]} \leq \mathbb E\left[V\right]-V \bigg)\\
		&\leq\mathbb P\bigg(\big| V - \mathbb E[V]\big| \geq\sqrt{\frac1\beta\operatorname{Var}[V]}\bigg)\\
		&\leq \beta.
	\end{align*}
	Then we have
	\begin{align*}
		\mathbb P\left(\mathcal A_V\right)&= \mathbb P\left(\mathcal A_V,\mathcal B_\beta \right)+ \mathbb P\left(\mathcal A_V,\mathcal B_\beta^c \right)\\
		&\leq\beta+\mathbb P\left(\mathcal A_V\,|\,\mathcal B_\beta^c \right)\mathbb P\left(\mathcal B_\beta^c \right)\\
		&=\beta+0=\beta,
	\end{align*}
	and we can get a similar result with $\mathcal A_U.$
	Therefore, our focus is on carefully identifying an event $\mathcal B_\beta$ and demonstrating that $\mathbb P(\mathcal B_\beta)=1$ for sufficiently large ${n_1},{n_2}$ and $R$.  To obtain such $\mathcal B_\beta$, we take a lower bound on the left-hand side and upper bound on the right-hand side in Equation \eqref{eventt} and Equation \eqref{eventu}. Note that, as shown in Equation \eqref{corproof}, the test statistic $V$ can be decomposed as
	\begin{equation}\label{thm5decom}
    \begin{aligned}
	V=U+W,
\end{aligned}
\end{equation}
where $$
W=\left(\frac {\kappa(0)}{n_1-1}-\frac{1}{n_1-1}\bigg\| \frac1{n_1}\sum^{n_1}_{i=1}\boldsymbol{\psi}_{\boldsymbol{\omega}_R}(X_i)\bigg\|_{\mathbb R^{2R}}^2+\frac{\kappa(0)}{n_2-1}-\frac{1}{n_2-1}\bigg\| \frac1{n_2}\sum^{n_2}_{i=1}\boldsymbol{\psi}_{\boldsymbol{\omega}_R}(Y_i)\bigg\|_{\mathbb R^{2R}}^2\right),$$
	and for all $x,y\in \mathbb R^d,$ 
	\begin{align}\label{hatkbound}
		|\langle\boldsymbol{\psi}_{\boldsymbol{\omega}_R}(x),\boldsymbol{\psi}_{\boldsymbol{\omega}_R}(y)\rangle|\leq\frac1R\sum_{r=1}^R \kappa(0)|\cos({\omega_r^\top(x-y)})|\leq \kappa(0).
	\end{align}
	Therefore, $\big\| \frac{1}{n_1}\sum^{n_1}_{i=1}\boldsymbol{\psi}_{\boldsymbol{\omega}_R}(X_i)\big\|_{\mathbb R^{2R}}^2$ and $\big\| \frac{1}{n_2}\sum^{n_2}_{i=1}\boldsymbol{\psi}_{\boldsymbol{\omega}_R}(Y_i)\big\|_{\mathbb R^{2R}}^2$ are less than or equal to $\kappa(0)$, and this implies that $W$ satisfies $0\leq W\leq\kappa(0)\big(({n_1}-1)^{-1}+({n_2}-1)^{-1}\big).$ Now, as a lower bound for $\mathbb E\left[V\right]$ and $\mathbb E\left[U\right]$, we observe that 
	\begin{align*}
		\mathbb E\left[V\right]&=\mathbb E\left[U+W\right]\\
		&=\mathbb E\left[U\right]+\mathbb E\left[W\right]\\
		&\geq\mathbb E\left[U\right]. 
	\end{align*}  
	
	On the other hand, we note that $\operatorname{Var}\left[V\right]$ and $\operatorname{Var}\left[U\right]$ are both upper bounded by
	\begin{align*}
		\sqrt{\frac1\beta\operatorname{Var}\left[V\right]}&\leq \sqrt{\frac2\beta\operatorname{Var}\left[U\right]+\frac2\beta\operatorname{Var}\left[W\right]}\\
		&\leq \sqrt{\frac2\beta\operatorname{Var}\left[U\right]}+\sqrt{\frac2\beta\operatorname{Var}\left[W\right]}\\
		&\stackrel{(a)}{\leq} \sqrt{\frac2\beta\operatorname{Var}\left[U\right]}+\sqrt{\frac1{2\beta}\kappa(0)^2\bigg(\frac1{{n_1}-1}+\frac1{{n_2}-1}\bigg)^2}\\
		&\stackrel{(b)}{\leq} \sqrt{\frac2\beta\operatorname{Var}\left[U\right]}+\sqrt{\frac2\beta}\kappa(0)\bigg(\frac1{{n_1}}+\frac1{{n_2}}\bigg),
	\end{align*}
	where the inequality (a) follows the fact that the variance of bounded variable is also bounded, (b) follows from the fact that $(x-1)^{-1}\leq 2x^{-1}$ for all $x\geq 2$. For the critical value term, recall that the critical value is $q_{{n_1},{n_2},1-\alpha}=\inf\{t:F_{V}^{\pi}(t)\geq1-\alpha\},$  and if we substitute the test statistic $V$ with $U,$ then the critical value is $q_{{n_1},{n_2},1-\alpha}^u=\inf\{t:F_{U}^{\pi}(t)\geq1-\alpha\}.$ We claim that $q_{{n_1},{n_2},1-\alpha}$ and $q_{{n_1},{n_2},1-\alpha}^u$ are both upper bounded by 
	\begin{align*}
		q_{{n_1},{n_2},1-\alpha}\leq q_{{n_1},{n_2},1-\alpha}^u+\kappa(0)\bigg(\frac1{{n_1}-1}+\frac1{{n_2}-1}\bigg).
	\end{align*}
	To see this, note that we have $|V-U|\leq \kappa(0)\big(({n_1}-1)^{-1}+({n_2}-1)^{-1}\big)$ from Equation \eqref{thm5decom}. Based on this fact, for given $(Z_1,\ldots,Z_N)=(\mathcal X_{n_1},\mathcal Y_{n_2})$ and $\boldsymbol{\omega}_R$, if a permutation $\pi\in\Pi_N$ satisfies $U(Z_{\pi(1)},\dots,Z_{\pi(N)};\boldsymbol{\omega}_R)\leq q^u_{{n_1},{n_2},1-\alpha},$ then we also have $V(Z_{\pi(1)},\dots,Z_{\pi(N)};\boldsymbol{\omega}_R)\leq q^u_{{n_1},{n_2},1-\alpha}+\kappa(0)\big(({n_1}-1)^{-1}+({n_2}-1)^{-1}\big).$ This yields the desired result $q_{{n_1},{n_2},1-\alpha}\leq q_{{n_1},{n_2},1-\alpha}^u+\kappa(0)\big(({n_1}-1)^{-1}+({n_2}-1)^{-1}\big),$ and further, we also get $q_{{n_1},{n_2},1-\alpha}\leq q_{{n_1},{n_2},1-\alpha}^u+2\kappa(0)\big({n_1}^{-1}+{n_2}^{-1}\big),$ using $(x-1)^{-1}\leq 2x^{-1}$ for all $x\geq 2.$
	
	Combining the above results, we define an event 
	\begin{align}\label{thm5prob_kappa}
		\mathcal B_{\beta}:=\bigg\{\mathbb E\left[U\right] \geq \sqrt{\frac2\beta\operatorname{Var}\left[U\right]}+q_{{n_1},{n_2},1-\alpha}^u+\kappa(0)\bigg(\sqrt{\frac{2}{ \beta}}+2\bigg)\bigg(\frac1{{n_1}}+\frac1{{n_2}}\bigg)\bigg\},
	\end{align}
	then it is straightforward to see that $\mathcal B_\beta \subseteq \mathcal B_{V,\beta} \cap \mathcal B_{U,\beta}.$ Now, our strategy is to show that the probability $\mathbb P(\mathcal B_\beta)$ becomes 1 with sufficiently large ${n_1},{n_2}$, and $R$. To start with, similar to Corollary \ref{negcor}, we can assume $\kappa(0)=1$ throughout the proof. Then the event $\mathcal B_\beta$ becomes
	\begin{align*}
		\mathcal B_{\beta}=\bigg\{\mathbb E\left[U\right] \geq \sqrt{\frac2\beta\operatorname{Var}\left[U\right]}+q_{{n_1},{n_2},1-\alpha}^u+\bigg(\sqrt{\frac{2}{ \beta}}+2\bigg)\bigg(\frac1{{n_1}}+\frac1{{n_2}}\bigg)\bigg\},
	\end{align*}
	and now we examine the three terms in the above event.
	
	\subsubsection*{Expectation of $U$}
	For $\mathbb E\left[U\right],$ we note that the test statistic $U$ is an unbiased estimator, and hence
	\begin{equation}
		\begin{aligned}\label{thm5lhs}
			\mathbb E\left[U\right]&=\mathbb E_{X\times Y}\big[\mathbb E_{\omega}\left[U\,|\,\mathcal X_{n_1},\mathcal Y_{n_2}\right]\big]\\
			&=\mathbb E_{X\times Y}\big[\widehat {\mathrm{MMD}}_u^2( \mathcal{X}_{n_1}, \mathcal{Y}_{n_2};\mathcal H_k)\big]\\
			&=\mathrm{MMD}^2\left(P_X,P_Y;\mathcal H_k\right).
		\end{aligned}
	\end{equation}

	\subsubsection*{Upper bound for the variance of $U$}
	For $\sqrt{\frac2\beta\operatorname{Var}\left[U\right]},$  consider the decomposition of $\operatorname{Var}\left[U\right]$ as follows:
	\begin{equation}
		\begin{aligned}\label{thm5evve}
			\operatorname{Var}\left[U\right]&=\mathbb E_{X\times Y}\big[\operatorname{Var}_{\omega}[U\,|\, \mathcal X_{n_1},\mathcal Y_{n_2} ] \big]+\operatorname{Var}_{X\times Y}\big[\mathbb E_{\omega}[U\,|\, \mathcal X_{n_1},\mathcal Y_{n_2} ] \big]\\
			&=\mathbb E_{X\times Y}\big[\operatorname{Var}_{\omega}[U\,|\, \mathcal X_{n_1},\mathcal Y_{n_2} ] \big]+\operatorname{Var}_{X\times Y}\big[\widehat {\mathrm{MMD}}_u^2( \mathcal{X}_{n_1}, \mathcal{Y}_{n_2};\mathcal H_k) \big].
		\end{aligned}
	\end{equation}
	For the first term in the last equation, recall that the statistic $U$ is 
	\begin{align*}
		U(\mathcal X_{n_1},\mathcal Y_{n_2};\boldsymbol{\omega}_R)
		&= \frac{1}{{n_1}({n_1}-1)} \sum_{1\leq i\neq j \leq {n_1}} \langle \boldsymbol{\psi}_{\boldsymbol{\omega}_R}(X_i),\boldsymbol{\psi}_{\boldsymbol{\omega}_R}(X_j)\rangle-\frac{2}{{n_1} {n_2}} \sum_{i=1}^{n_1} \sum_{j=1}^{n_2} \langle \boldsymbol{\psi}_{\boldsymbol{\omega}_R}(X_i),\boldsymbol{\psi}_{\boldsymbol{\omega}_R}(Y_j)\rangle\\ &\quad +\frac{1}{{n_2}({n_2}-1)} \sum_{1\leq i\neq j \leq {n_2}}  \langle \boldsymbol{\psi}_{\boldsymbol{\omega}_R}(Y_i),\boldsymbol{\psi}_{\boldsymbol{\omega}_R}(Y_j)\rangle.
	\end{align*}
	Here, we emphasize that the inner product $\langle \boldsymbol{\psi}_{\boldsymbol{\omega}_R}(X_i),\boldsymbol{\psi}_{\boldsymbol{\omega}_R}(X_j)\rangle$ and similar terms are actually the sample means. To be specific, observe that
	\begin{align}
	\langle \boldsymbol{\psi}_{\boldsymbol{\omega}_R}(X_i),\boldsymbol{\psi}_{\boldsymbol{\omega}_R}(X_j)\rangle=\frac1R\sum^R_{r=1}\langle {\psi_{\omega_r}}(X_i),{\psi_{\omega_r}}(X_j)\rangle.
	\end{align}
	Therefore, conditioning on the samples $\mathcal X_{n_1}$ and $\mathcal Y_{n_2}$, the statistic $U$ can be written as the average of $R$ conditionally i.i.d. random variables,
	\begin{align}\label{U is mean of U1}
		U(\mathcal X_{n_1},\mathcal Y_{n_2};\boldsymbol{\omega}_R)=\frac1R\sum_{r=1}^R U_1(\mathcal X_{n_1},\mathcal Y_{n_2};\omega_r),
	\end{align}
	where the randomness comes only from the i.i.d. random frequencies $\omega_1,\ldots,\omega_R$ and $U_1$ is defined as
	\begin{equation}    \label{U_1}
		\begin{aligned}
			&U_1(\mathcal X_{n_1},\mathcal Y_{n_2};\omega)\\:=&\frac{1}{{n_1}({n_1}-1)} \sum_{1\leq i\neq j \leq {n_1}} \langle {\psi_{\omega}}(X_i),{\psi_{\omega}}(X_j)\rangle-\frac{2}{{n_1} {n_2}} \sum_{i=1}^{n_1} \sum_{j=1}^{n_2} \langle {\psi_{\omega}}(X_i),{\psi_{\omega}}(Y_j)\rangle\\
			&\quad+\frac{1}{{n_2}({n_2}-1)} \sum_{1\leq i\neq j \leq {n_2}}  \langle {\psi_{\omega}}(Y_i),{\psi_{\omega}}(Y_j)\rangle\\
			=&\frac{1}{{n_1}({n_1}-1)} \sum_{1\leq i\neq j \leq {n_1}} \kappa(0)\cos\big({\omega^{\top}\left(X_i-X_j\right)}\big)-\frac{2}{{n_1} {n_2}} \sum_{i=1}^{n_1} \sum_{j=1}^{n_2} \kappa(0)\cos\big({\omega^{\top}\left(X_i-Y_j\right)}\big)\\
			&\quad+\frac{1}{{n_2}({n_2}-1)} \sum_{1\leq i\neq j \leq {n_2}} \kappa(0)\cos\big({\omega^{\top}\left(Y_i-Y_j\right)}\big).
		\end{aligned}
	\end{equation}
	Hence, the conditional variance of $U$ in Equation \eqref{thm5evve} can be written as
	\begin{align}
		\operatorname{Var}[U\,|\, \mathcal X_{n_1},\mathcal Y_{n_2} ]
		=\frac{1}{R} \operatorname{Var}[U_1\,|\, \mathcal X_{n_1},\mathcal Y_{n_2} ].
	\end{align}
	Also, since $|\!\cos(x)|\leq1$ for all $x\in\mathbb R$, we note that $|U_1|\leq4\kappa(0).$ Therefore, since its variance is also bounded, we conclude that the first term in Equation \eqref{thm5evve} is bounded by
	\begin{equation}    \label{thm5evve1}
		\begin{aligned}
			\mathbb E_{X\times Y}\big[\operatorname{Var}_{\omega}[U\,|\, \mathcal X_{n_1},\mathcal Y_{n_2} ] \big]&=\frac{1}{R}  \mathbb E_{X\times Y}\big[\operatorname{Var}_{\omega}[U_1\,|\, \mathcal X_{n_1},\mathcal Y_{n_2} ] \big]\\
			&\leq\frac{16\kappa(0)^2}{R}.
		\end{aligned}
	\end{equation}
	
	For the second term in Equation \eqref{thm5evve}, we leverage the result of \citet[Appendix F]{Kim2022}.
	Let $h(x_1,x_2,y_1,y_2):=k(x_1,x_2)+k(y_1,y_2)-k(x_1,y_2)-k(x_2,y_1).$ Then, there exists some positive constant $C_1$ such that the variance of the unbiased estimator of MMD can be bounded as
	$$
	\operatorname{Var}\big[\widehat {\mathrm{MMD}}_u^2( \mathcal{X}_{n_1}, \mathcal{Y}_{n_2};\mathcal H_k) \big]\leq C_1\bigg(\frac{\sigma^2_{1,0}}{{n_1}}+\frac{\sigma^2_{0,1}}{{n_2}}+\bigg(\frac{1}{{n_1}}+\frac1{{n_2}}\bigg)^2\sigma^2_{2,2}\bigg)
	$$
	for
	\begin{align*}
		\sigma^2_{1,0}&:=\operatorname{Var}\Big[\mathbb E_{X',Y,Y'}\big[h(X,X',Y,Y') \big]\Big],\\
		\sigma^2_{0,1}&:=\operatorname{Var}\Big[\mathbb E_{X,X',Y'}\big[h(X,X',Y,Y') \big]\Big],\\
		\sigma^2_{2,2}&:=\operatorname{Var}\Big[h(X,X',Y,Y') \Big],
	\end{align*}
	where $X'$ is an independent copy of $X$, and $Y'$ is an independent copy of $Y.$ We note that the kernel $k$ is bounded and Bochner's theorem (Lemma \ref{bochner}) guarantees the existence of the nonnegative Borel measure $\Lambda$ that satisfies 
	$$
	k(x,y)=\kappa(x-y)=\int_{\mathbb R^d} \cos\bigl({\omega^\top(x-y)}\bigr)d\Lambda(\omega).
	$$
	Since $|\!\cos(x)|\leq1$ for all $x\in \mathbb R$ and the measure $\Lambda$ is nonnegative, we have
	\begin{align} \label{kbound}
		|k(x,y)|=\bigg|\int_{\mathbb R^d} \cos\bigl({\omega^\top(x-y)}\bigr)d\Lambda(\omega)\bigg|\leq\int_{\mathbb R^d} |\cos\bigl({\omega^\top(x-y)}\bigr)|d\Lambda(\omega)\leq\int_{\mathbb R^d} 1d\Lambda(\omega)=\kappa(0)
	\end{align}
	for all $x,y\in \mathbb R^d.$
	Therefore, the kernel $k$ is bounded by $\kappa(0),$ and the term $|h(X,X',Y,Y')|$ is bounded by $4\kappa(0)$. This yields 
	$$
	\max\left(\sigma^2_{1,0},\sigma^2_{0,1},\sigma^2_{2,2}\right)\leq16\kappa(0)^2.
	$$
	Now, we conclude that the second term in Equation \eqref{thm5evve} is bounded by
	\begin{align}\label{thm5evve2}
		\operatorname{Var}\big[\widehat {\mathrm{MMD}}_u^2( \mathcal{X}_{n_1}, \mathcal{Y}_{n_2};\mathcal H_k) \big]&\leq 16C_1\kappa(0)^2\left(\frac{1}{{n_1}}+\frac{1}{{n_2}}+\bigg(\frac{1}{{n_1}}+\frac{1}{{n_2}}\bigg)^2\right).
	\end{align}

	To sum up, combining results in Equations  \eqref{thm5evve1} and \eqref{thm5evve2}, we have
	\begin{align*}
		\operatorname{Var}\left[U\right]&\leq\mathbb E_{X\times Y}\big[\operatorname{Var}[U\,|\, \mathcal X_{n_1},\mathcal Y_{n_2} ] \big]+\operatorname{Var}\big[\widehat {\mathrm{MMD}}_u^2( \mathcal{X}_{n_1}, \mathcal{Y}_{n_2};\mathcal H_k) \big]\\
		&\leq\frac{16\kappa(0)^2}{R}+16C_1\kappa(0)^2\Bigg(\frac1{{n_1}}+\frac1{{n_2}}+\bigg(\frac1{{n_1}}+\frac1{{n_2}}\bigg)^2\Bigg)\\
		&\leq\kappa(0)^2\Bigg(\frac{16}{R}+C_2\bigg(\frac1{{n_1}}+\frac1{{n_2}}\bigg)\Bigg),
	\end{align*}
	for $C_2:=32C_1$, as $\left({n_1}^{-1}+{n_2}^{-1}\right)^2\leq {n_1}^{-1}+{n_2}^{-1}$ for ${n_1},{n_2}\geq 2$.
	Therefore, we have
	\begin{equation}
		\begin{aligned}\label{thm5rhs1}
			\sqrt{\frac{2}\beta\operatorname{Var}\left[U\right]}&\leq\frac{\kappa(0)}{\sqrt\beta}\sqrt{\frac{32}{R}+2C_2\bigg(\frac1{{n_1}}+\frac1{{n_2}}\bigg)}\\
			&\leq\frac{1}{\sqrt\beta}\Bigg(\frac6{\sqrt R}+C_3\bigg(\frac1{\sqrt {n_1}}+\frac1{\sqrt {n_2}}\bigg)\Bigg),
		\end{aligned}
	\end{equation}
	for $C_3=\sqrt{2C_2},$ where we use the fact that $\sqrt{x+y}\leq \sqrt x + \sqrt y$ for $x,y \geq 0,$ and the assumption $\kappa(0)=1.$
	\subsubsection*{Upper bound for the critical value $q_{{n_1},{n_2},1-\alpha}^u$}
	In order to derive an upper bound for $q_{{n_1},{n_2},1-\alpha}^u,$ we use the property of U-statistics as done by \citet[Appendix E, F]{Kim2022}. First, observe that Chebyshev's inequality yields 
	$$
	\mathbb P_\pi \bigg(\big|U_\pi-\mathbb E_\pi[U_\pi\,|\,\mathcal X_{n_1},\mathcal Y_{n_2},\boldsymbol{\omega}_R]  \big|\geq\sqrt{\frac1\alpha\operatorname{Var}_\pi[U_\pi\,|\,\mathcal X_{n_1},\mathcal Y_{n_2},\boldsymbol{\omega}_R]}\,\bigg|\,\mathcal X_{n_1},\mathcal Y_{n_2},\boldsymbol{\omega}_R\bigg)\leq \alpha,
	$$
	and by the definition of quantile, we have an upper bound of $q_{{n_1},{n_2},1-\alpha}^u$:
	\begin{align*}
		q_{{n_1},{n_2},1-\alpha}^u \leq \mathbb E_\pi[U_\pi\,|\,\mathcal X_{n_1},\mathcal Y_{n_2},\boldsymbol{\omega}_R]+\sqrt{\frac1\alpha\operatorname{Var}_\pi[U_\pi\,|\,\mathcal X_{n_1},\mathcal Y_{n_2},\boldsymbol{\omega}_R]}.
	\end{align*}
	For the first term of the right-hand side, since the U-statistic is centered at zero under the permutation law \citep[see e.g.,][Appendix F]{Kim2022}, we can deduce that $\mathbb E_\pi[U_\pi\,|\,\mathcal X_{n_1},\mathcal Y_{n_2},\boldsymbol{\omega}_R]=0.$ Similarly, for the second term, observe that 
	\begin{align*}
		\operatorname{Var}_\pi[U_\pi\,|\,\mathcal X_{n_1},\mathcal Y_{n_2},\boldsymbol{\omega}_R] &= \mathbb E_\pi\big[(U_\pi)^2\,|\,\mathcal X_{n_1},\mathcal Y_{n_2},\boldsymbol{\omega}_R\big]- \big(\mathbb E_\pi[U_\pi\,|\,\mathcal X_{n_1},\mathcal Y_{n_2},\boldsymbol{\omega}_R] \big)^2\\
		&=\mathbb E_\pi\big[(U_\pi)^2\,|\,\mathcal X_{n_1},\mathcal Y_{n_2},\boldsymbol{\omega}_R\big].
	\end{align*}
	Here, we note that this statistic has been carefully studied in \citet[Appendix F]{Kim2022}, and the following result holds true:
 \begin{equation}\label{kim appendix f}
     \begin{aligned}
	&\mathbb E_\pi\big[(U_\pi)^2\,|\,\mathcal X_{n_1},\mathcal Y_{n_2},\boldsymbol{\omega}_R\big] \\=&\frac{1}{{n_1}^2({n_1}-1)^2{n_2}^2({n_2}-1)^2} \sum_{(i_1,\dots,j'_2)\in\mathbf I}\mathbb E_\pi\Big[\hat h\big(Z_{\pi(i_1)},Z_{\pi(i_2)}; Z_{\pi(n_1+j_1)},Z_{\pi(n_1+j_2)}\big)\\& 
	 \,\qquad \qquad \qquad \qquad \qquad \qquad \qquad \qquad \times \hat h\big(Z_{\pi(i'_1)},Z_{\pi(i'_2)}; Z_{\pi(n_1+j'_1)},Z_{\pi(n_1+j'_2)}\big)\,\Big|\,\mathcal X_{n_1},\mathcal Y_{n_2},\boldsymbol{\omega}_R\Big],
\end{aligned}
 \end{equation}
	
	where $\hat h(x_1,x_2;y_1,y_2)$ is a kernel defined as $\hat h(x_1,x_2;y_1,y_2):=\hat k(x_1,x_2)+\hat k(y_1,y_2)-\hat k(x_1,y_2)-\hat k(x_2,y_1)$, and $\mathbf I$ is a set of indices defined as $\mathbf I:=\{(i_1,i_2,i'_1,i'_2,j_1,j_2,j'_1,j'_2)\in\mathbb N^8_+ : (i_1,i_2),(i'_1,i'_2)\in \mathbf i^{n_1}_2, (j_1,j_2),(j'_1,j'_2)\in \mathbf i^{n_2}_2,\#|\{i_1,i_2\}\cap\{i'_1,i'_2\}|+\#|\{j_1,j_2\}\cap\{j'_1,j'_2\}|>1  \}.$ Here $\#|A|$ denotes the cardinality of a set $A$, and $(l_1,l_2)\in\mathbf i^k_2$ implies $1\leq l_1\neq l_2\leq k.$ Recall that $\hat k(x,y)=\langle\boldsymbol{\psi}_{\boldsymbol{\omega}_R}(x),\boldsymbol{\psi}_{\boldsymbol{\omega}_R}(y)\rangle$ is the approximated kernel defined in Equation \eqref{hatk}, and we have a bound $|\hat k(x,y)|\leq \kappa(0)$ for all $x,y\in \mathbb R^d$, in Equation \eqref{hatkbound}. This implies $|\hat h(\cdot)|\leq 4\kappa(0),$ and thus $|\hat h(\cdot)\times \hat h(\cdot)|\leq 16\kappa(0)^2.$ Using this observation and counting the number of $\mathbf I$ \citep[Appendix F]{Kim2022} yields
	\begin{align*}
		\mathbb E_\pi\big[(U_\pi)^2\,|\,\mathcal X_{n_1},\mathcal Y_{n_2},\boldsymbol{\omega}_R\big]&\leq \frac{16\kappa(0)^2}{{n_1^2}({n_1}-1)^2{n_2^2}({n_2}-1)^2}\sum_{(i_1,\dots,j'_2)\in\mathbf I} 1 \\
		&\leq C_4\kappa(0)^2\bigg(\frac1{{n_1}}+\frac1{{n_2}}\bigg)^2
	\end{align*}
	for some positive constant $C_4,$ regardless of the realized values of $\mathcal X_{n_1},\mathcal Y_{n_2}$ and $\boldsymbol{\omega}_R.$ Therefore, we get
	\begin{align*}
		\sqrt{\frac{1}{\alpha}\operatorname{Var}_\pi[U_\pi\,|\,\mathcal X_{n_1},\mathcal Y_{n_2},\boldsymbol{\omega}_R]} &\leq \sqrt{\frac{C_4}{\alpha}}\kappa(0)\bigg(\frac1{{n_1}}+\frac1{{n_2}}\bigg)
	\end{align*}
	and we conclude that 
	\begin{equation}
		\begin{aligned}\label{crit.ub2}
			q_{{n_1},{n_2},1-\alpha}^u&\leq  \sqrt{\frac{1}{\alpha}\operatorname{Var}_\pi[U_\pi\,|\,\mathcal X_{n_1},\mathcal Y_{n_2},\boldsymbol{\omega}_R]}\\
			&\leq\sqrt{\frac{C_4}{\alpha}}\kappa(0)\bigg(\frac1{{n_1}}+\frac1{{n_2}}\bigg)\\
			&\leq C_5(\alpha)\bigg(\frac1{{n_1}}+\frac1{{n_2}}\bigg)
		\end{aligned}
	\end{equation}
	for $C_5(\alpha):=\sqrt{\alpha^{-1}C_4}.$

	\subsubsection*{Finding $\beta_{{n_1},{n_2},R}$}
	Based on Equations \eqref{thm5rhs1} and \eqref{crit.ub2} that we obtained so far, we can derive the result as follows:
	\begin{align*}
		&\quad\sqrt{\frac2\beta\operatorname{Var}\left[U\right]}+q_{{n_1},{n_2},1-\alpha}^u+\bigg(\sqrt{\frac{2}{ \beta}}+2\bigg)\bigg(\frac1{{n_1}}+\frac1{{n_2}}\bigg)\\
		&\leq\frac{1}{\sqrt \beta}\Bigg(\frac6{\sqrt R}+C_3\bigg(\frac1{\sqrt {n_1}}+\frac1{\sqrt {n_2}}\bigg)+ \sqrt2\bigg(\frac1{{n_1}}+\frac1{{n_2}}\bigg)\Bigg)+\left(C_5(\alpha)+2\right)\bigg(\frac1{{n_1}}+\frac1{{n_2}}\bigg)\\
		&\leq C(\alpha)\bigg(\frac1{\sqrt{\beta R}}+\frac1{\sqrt{\beta {n_1}}}+\frac1{\sqrt{\beta {n_2}}}+\frac1{{n_1}}+\frac1{{n_2}}\bigg)
	\end{align*}
	for a constant $C(\alpha):=\max\left\{6,C_3+\sqrt2,C_5(\alpha)+2\right\}.$ 
	Here, we consider a sequence that converges to zero, $\beta_{{n_1},{n_2},R}=\max\big\{\frac{1}{\log({n_1})},\frac{1}{\log({n_2})},\frac{1}{\log(R)}\big\}.$ Then we get
	\begin{align*}
		&\quad\sqrt{\frac2{\beta_{{n_1},{n_2},R}}\operatorname{Var}\left[U\right]}+q_{{n_1},{n_2},1-\alpha}^u+\bigg(\sqrt{\frac{2}{ \beta_{{n_1},{n_2},R}}}+2\bigg)\bigg(\frac1{{n_1}}+\frac1{{n_2}}\bigg)\\&\leq C(\alpha) \Bigg(\bigg(\frac{\log R}{R}\bigg)^{1/2}+\bigg(\frac{\log {n_1}}{{n_1}}\bigg)^{1/2}+\bigg(\frac{\log {n_2}}{{n_2}}\bigg)^{1/2}+\frac{1}{n_1}+\frac1{{n_2}}\Bigg).
	\end{align*}
	Since $\lim_{x\rightarrow\infty}(\log x /x)^{1/2}=0$ and $\lim_{x\rightarrow\infty}(1/x)=0$, there exist $N_{(P_X,P_Y)}$ and 
	$R_{(P_X,P_Y)}$ such that ${n_1},{n_2}\geq N_{(P_X,P_Y)}$ and $R\geq R_{(P_X,P_Y)}$ implies
	\begin{align*}
		\mathrm{MMD}^2\left(P_X,P_Y;\mathcal H_k\right)\geq
		C(\alpha) \left(\bigg(\frac{\log R}{R}\bigg)^{1/2}+\bigg(\frac{\log {n_1}}{{n_1}}\bigg)^{1/2}+\bigg(\frac{\log {n_2}}{{n_2}}\bigg)^{1/2}+\frac{1}{n_1}+\frac1{{n_2}}\right).
	\end{align*}
	Then we can deduce that
	\begin{align*}
		\mathbb P(\mathcal B_{\beta_{{n_1},{n_2},R}})&=\mathbb P \left\{\mathbb E\left[U\right]\geq \sqrt{\frac2{\beta_{{n_1},{n_2},R}}\operatorname{Var}\left[U\right]}+q_{{n_1},{n_2},1-\alpha}^u+\bigg(\sqrt{\frac{2}{ \beta_{{n_1},{n_2},R}}}+2\bigg)\bigg(\frac1{{n_1}}+\frac1{{n_2}}\bigg)\right\}\\
		&\geq\mathbb P \left\{\mathrm{MMD}^2\left(P_X,P_Y;\mathcal H_k\right)\geq \sqrt{\frac2{\beta_{{n_1},{n_2},R}}\operatorname{Var}\left[U\right]}+q_{{n_1},{n_2},1-\alpha}^u+\bigg(\sqrt{\frac{2}{ \beta_{{n_1},{n_2},R}}}+2\bigg)\bigg(\frac1{{n_1}}+\frac1{{n_2}}\bigg)\right\}\\
		&\geq\mathbb P \left\{\mathrm{MMD}^2\left(P_X,P_Y;\mathcal H_k\right)\geq C(\alpha) \left(\bigg(\frac{\log R}{R}\bigg)^{1/2}+\bigg(\frac{\log {n_1}}{{n_1}}\bigg)^{1/2}+\bigg(\frac{\log {n_2}}{{n_2}}\bigg)^{1/2}+\frac{1}{n_1}+\frac1{{n_2}}\right)\right\}\\
		&=1
	\end{align*}
	for ${n_1},{n_2}\geq N_{(P_X,P_Y)}$ and $R\geq R_{(P_X,P_Y)}$. Note that the sequence $\beta_{{n_1},{n_2},R}$ converges to 0 and this completes the proof.

	\subsection{Proof of Theorem \ref{L2thm}} \label{Section: proof of L2thm}
	To begin with, we introduce some assumptions and useful facts for ease of analysis. Note that we use a translation invariant kernel which can be decomposed as $$
	k_\lambda(x,y)=\kappa_\lambda(x-y):=\prod^d_{i=1}\frac1{\lambda_i}\kappa_i\left(\frac{x_i-y_i}{\lambda_i}\right)
	$$
	for $\lambda=(\lambda_1,\ldots,\lambda_d)\in(0,\infty)^d.$ Here, without loss of generality, we assume that $\prod^d_{i=1}\kappa_i(0)=1.$ If not, this can be done by scaling the bandwidth and $\kappa$ with a constant while the kernel $k$ remains unchanged. To be specific,
	$$
	k(x,y)=\prod^d_{i=1}\frac1{\lambda_i}\kappa_i\left(\frac{x_i-y_i}{\lambda_i}\right)=\prod^d_{i=1}\frac1{\lambda^*_i}\kappa^*_i\left(\frac{x_i-y_i}{\lambda^*_i}\right)
	$$
	holds where $\kappa^*_i(x):=\kappa_i(x/\kappa_i(0))/\kappa_i(0),~\lambda^*_i=\lambda_i/\kappa_i(0),$ and then $\prod^d_{i=1}\kappa^*_i(0)=1.$ Now, note that our assumption yields $\kappa_\lambda(0)=(\lambda_1\cdots\lambda_d)^{-1}.$
	Also, let $C_0,C'_0>0$ be constants that satisfy 
		\begin{equation}\label{CN1}
			\begin{aligned}
				\frac1{{n_1}-1}+\frac1{{n_2}-1}\leq \frac {C_0}{n},
			\end{aligned}
		\end{equation}
		and
		\begin{equation}\label{CN2}
			\begin{aligned}
				\frac1{({n_1}-1)^2}+\frac1{({n_2}-1)^2}\leq \frac {C'_0}{{n}^2},
			\end{aligned}
		\end{equation}
		respectively.
  
	For the proof of \Cref{L2thm}, we follow a similar approach taken in \citet{schrab2023mmdaggregated} to derive an upper bound for the uniform separation rate.
	First, as in the proof of Theorem \ref{conthm}, we define an event that can be utilized concurrently for analyzing both tests $\Delta_{{n_1},{n_2},R}^{\alpha}$ and $\Delta_{{n_1},{n_2},R}^{\alpha,u}$. Consider the following two events
	\begin{align*}
		\mathcal B_{V,\beta/2}&:=\bigg\{\mathbb E\left[V\right] \geq \sqrt{\frac2\beta\operatorname{Var}\left[V\right]}+q_{{n_1},{n_2},1-\alpha}\bigg\} \quad \text{and} \\ 
		\mathcal B_{U,\beta/2}&:=\bigg\{\mathbb E\left[U\right] \geq \sqrt{\frac2\beta\operatorname{Var}\left[U\right]}+q^u_{{n_1},{n_2},1-\alpha}\bigg\},
	\end{align*}
	and suppose that there exists an event $\mathcal B_{\beta/2}\subseteq\mathcal B_{V,\beta/2}\cap\mathcal B_{U,\beta/2}$ and $\mathbb P (\mathcal B_{\beta/2})\geq1-\beta / 2$. Also, for the following two events, 
	\begin{align*}
		\mathcal A_V&:=\{V \leq q_{{n_1},{n_2},1-\alpha}\} \quad \text{and} \\ 
		\mathcal A_U&:=\{U \leq q^u_{{n_1},{n_2},1-\alpha}\},
	\end{align*}
	recall that $\mathbb P(\mathcal A_V)\leq\beta$ implies $\mathbb{P}_{X\times Y}(\Delta_{{n_1},{n_2},R}^{\alpha}(\mathcal{X}_{n_1},\mathcal{Y}_{n_2})=0)\leq \beta,$
	and similarly $\mathbb P(\mathcal A_U)\leq\beta$ implies $\mathbb{P}_{X\times Y}(\Delta_{{n_1},{n_2},R}^{\alpha,u}(\mathcal{X}_{n_1},\mathcal{Y}_{n_2})=0)\leq \beta.$
	Then Chebyshev's inequality yields the desired result as
	\begin{align*}
		\mathbb P(\mathcal A_V)&= \mathbb P(\mathcal A_V,\mathcal B_{\beta/2} )+ \mathbb P(\mathcal A_V\,|\,\mathcal B_{\beta/2}^c )\mathbb P(\mathcal B_{\beta/2}^c)\\
		&\leq\mathbb P(\mathcal A_V,\mathcal B_{V,\beta/2} )+\frac\beta2\\
		&\leq\mathbb P\bigg(V \leq \mathbb E[V]-\sqrt{\frac2\beta\operatorname{Var}[V]}\bigg)+\frac\beta2\\
		&=\mathbb P\bigg(\sqrt{\frac2\beta\operatorname{Var}\left[V\right]} \leq \mathbb E\left[V\right]-V \bigg)+\frac\beta2\\
		&\leq\mathbb P\bigg(| V - \mathbb E[V]| \geq\sqrt{\frac2\beta\operatorname{Var}\left[V\right]}\bigg)+\frac\beta2\\
		&\leq \frac\beta2+\frac\beta2=\beta,
	\end{align*}
	and similarly we can get $\mathbb P\left(\mathcal A_U\right)\leq\beta.$ 
	
	Therefore, our strategy is to identify such event $\mathcal B_{\beta/2}.$
	To begin with, we carefully analyze the difference between the statistics $V$ and $U$, following the logic similar to \citet[Appendix E.11]{kim2023diff}. From Equation \eqref{thm5decom}, the statistic $V$ can be decomposed as
	\begin{equation*}
		\begin{aligned}
			V&=U+\left(\frac {\kappa_\lambda(0)}{{n_1}-1}-\frac{1}{{n_1}-1}\bigg\| \frac{1}{n_1}\sum^{n_1}_{i=1}\boldsymbol{\psi}_{\boldsymbol{\omega}_R}(X_i)\bigg\|_{\mathbb R^{2R}}^2+\frac{\kappa_\lambda(0)}{{n_2}-1}-\frac{1}{{n_2}-1}\bigg\| \frac1{{n_2}}\sum^{n_2}_{i=1}\boldsymbol{\psi}_{\boldsymbol{\omega}_R}(Y_i)\bigg\|_{\mathbb R^{2R}}^2\right)\\
			&=U+\frac {\kappa_\lambda(0)}{{n_1}-1}-\frac {\kappa_\lambda(0)}{{n_1}({n_1}-1)}+\frac {\kappa_\lambda(0)}{{n_2}-1}-\frac {\kappa_\lambda(0)}{{n_2}({n_2}-1)}\\
			&\quad-\bigg(\underbrace{\frac{1}{{n_1^2}({n_1}-1)}\sum_{1\leq i\neq j\leq {n_1}}\hat k(X_i,X_j)+\frac{1}{{n_2^2}({n_2}-1)}\sum_{1\leq i\neq j\leq {n_2}}\hat k(Y_i,Y_j)}_{:=W'}\bigg)\\
			&=U-W'+\kappa_\lambda(0)\bigg(\frac1{{n_1}}+\frac1{{n_2}}\bigg).
		\end{aligned}
	\end{equation*}
	
	Here, we claim that the event
	$$
	\mathcal B_{\beta/2}:= \Bigg\{\mathbb E\left[U\right] \geq \sqrt{\frac4\beta\operatorname{Var}\left[U\right]}+q_{{n_1},{n_2},1-\alpha}^u+\mathbb E\left[W'\right]+\sqrt{\frac4\beta\operatorname{Var}\left[W'\right]}+C'_0\frac{\kappa_\lambda(0)}{{n}^2}\Bigg\}
	$$
	with a positive constant $C'_0>0$ defined in \Cref{CN2} satisfies $\mathcal B_{\beta/2}\subseteq\mathcal B_{V,\beta/2}\cap\mathcal B_{U,\beta/2}.$
	To see this, we would first show $\mathcal B_{\beta/2}\subseteq\mathcal B_{U,\beta/2}$ and then show $\mathcal B_{\beta/2}\subseteq\mathcal B_{V,\beta/2}.$ Now, note that the nonnegativity of the kernel $k$ guarantees that the inequality
	{\allowdisplaybreaks
		\begin{align*}
		\mathbb E\left[W'\right]&=\mathbb E_{X\times Y}\big[\mathbb E_{\omega}[W'\,|\, \mathcal X_{n_1}, \mathcal Y_{n_2} ] \big]\\
		&=\mathbb E_{X}\bigg[\frac{1}{{n_1^2}({n_1}-1)}\sum_{1\leq i\neq j\leq {n_1}}\mathbb E_{\omega}\big[\hat k(X_i,X_j)\,\big|\,\mathcal X_{n_1},\mathcal Y_{n_2}\big]\bigg]\\&\quad+\mathbb E_{Y}\bigg[\frac{1}{{n_2^2}({n_2}-1)}\sum_{1\leq i\neq j\leq {n_2}}\mathbb E_{\omega}\big[\hat k(Y_i,Y_j)\,\big|\,\mathcal X_{n_1},\mathcal Y_{n_2}\big]\bigg]\\
		&=\mathbb E_{X}\bigg[\frac{1}{{n_1^2}({n_1}-1)}\sum_{1\leq i\neq j\leq {n_1}} k(X_i,X_j)\bigg]+\mathbb E_{Y}\bigg[\frac{1}{{n_2^2}({n_2}-1)}\sum_{1\leq i\neq j\leq {n_2}}k(Y_i,Y_j)\bigg]\\
		&=\frac{1}{n_1}\mathbb E_{X_1 \times X_2}\big[k(X_1,X_2)\big]  +\frac1{{n_2}}\mathbb E_{Y_1 \times Y_2}\big[k(Y_1,Y_2)\big]\\
		&\geq 0.
	\end{align*}
	}
	Based on this observation, it can be shown that the right-hand side in the event $\mathcal B_{\beta/2}$ is an upper bound for the right-hand side in the event $\mathcal B_{U,\beta/2}$, i.e., 
	\begin{align*}
		\sqrt{\frac2\beta\operatorname{Var}\left[U\right]}+q^u_{{n_1},{n_2},1-\alpha}\leq\sqrt{\frac4\beta\operatorname{Var}\left[U\right]}+q_{{n_1},{n_2},1-\alpha}^u+\mathbb E\left[W'\right]+\sqrt{\frac4\beta\operatorname{Var}\left[W'\right]}+C'_0\frac{\kappa_\lambda(0)}{{n}^2},
	\end{align*}
	and this implies $\mathcal B_{\beta/2}\subseteq\mathcal B_{U,\beta/2}.$
	
	For the inequality $\mathcal B_{\beta/2}\subseteq\mathcal B_{V,\beta/2},$ observe that
	$$
	\mathbb E[V]=\mathbb E [U]-\mathbb E[W']+\kappa_\lambda(0)\bigg(\frac1{{n_1}}+\frac1{{n_2}}\bigg)
	$$
	and plugging this equality into the event $\mathcal B_{V,\beta/2}$ yields
	\begin{align*}
		\mathcal B_{V,\beta/2}&=\bigg\{\mathbb E\left[V\right] \geq \sqrt{\frac2\beta\operatorname{Var}\left[V\right]}+q_{{n_1},{n_2},1-\alpha}\bigg\}\\
		&=\bigg\{\mathbb E\left[U\right] \geq \sqrt{\frac2\beta\operatorname{Var}\left[V\right]}+\mathbb E[W']+q_{{n_1},{n_2},1-\alpha}-\kappa_\lambda(0)\bigg(\frac1{{n_1}}+\frac1{{n_2}}\bigg)\bigg\}.
	\end{align*}

	Here, note that we have
	\begin{align*}
		\sqrt{\frac{2}{\beta}\operatorname{Var}[V]}&\leq\sqrt{\frac{2}{\beta}\operatorname{Var}[U-W']}\\
		&\leq\sqrt{\frac{4}{\beta}\operatorname{Var}[U]+\frac{4}{\beta}\operatorname{Var}[W']}\\
		&\leq\sqrt{\frac{4}{\beta}\operatorname{Var}[U]}+\sqrt{\frac{4}{\beta}\operatorname{Var}[W']}
	\end{align*}
	and
	\begin{align*}
		q_{{n_1},{n_2},1-\alpha}-\kappa_\lambda(0)\bigg(\frac1{{n_1}}+\frac1{{n_2}}\bigg)&\leq q_{{n_1},{n_2},1-\alpha}^u+\kappa_\lambda(0)\bigg(\frac1{{n_1}-1}+\frac1{{n_2}-1}\bigg)-\kappa_\lambda(0)\bigg(\frac1{{n_1}}+\frac1{{n_2}}\bigg)\\
		&=q_{{n_1},{n_2},1-\alpha}^u+\kappa_\lambda(0)\bigg(\frac1{{n_1}({n_1}-1)}+\frac1{{n_2}({n_2}-1)}\bigg)\\
		&\leq q_{{n_1},{n_2},1-\alpha}^u+C'_0\frac{\kappa_\lambda(0)}{{n}^2}
	\end{align*}
	for the constant $C'_0>0$ defined in \Cref{CN2}. These two facts guarantee that the right-hand side in the event $\mathcal B_{\beta/2}$ is an upper bound for the right-hand side in the event $\mathcal B_{V,\beta/2}$, i.e., 
	\begin{align*}
		&\sqrt{\frac2\beta\operatorname{Var}\left[V\right]}+\mathbb E[W']+q_{{n_1},{n_2},1-\alpha}-\kappa_\lambda(0)\bigg(\frac1{{n_1}}+\frac1{{n_2}}\bigg)\\ \leq&\sqrt{\frac4\beta\operatorname{Var}\left[U\right]}+q_{{n_1},{n_2},1-\alpha}^u+\mathbb E\left[W'\right]+\sqrt{\frac4\beta\operatorname{Var}\left[W'\right]}+C'_0\frac{\kappa_\lambda(0)}{{n}^2}
	\end{align*}
	and we conclude that $\mathcal B_{\beta/2}\subseteq\mathcal B_{V,\beta/2}.$
	
	Now, we move our focus to find a sufficient condition for $\mathbb P (\mathcal B_{\beta/2})\geq1-\beta / 2.$ Observe that Chebyshev's inequality yields
	$$
	\mathbb P_\pi \Bigg(\big|U_\pi-\mathbb E_\pi[U_\pi\,|\,\mathcal X_{n_1},\mathcal Y_{n_2},\boldsymbol{\omega}_R] \big| \geq\sqrt{\frac1\alpha\operatorname{Var}_\pi[U_\pi\,|\,\mathcal X_{n_1},\mathcal Y_{n_2},\boldsymbol{\omega}_R]}\,\Bigg|\,\mathcal X_{n_1},\mathcal Y_{n_2},\boldsymbol{\omega}_R\Bigg)\leq \alpha,
	$$
	and by the definition of a quantile, we have an upper bound for $q_{{n_1},{n_2},1-\alpha}^u$:
	\begin{align*}
		q_{{n_1},{n_2},1-\alpha}^u \leq \mathbb E_\pi[U_\pi\,|\,\mathcal X_{n_1},\mathcal Y_{n_2},\boldsymbol{\omega}_R]+\sqrt{\frac1\alpha\operatorname{Var}_\pi[U_\pi\,|\,\mathcal X_{n_1},\mathcal Y_{n_2},\boldsymbol{\omega}_R]}.
	\end{align*}
	For the first term of the right-hand side, since the U-statistic is centered at zero under the permutation law \citep[see e.g.,][Appendix F]{Kim2022}, we can deduce that $\mathbb E_\pi[U_\pi\,|\,\mathcal X_{n_1},\mathcal Y_{n_2},\boldsymbol{\omega}_R]=0.$ Then, since Markov's inequality yields
	$$
	\mathbb P\bigg(\sqrt{\frac1\alpha\operatorname{Var}_\pi[U_\pi\,|\,\mathcal X_{n_1},\mathcal Y_{n_2},\boldsymbol{\omega}_R]}<\sqrt{\frac2{\alpha\beta}\mathbb E\big[\operatorname{Var}_\pi[U_\pi\,|\,\mathcal X_{n_1},\mathcal Y_{n_2},\boldsymbol{\omega}_R]\big]}\bigg)\geq 1-\frac\beta2,
	$$
	we conclude that
	\begin{align}\label{twomoment}
		\mathbb E\left[U\right] \geq \sqrt{\frac4\beta\operatorname{Var}\left[U\right]}+\sqrt{\frac2{\alpha\beta}\mathbb E\big[\operatorname{Var}_\pi[U_\pi\,|\,\mathcal X_{n_1},\mathcal Y_{n_2},\boldsymbol{\omega}_R]\big]}+\mathbb E\left[W'\right]+\sqrt{\frac4\beta\operatorname{Var}\left[W'\right]}+C'_0\frac{\kappa_\lambda(0)}{{n}^2}
	\end{align}
	is a sufficient condition for $\mathbb P (\mathcal B_{\beta/2})\geq1-\beta/2.$
	Therefore, our goal is to analyze the above equation and to find a proper rate of $R$ and the bandwidth $\lambda_1,\dots,\lambda_d$ in terms of $R,{n_1}$ and ${n_2}$ to uniformly control both types of errors. 
	
	\subsubsection*{Lower bound for $\mathbb E\left[U\right]$}
	We first note that $\mathbb E\left[U\right]=\mathrm{MMD}^2\left(P_X,P_Y;\mathcal H_{k_\lambda}\right)$, and $\mathrm{MMD}^2\left(P_X,P_Y;\mathcal H_{k_\lambda}\right)$ can be written in $L_2$ sense \citep[Appendix E.5]{schrab2023mmdaggregated}:
	\begin{align*}
		\mathrm{MMD}^2\left(P_X,P_Y;\mathcal H_{k_\lambda}\right)&=\langle\xi,\xi\ast\kappa_\lambda\rangle_2\\
		&=\frac 12\left(\|\xi\|^2_2+\|\xi\ast\kappa_\lambda\|^2_2 -\|\xi-\xi\ast\kappa_\lambda\|^2_2\right),
	\end{align*}
	where $\xi:=p_X-p_Y,$ $\kappa_\lambda(u)=\prod^d_{i=1}(1/{\lambda_i})\kappa_i\left( u_i/{\lambda_i}\right)$ for $u\in\mathbb R^d,$ $\ast$ denotes convolution, and $\langle \cdot,\cdot \rangle_2$ is an inner product defined on $L^2(\mathbb R^d),$ i.e., $\langle f,~ g \rangle_2 = \int_{\mathbb R^d} f(x)g(x) \: dx$ for $f, g \in L^2(\mathbb R^d).$ Hence, 
	\begin{align*}
		\mathbb E\left[U\right]= \frac 12\left(\|\xi\|^2_2+\|\xi\ast\kappa_\lambda\|^2_2 -\|\xi-\xi\ast\kappa_\lambda\|^2_2\right).
	\end{align*}
	Now, we want to upper bound $\|\xi-\xi\ast\kappa_\lambda\|^2_2$, and recall that we assumed the difference of the densities $p_X-p_Y$ lying in a Sobolev ball $\mathcal S_d^{s}(M_1)$. In this setting, as shown in \citet[Appendix E.6]{schrab2023mmdaggregated}, we have
	$$
	\|\xi-\xi\ast\kappa_\lambda\|^2_2\leq S^2\|\xi\|^2_2+C_1(M_1,d,s)\sum^d_{i=1}\lambda_i^{2s}
	$$
	for some fixed constant $S\in(0,1)$ and positive constant $C_1(M_1,d,s).$ Therefore, we conclude that
	\begin{equation}\label{lbexp}
		\begin{aligned}
			\mathbb E\left[U\right]
			&= \frac 12\left(\|\xi\|^2_2+\|\xi\ast\kappa_\lambda\|^2_2-\|\xi-\xi\ast\kappa_\lambda\|^2_2\right)\\
			&\geq \frac {1-S^2}2\|\xi\|^2_2+\frac12 \|\xi\ast\kappa_\lambda\|^2_2-C_1'(M_1,d,s)\sum^d_{i=1}\lambda_i^{2s}, 
		\end{aligned}
	\end{equation}
	for $C_1'(M_1,d,s):=\frac12C_1(M_1,d,s).$
	
	\subsubsection*{Upper bound for $\sqrt{\frac4\beta\operatorname{Var}\left[U\right]}$}
	Recall the statistic $U_1$ that estimates the squared MMD with a single random feature, defined in Equation \eqref{U_1}. Note that as shown in Equation \eqref{U is mean of U1}, when the samples $\mathcal X_{n_1}$ and $\mathcal Y_{n_2}$ are given, then the statistic $U$ can be seen as the average of $R$ observations of $U_1(\omega_r)$, functions of i.i.d.~random variables $\omega_1,\ldots,\omega_R.$ Hence, we can decompose the variance of $U$ as follows:
	\begin{equation*}
		\begin{aligned}
			\operatorname{Var}\left[U\right]&=\mathbb E_{X\times Y}\big[\operatorname{Var}_{\omega}[U\,|\, \mathcal X_{n_1}, \mathcal Y_{n_2} ] \big]+\operatorname{Var}_{X\times Y}\big[\mathbb E_{\omega}[U\,|\, \mathcal X_{n_1}, \mathcal Y_{n_2} ] \big]\\
			&=\mathbb E_{X\times Y}\bigg[\frac1R\operatorname{Var}_{\omega}[U_1\,|\, \mathcal X_{n_1}, \mathcal Y_{n_2} ] \bigg]+\operatorname{Var}_{X\times Y}\big[\widehat{\mathrm{MMD}}_u^2(\mathcal X_{n_1}, \mathcal Y_{n_2};\mathcal H_{k_\lambda}) \big]\\
			&\leq\frac1R\mathbb E_{X\times Y}\Big[\mathbb E_{\omega}\big[(U_1)^2\,\big|\, \mathcal X_{n_1}, \mathcal Y_{n_2} \big] \Big]+\operatorname{Var}_{X\times Y}\big[\widehat{\mathrm{MMD}}_u^2(\mathcal X_{n_1}, \mathcal Y_{n_2};\mathcal H_{k_\lambda}) \big]\\
			&=\frac1R\mathbb E_{\omega}\Big[\mathbb E_{X\times Y}\big[(U_1)^2\,\big|\, \omega \big] \Big]+\operatorname{Var}_{X\times Y}\big[\widehat{\mathrm{MMD}}_u^2(\mathcal X_{n_1}, \mathcal Y_{n_2};\mathcal H_{k_\lambda}) \big]\\
			&=\frac1R\mathbb E_{\omega}\Big[\operatorname{Var}_{X\times Y}[U_1\,|\, \omega ]+\big(\mathbb E_{X\times Y}[U_1\,|\, \omega ]\big)^2\Big]+\operatorname{Var}_{X\times Y}\big[\widehat{\mathrm{MMD}}_u^2(\mathcal X_{n_1}, \mathcal Y_{n_2};\mathcal H_{k_\lambda}) \big]\\
			&=\frac1R\mathbb E_{\omega}\big[\operatorname{Var}_{X\times Y}[U_1\,|\, \omega ]\big]+\frac1R\mathbb E_{\omega}\big[\big(\mathbb E_{X\times Y}[U_1\,|\, \omega ]\big)^2\big]+\operatorname{Var}_{X\times Y}\big[\widehat{\mathrm{MMD}}_u^2(\mathcal X_{n_1}, \mathcal Y_{n_2};\mathcal H_{k_\lambda}) \big].
		\end{aligned}
	\end{equation*}
	Therefore, we have
	\begin{equation}    \label{uvardecom}
		\begin{aligned}
			\sqrt{\frac4\beta\operatorname{Var}\left[U\right]}&\leq \sqrt{\frac4{\beta R}\mathbb E_{\omega}\big[\operatorname{Var}_{X\times Y}[U_1\,|\, \omega ]\big]}+\sqrt{\frac4{\beta R}\mathbb E_{\omega}\big[\big(\mathbb E_{X\times Y}[U_1\,|\, \omega ]\big)^2\big]}\\
			&\quad +\sqrt{\frac4\beta\operatorname{Var}_{X\times Y}\big[\widehat{\mathrm{MMD}}_u^2(\mathcal X_{n_1}, \mathcal Y_{n_2};\mathcal H_{k_\lambda}) \big]}.
		\end{aligned}
	\end{equation}
	We start by analyzing the first term of the right hand side, $\sqrt{\frac4{\beta R}\mathbb E_{\omega}\big[\operatorname{Var}_{X\times Y}[U_1\,\big|\, \omega ]\big]}$. When $\omega$ is fixed, we note that $U_1$ is a two-sample U-statistic. We use the exact variance formula of the two-sample U-statistic~\citep[see e.g., page 38 of][]{lee1990ustatistics}. To do so, let us define a kernel for a two-sample U-statistic,
	$$
	h_\omega(x_1,x_2;y_1,y_2):=\langle\psi_{\omega}(x_1),\psi_{\omega}(x_2)\rangle+\langle\psi_{\omega}(y_1),\psi_{\omega}(y_2)\rangle-\langle\psi_{\omega}(x_1),\psi_{\omega}(y_2)\rangle-\langle\psi_{\omega}(x_2),\psi_{\omega}(y_1)\rangle,
	$$
	where $\psi_{\omega}(x)=[\sqrt{\kappa_{\lambda}(0)}\cos (\omega^{\top} x),\sqrt{\kappa_{\lambda}(0)}\sin(\omega^{\top} x)]^{\top}$ for a given $\omega$, and write the symmetrized kernel as
	\begin{align*}
		\bar h_\omega(x_1,x_2;y_1,y_2):&=\frac{1}{2!2!}\sum_{1\leq i_1 \neq i_2\leq 2}\sum_{1\leq j_1 \neq j_2\leq 2}h_\omega(x_{i_1},x_{i_2};y_{j_1},y_{j_2}) \\
		&=\frac12\langle\psi_{\omega}(x_1)-\psi_{\omega}(y_1),\psi_{\omega}(x_2)-\psi_{\omega}(y_2)\rangle+\frac12\langle\psi_{\omega}(x_1)-\psi_{\omega}(y_2),\psi_{\omega}(x_2)-\psi_{\omega}(y_1)\rangle.
	\end{align*}
	Also, let
	$$
	\bar h_{\omega,c,d}(x_1,\dots,x_c;y_1,\dots,y_d)=\mathbb E_{X\times Y}[\bar h_\omega(x_1,\dots,x_c,X_{c+1},\dots,X_{2};y_1,\dots,y_d,Y_{d+1},\dots,Y_{2})],
	$$
	and 
	$$
	\check\sigma_{\omega,c,d}^2=\operatorname{Var}_{X\times Y}\left[\bar h_{\omega,c,d}(X_1,\dots,X_c;Y_1,\dots,Y_d)\right],
	$$
	for $0\leq c,d\leq 2.$ 
	Then, the variance of the two-sample U-statistic is
	\begin{align}\label{uvarformula}
		\operatorname{Var}_{X\times Y}\left[U_1\,|\,\omega\right]=\sum_{c=0}^{2} \sum_{d=0}^{2} \binom{2}{c}\binom2d\binom{{n_1}-2}{2-c}\binom{{n_2}-2}{2-d}\binom{{n_1}}{2}^{-1}\binom{{n_2}}{2}^{-1} \check\sigma_{\omega,c, d}^2.
	\end{align}
	Here, note that we have $\check\sigma_{\omega,c,d}^2 \leq 16\kappa_{\lambda}(0)^2$ for all $0\leq c,d \leq 2$, since $|\langle\psi_{\omega}(x),\psi_{\omega}(y)\rangle| \leq \kappa_{\lambda}(0) $ for all $x,y\in \mathbb R^d.$
	Also, denote $\mu_{\omega,X}=\mathbb E_X\left[\psi_{\omega}(X)\right]$ and $\mu_{\omega,Y}=\mathbb E_Y\left[\psi_{\omega}(Y)\right]$, and observe that
	\begin{equation}\label{sigma10}
		\begin{aligned}
			\check\sigma_{\omega,1, 0}^2&=\mathbb E_{X_1}\Big[\big(\mathbb E_{X_2,Y_1,Y_2}\left[\bar h_\omega(X_1,X_2;Y_1,Y_2)\,\big|\, X_1 \right]-\left\|\mu_{\omega,X}-\mu_{\omega,Y}\right\|^2 \big)^2 \Big]\\
			&=\mathbb E_{X_1}\Big[\big( \langle \psi_{\omega}(X_1)-\mu_{\omega,X},\mu_{\omega,X}-\mu_{\omega,Y} \rangle  \big)^2 \Big]\\
			&\stackrel{(a)}{\leq} \mathbb E_{X_1}\left\| \psi_{\omega}(X_1)-\mu_{\omega,X} \right\|^2\cdot \left\| \mu_{\omega,X}-\mu_{\omega,Y} \right\|^2\\
			&\stackrel{(b)}{\leq}4\kappa_{\lambda}(0)\left\| \mu_{\omega,X}-\mu_{\omega,Y} \right\|^2,
		\end{aligned}
	\end{equation}
	where inequality~(a) is by the Cauchy--Schwarz inequality and inequality~(b) is by the fact that $-\kappa_{\lambda}(0)\leq \langle\psi_{\omega}(x),\psi_{\omega}(y)\rangle \leq \kappa_{\lambda}(0) $ for all $x,y\in \mathbb R^d.$ Similarly, we can get $\check\sigma_{\omega,0,1}^2 \leq 4\kappa_{\lambda}(0)\left\| \mu_{\omega,X}-\mu_{\omega,Y} \right\|^2.$ Now, followed by \eqref{uvarformula} and $-\kappa_{\lambda}(0)\leq \langle\psi_{\omega}(x),\psi_{\omega}(y)\rangle \leq \kappa_{\lambda}(0) $ for all $x,y\in \mathbb R^d$, we can show that there exist universal constants $C_2,C_2', C_3,C_3'>0$ such that 
	\begin{align*}
		\operatorname{Var}_{X\times Y}\left[U_1\,|\,\omega\right] &\leq C_2\kappa_{\lambda}(0)\left\| \mu_{\omega,X}-\mu_{\omega,Y} \right\|^2\bigg(\frac1{{n_1}}+\frac1{{n_2}}\bigg)+C_3\kappa_{\lambda}(0)^2\bigg(\frac1{{n_1^2}}+\frac1{{n_2^2}}+\frac1{{n_1}{n_2}}\bigg)\\
		&\leq C_2'\frac{\kappa_{\lambda}(0)}{n}\left\| \mu_{\omega,X}-\mu_{\omega,Y} \right\|^2+C_3'\frac{\kappa_{\lambda}(0)^2}{{n}^2}.
	\end{align*}
	Then, observe
	\begin{equation} \label{EVarU1}
		\begin{aligned}
			\mathbb E_{\omega}\big[\operatorname{Var}_{X\times Y}\left[U_1\,|\,\omega\right]\big] &\stackrel{(a)}\leq C_2'\frac{\kappa_{\lambda}(0)}{n}\mathrm{MMD}^2(P_X,P_Y;\mathcal H_{k_\lambda})+C_3'\frac{\kappa_{\lambda}(0)^2}{{n}^2}\\
			&\stackrel{(b)}\leq C_2'\frac{\kappa_{\lambda}(0)}{n}\|\xi\|^2_2+C_3'\frac{\kappa_{\lambda}(0)^2}{{n}^2},
		\end{aligned}
	\end{equation}
	where (a) is according to the equality $\mathbb E_{\omega}\big[\left\| \mu_{\omega,X}-\mu_{\omega,Y} \right\|^2 \big]=\mathrm{MMD}^2(P_X,P_Y;\mathcal H_{k_\lambda})$ and (b) follows from the fact that Young's convolution inequality (Lemma \ref{young}) yields $\mathrm{MMD}^2(P_X,P_Y;\mathcal H_{k_\lambda})=\langle\xi,\xi\ast\kappa_\lambda\rangle_2\leq\|\xi\|_2\|\xi\ast\kappa_\lambda\|_2 \leq\|\xi\|^2_2\|\kappa_\lambda\|_1= \|\xi\|^2_2.$ Since we have $\int_{\mathbb R^d} \kappa_\lambda(x) dx =1$ and $\kappa_\lambda(x) \geq 0$ for all $x\in \mathbb R^d.$
	Hence, using $\sqrt{x+y}\leq\sqrt x +\sqrt y$ for all $x,y\geq0,$ we can conclude that
	\begin{equation}    \label{uevar}
		\begin{aligned}
			\sqrt{\frac4{\beta R}\mathbb E_{\omega}\big[\operatorname{Var}_{X\times Y}[U_1\,|\, \omega ]\big]}&\leq C_2''(\beta)\sqrt{\frac{\kappa_{\lambda}(0)}{R{n}}}\|\xi\|_2+C_3''(\beta)\frac{\kappa_{\lambda}(0)}{\sqrt R{n}},
		\end{aligned}
	\end{equation}
	for $C_2''(\beta):=\sqrt {{4C_2'}/\beta}$ and $C_3''(\beta):=\sqrt {{4C_3'}/\beta}.$
	
	Now, we analyze the second term in the right hand side of Equation \eqref{uvardecom}, $\sqrt{\frac4{\beta R}\mathbb E_{\omega}\big[\big(\mathbb E_{X\times Y}[U_1\,|\, \omega ]\big)^2\big]}$. 
	Observe that we have
	\begin{equation}
	\begin{aligned}\label{eq: modification}
		\mathbb E_\omega\big[\big(\mathbb E_{X\times Y}[U_1\,|\, \omega]\big)^2\big]&\leq \mathbb E_\omega\Big[\big|\mathbb E_{X\times Y}[U_1\,|\, \omega]\big|\cdot \sup_{\omega} \big|\mathbb E_{X\times Y}[U_1\,|\, \omega]\big|\Big]\\
		&=\sup_{\omega} \big|\mathbb E_{X\times Y}[U_1\,|\, \omega]\big|\cdot \mathbb E_\omega\Big[\big|\mathbb E_{X\times Y}[U_1\,|\, \omega]\big|\Big].
	\end{aligned}
	\end{equation}
	Recall the definition of $U_1$ in Equation \eqref{U_1}, and we obtain
	\begin{equation*}    
    \begin{aligned}
        &\mathbb E_{X\times Y}[U_1\,|\, \omega]\\=&\mathbb E_{X,X'}[ \langle {\psi_{\omega}}(X),{\psi_{\omega}}(X')\rangle]-2\mathbb E_{X,Y}[ \langle {\psi_{\omega}}(X),{\psi_{\omega}}(Y)\rangle]+\mathbb E_{Y,Y'}[ \langle {\psi_{\omega}}(Y),{\psi_{\omega}}(Y')\rangle]\\
        =&\langle \mathbb E_X[{\psi_{\omega}}(X)],\mathbb E_{X'}[{\psi_{\omega}}(X')]\rangle-2\langle \mathbb E_X[{\psi_{\omega}}(X)],\mathbb E_{Y}[{\psi_{\omega}}(Y)]\rangle+\langle \mathbb E_Y[{\psi_{\omega}}(Y)],\mathbb E_{Y'}[{\psi_{\omega}}(Y')]\rangle\\
        =&\big\|\mathbb E_X[{\psi_{\omega}}(X)]-\mathbb E_Y[{\psi_{\omega}}(Y)]\big\|^2_2\\
        \geq &0.
    \end{aligned}
	\end{equation*}
		Thus we get
	\begin{align*}
	&\mathbb E_\omega\Big[\big|\mathbb E_{X\times Y}[U_1\,|\, \omega]\big|\Big]\\=&\mathbb E_\omega\Big[\mathbb E_{X\times Y}[U_1\,|\, \omega]\Big]\\
	=&\mathbb E_{X\times Y}\Big[\mathbb E_\omega\big[U_1\,|\,\mathcal X_{n_1},\mathcal Y_{n_2}\big]\Big]\\
	=&\mathbb E_{X\times Y}\bigg[\frac{1}{{n_1}({n_1}-1)} \sum_{1\leq i\neq j \leq {n_1}} k(X_i,X_j)-\frac{2}{{n_1} {n_2}} \sum_{i=1}^{n_1} \sum_{j=1}^{n_2} k(X_i,Y_j)+\frac{1}{{n_2}({n_2}-1)} \sum_{1\leq i\neq j \leq {n_2}}  k(Y_i,Y_j)\bigg]\\
	=&\mathrm{MMD}^2\left(P_X,P_Y;\mathcal H_{k_\lambda}\right).
	\end{align*}
	Therefore, note that Equation \eqref{eq: modification} can be further upper bounded by
	\begin{align*}
    \mathbb E_\omega\big[\big(\mathbb E_{X\times Y}[U_1\,|\, \omega]\big)^2\big]&\leq \sup_{\omega} \big|\mathbb E_{X\times Y}[U_1\,|\, \omega]\big|\cdot \mathbb E_\omega\Big[\big|\mathbb E_{X\times Y}[U_1\,|\, \omega]\big|\Big]\\
    &\leq \sup_{\omega} \big|\mathbb E_{X\times Y}[U_1\,|\, \omega]\big|\cdot \mathrm{MMD}^2\left(P_X,P_Y;\mathcal H_{k_\lambda}\right)\\
    &\leq \|\xi\|^2_2\sup_{\omega} \big|\mathbb E_{X\times Y}[U_1\,|\, \omega]\big|,
	\end{align*}
	provided that $\mathrm{MMD}^2\left(P_X,P_Y;\mathcal H_{k_\lambda}\right)\leq \|\xi\|^2_2$ by Young's convolution inequality.
	
	Note that $\mathbb E_{X\times Y}[U_1\,|\, \omega ]$ can be written as
	\begin{align*}
		\mathbb E_{X\times Y}[U_1\,|\, \omega ]=\iint_{[-M_3,M_3]^d\times [-M_3,M_3]^d}\kappa_{\lambda}(0)\cos\left(\omega^{\top}(x-y)\right)\left(p_X(x)-p_Y(x)\right)\left(p_X(y)-p_Y(y)\right)dxdy.
	\end{align*}
	Therefore, for some positive constant $C_4(M_3,d),$ we have
	\begin{align*}
    &\quad\sup_{\omega} \big|\mathbb E_{X\times Y}[U_1\,|\, \omega]\big|\\
	&=\sup_{\omega}\bigg|\iint_{[-M_3,M_3]^d\times [-M_3,M_3]^d}\kappa_{\lambda}(0)\cos\left(\omega^{\top}(x-y)\right)\left(p_X(x)-p_Y(x)\right)\left(p_X(y)-p_Y(y)\right)dxdy\bigg|\\
    &\leq \sup_{\omega}\iint_{[-M_3,M_3]^d\times [-M_3,M_3]^d}\left|\kappa_{\lambda}(0)\cos\left(\omega^{\top}(x-y)\right)\right|\left|p_X(x)-p_Y(x)\right|\left|p_X(y)-p_Y(y)\right|dxdy\\
    &\leq \kappa_{\lambda}(0)\bigg(\iint_{[-M_3,M_3]^d\times [-M_3,M_3]^d}\left|p_X(x)-p_Y(x)\right|\left|p_X(y)-p_Y(y)\right|dxdy\bigg)\\
    &=\kappa_{\lambda}(0) \bigg(\int_{[-M_3,M_3]^d} \left|p_X(x)-p_Y(x)\right|dx\bigg)^2\\
    &\stackrel{(*)}{\leq} C_4(M_3,d)\kappa_{\lambda}(0) \bigg(\int_{[-M_3,M_3]^d} \left|p_X(x)-p_Y(x)\right|^2dx\bigg)\\
    &=C_4(M_3,d)\kappa_{\lambda}(0)\|\xi\|^2_2,
	\end{align*}
	where $(*)$ follows from Cauchy-Schwarz inequality,
	and this implies
	$$
	\mathbb E_\omega\big[\big(\mathbb E_{X\times Y}[U_1\,|\, \omega]\big)^2\big]\leq C_4(M_3,d)\kappa_{\lambda}(0)\|\xi\|^4_2.
	$$

	Hence, we can get
	\begin{equation}\label{uvare}
		\begin{aligned}
			\sqrt{\frac4{\beta R}\mathbb E_{\omega}\big[\big(\mathbb E_{X\times Y}[U_1\,|\, \omega ]\big)^2\big]} &\leq \sqrt {\frac{4C_4(M_3,d)\kappa_{\lambda}(0)}{\beta R}}\|\xi\|^2_2\\
			&= C_4'(M_3,\beta,d)\sqrt{\frac{\kappa_{\lambda}(0)}{R}}\|\xi\|^2_2
		\end{aligned}
	\end{equation}
	for $C_4'(M_3,\beta,d):=2\sqrt {{C_4(M_3,d)}/\beta}.$
	
	For the final term, $\sqrt{\frac{4}{\beta}\operatorname{Var}\big[\widehat {\mathrm{MMD}}_u^2( \mathcal{X}_{n_1}, \mathcal{Y}_{n_2};\mathcal H_{k_\lambda}) \big]},$ \citet[Proposition 3]{schrab2023mmdaggregated} guarantees that there exists a positive constant $C_5(M_2,d)$ such that
	\begin{align*}
		\operatorname{Var}\big[\widehat {\mathrm{MMD}}_u^2( \mathcal{X}_{n_1}, \mathcal{Y}_{n_2};\mathcal H_{k_\lambda}) \big]\leq C_5(M_2,d) \bigg(\frac{\|\xi\ast\kappa_\lambda\|^2_2}{n}+\frac{\kappa_{\lambda}(0)}{{n}^2}\bigg).
	\end{align*}
	Then, similar to the proof of \citet[Appendix E.5]{schrab2023mmdaggregated}, we have
	\begin{equation}\label{mmduvar}
		\begin{aligned}
			\sqrt{\frac{4}{\beta}\operatorname{Var}\big[\widehat {\mathrm{MMD}}_u^2( \mathcal{X}_{n_1}, \mathcal{Y}_{n_2};\mathcal H_{k_\lambda}) \big]}&\leq \sqrt{\frac{4C_5(M_2,d)\|\xi\ast\kappa_\lambda\|^2_2}{\beta  {n}}+\frac{4C_5(M_2,d)\kappa_{\lambda}(0)}{\beta {n}^2}}\\
			&\stackrel{(a)}{\leq} 2\sqrt{\frac12\|\xi\ast\kappa_\lambda\|^2_2\frac{2C_5(M_2,d)}{\beta {n}}}+\frac{2\sqrt{C_5(M_2,d)\kappa_{\lambda}(0)}}{\sqrt{\beta}{n}}\\ 
			&\stackrel{(b)}\leq \frac12\|\xi\ast\kappa_\lambda\|^2_2+\frac{2C_5(M_2,d)}{\beta {n}}+\frac{2\sqrt{C_5(M_2,d)\kappa_{\lambda}(0)}}{\sqrt{\beta}{n}}\\
			&\leq\frac12\|\xi\ast\kappa_\lambda\|^2_2+\frac{C_5'(M_2,\beta,d)}{n}+C_5'(M_2,\beta,d)\frac{\sqrt{\kappa_{\lambda}(0)}}{{n}},
		\end{aligned}
	\end{equation}
	where $(a)$ used the fact that $\sqrt{x+y}\leq \sqrt x+ \sqrt y$ for all $x,y>0,$ $(b)$ used $2\sqrt{xy}\leq x+y$ for all $x,y>0,$ and the last inequality holds with $C_5'(M_2,\beta,d):=\max\{2C_5(M_2,d)/\beta,2\sqrt{C_5(M_2,d)/\beta}\}$.
	
	To sum up, given Equations \eqref{uvardecom}, \eqref{uevar}, \eqref{uvare} and \eqref{mmduvar}, a valid upper bound for $\sqrt{\frac4\beta\operatorname{Var}\left[U\right]}$ is
	\begin{equation}\label{ubvar}
		\begin{aligned}
			\sqrt{\frac4\beta\operatorname{Var}\left[U\right]}&\leq \sqrt{\frac4{\beta R}\mathbb E_{\omega}\big[\operatorname{Var}_{X\times Y}[U_1\,|\, \omega ]\big]}+\sqrt{\frac4{\beta R}\mathbb E_{\omega}\big[\big(\mathbb E_{X\times Y}[U_1\,|\, \omega ]\big)^2\big]}\\
			&\quad +\sqrt{\frac4\beta\operatorname{Var}_{X\times Y}\big[\widehat{\mathrm{MMD}}_u^2(\mathcal X_{n_1}, \mathcal Y_{n_2};\mathcal H_{k_\lambda}) \big]}\\
			&\leq C_2''(\beta)\sqrt{\frac{\kappa_{\lambda}(0)}{R{n}}}\|\xi\|_2+C_3''(\beta)\frac{\kappa_{\lambda}(0)}{\sqrt R{n}}+C_4'(M_3,\beta,d)\sqrt{\frac{\kappa_{\lambda}(0)}{R}}\|\xi\|^2_2\\
			&\quad+\frac12\|\xi\ast\kappa_\lambda\|^2_2+\frac{C_5'(M_2,\beta,d)}{n}+C_5'(M_2,\beta,d)\frac{\sqrt{\kappa_{\lambda}(0)}}{{n}}.
		\end{aligned}
	\end{equation}

	\subsubsection*{Upper bound for $\sqrt{\frac2{\alpha\beta}\mathbb E\big[\operatorname{Var}_\pi[U_\pi\,|\,\mathcal X_{n_1},\mathcal Y_{n_2},\boldsymbol{\omega}_R]\big]}$}
	
	Since the U-statistic is centered at zero under the permutation law, we have 
	\begin{align*}
		\operatorname{Var}_\pi[U_\pi\,|\,\mathcal X_{n_1},\mathcal Y_{n_2},\boldsymbol{\omega}_R] &= \mathbb E_\pi\big[(U_\pi)^2\,\big|\,\mathcal X_{n_1},\mathcal Y_{n_2},\boldsymbol{\omega}_R\big]-\big(\mathbb E_\pi[U_\pi\,|\,\mathcal X_{n_1},\mathcal Y_{n_2},\boldsymbol{\omega}_R]\big)^2\\
		&=\mathbb E_\pi\big[(U_\pi)^2\,\big|\,\mathcal X_{n_1},\mathcal Y_{n_2},\boldsymbol{\omega}_R\big].
	\end{align*}
	Recall Equation \eqref{kim appendix f} and note that the following result holds true \citep[Appendix F]{Kim2022}:
	\begin{equation*}
     \begin{aligned}
		&\mathbb E_\pi\big[(U_\pi)^2\,|\,\mathcal X_{n_1},\mathcal Y_{n_2},\boldsymbol{\omega}_R\big] \\=&\frac{1}{{n_1}^2({n_1}-1)^2{n_2}^2({n_2}-1)^2} \sum_{(i_1,\dots,j'_2)\in\mathbf I}\mathbb E_\pi\Big[\hat h\big(Z_{\pi(i_1)},Z_{\pi(i_2)}; Z_{\pi(n_1+j_1)},Z_{\pi(n_1+j_2)}\big)\\& 
	 \,\qquad \qquad \qquad \qquad \qquad \qquad \qquad \qquad \times \hat h\big(Z_{\pi(i'_1)},Z_{\pi(i'_2)}; Z_{\pi(n_1+j'_1)},Z_{\pi(n_1+j'_2)}\big)\,\Big|\,\mathcal X_{n_1},\mathcal Y_{n_2},\boldsymbol{\omega}_R\Big],
	\end{aligned}
 \end{equation*}
	Also, it can be shown that there exists some positive constant $C_6$ such that for any $(i_1,\dots,j_2')\in\mathbf I,$
	\begin{align*}
		\bigg|\mathbb E_{X\times Y\times \omega}\bigg[\mathbb E_\pi&\Big[\hat h\big(Z_{\pi(i_1)},Z_{\pi(i_2)}; Z_{\pi({n_1}+j_1)},Z_{\pi({n_1}+j_2)}\big)\\&\times \hat h\big(Z_{\pi(i'_1)},Z_{\pi(i'_2)}; Z_{\pi({n_1}+j'_1)},Z_{\pi({n_1}+j'_2)}\big)\,\Big|\,\mathcal X_{n_1},\mathcal Y_{n_2},\boldsymbol{\omega}_R\Big]\bigg]\bigg|\\
		&\leq C_6{\tilde\sigma}^2_{2,2},
	\end{align*}
	where
	$$
	{\tilde\sigma}^2_{2,2}:=\max\Big\{\mathbb E\big[\hat k^2(X_1,X_2)\big],\mathbb E\big[\hat k^2(X_1,Y_1)\big],\mathbb E\big[\hat k^2(Y_1,Y_2)\big]\Big\}.
	$$
	Observe that
	\begin{align*}   
		\hat k^2(x,y)&= \bigg(\frac1R \sum\limits_{r=1}^R\langle \psi_{\omega_r}(x),\psi_{\omega_r}(y)\rangle\bigg)^2\\
		&=\frac{1}{R^2}\sum^R_{r=1}\langle \psi_{\omega_r}(x),\psi_{\omega_r}(y)\rangle\langle \psi_{\omega_r}(x),\psi_{\omega_r}(y)\rangle+\frac{1}{R^2}\sum_{1\leq r_1\neq r_2 \leq R}\langle \psi_{\omega_{r_1}}(x),\psi_{\omega_{r_1}}(y)\rangle\langle \psi_{\omega_{r_2}}(x),\psi_{\omega_{r_2}}(y)\rangle.
	\end{align*}
	Therefore, we have
	\begin{align*}   
		\mathbb E\big[\hat k^2(X_1,X_2)\big]&=\mathbb E_{{X_1}\times {X_2}}\Big[\mathbb E_{\omega}\big[\hat k^2(x_1,x_2)\,\big|\,X_1=x_1,X_2=x_2 \big]\Big]\\
		&=\mathbb E_{{X_1}\times {X_2}}\bigg[\frac1R\mathbb E_{\omega}\big[\langle \psi_{\omega}(x_1),\psi_{\omega}(x_2)\rangle^2\,\big|\,X_1=x_1,X_2=x_2 \big]+\frac{R(R-1)}{R^2}k^2(X_1,X_2)\bigg]\\
		&=\frac1R\mathbb E_{{X_1}\times {X_2}}\Big[\mathbb E_{\omega}\big[\langle \psi_{\omega}(x_1),\psi_{\omega}(x_2)\rangle^2\,\big|\,X_1=x_1,X_2=x_2 \big]\Big]+\frac{(R-1)}{R}\mathbb E_{{X_1}\times {X_2}}\big[k^2(X_1,X_2)\big]\\
		&\leq\frac{\kappa_{\lambda}(0)^2}R+M_2\varkappa\kappa_{\lambda}(0),
	\end{align*}
	where the last inequality follows from the fact that $-\kappa_{\lambda}(0)\leq \langle\psi_{\omega}(x),\psi_{\omega}(y)\rangle \leq \kappa_{\lambda}(0) $ for all $x,y\in \mathbb R^d,$ and $\mathbb E_{{X_1}\times {X_2}}\left[k^2(X_1,X_2)\right] \leq M_2\varkappa(\lambda_1\cdots\lambda_d)^{-1}$ where $\varkappa=\prod^d_{i=1}\int_{\mathbb R}\kappa_i(x_i)^2\text{d}x_i$, as shown in \citet[Appendix E.3]{schrab2023mmdaggregated}. A similar calculation shows that $\mathbb E\big[\hat k^2(X_1,Y_1)\big]$ and $\mathbb E\big[\hat k^2(Y_1,Y_2)\big]$ are also upper bounded by the bound in the above inequality, thus we get
	$$
	{\tilde\sigma}^2_{2,2}\leq\frac{\kappa_{\lambda}(0)^2}R+M_2\varkappa\kappa_{\lambda}(0).
	$$
	Using this observation and counting the number of $\mathbf I$ \citep[Appendix F]{Kim2022} yields
	\begin{align*}
		\mathbb E\big[\operatorname{Var}_\pi[U_\pi\,|\,\mathcal X_{n_1},\mathcal Y_{n_2},\boldsymbol{\omega}_R]\big]&\leq C_6{\tilde\sigma}^2_{2,2}\times \frac{1}{{n_1^2}({n_1}-1)^2{n_2^2}({n_2}-1)^2}\sum_{(i_1,\dots,j'_2)\in\mathbf I} 1 \\
		&\leq C_{7}\frac{\kappa_{\lambda}(0)^2}R\bigg(\frac1{{n_1}}+\frac1{{n_2}}\bigg)^2+C_{7}M_2\varkappa\kappa_{\lambda}(0)\bigg(\frac1{{n_1}}+\frac1{{n_2}}\bigg)^2\\
		&\leq C_{7}'\frac{\kappa_{\lambda}(0)^2}{R{n}^2}+C_{7}'M_2\varkappa\frac{\kappa_{\lambda}(0)}{{n}^2}
	\end{align*}
	for some positive constant $C_{7},C_7'>0.$ Therefore, using $\sqrt{x+y}\leq \sqrt x +\sqrt y$ for all $x,y\geq0,$ we get
	\begin{equation}\label{ubcrit}
		\begin{aligned}
			\sqrt{\frac{2}{\alpha\beta}\mathbb E\big[\operatorname{Var}_\pi[U_\pi\,|\,\mathcal X_{n_1},\mathcal Y_{n_2},\boldsymbol{\omega}_R]\big]} \leq C_8(\alpha,\beta)\frac{\kappa_{\lambda}(0)}{\sqrt R{n}}+C_9(M_2,\alpha,\beta)\frac{\sqrt{\kappa_{\lambda}(0)}}{{n}}
		\end{aligned}
	\end{equation}
	for some positive constants $C_8(\alpha,\beta),C_9(M_2,\alpha,\beta)>0$.
	
	\subsubsection*{Upper bound for $\mathbb E[W']$}
	Recall that $W'$ is defined as
	$$
	W'=\frac{1}{{n_1^2}({n_1}-1)}\sum_{1\leq i\neq j\leq {n_1}}\hat k(X_i,X_j)+\frac{1}{{n_2^2}({n_2}-1)}\sum_{1\leq i\neq j\leq {n_2}}\hat k(Y_i,Y_j)
	$$
	and its expectation is
	$$
	\mathbb E[W']=\frac{1}{n_1}\mathbb E_{X_1 \times X_2}\big[k(X_1,X_2)\big]  +\frac1{{n_2}}\mathbb E_{Y_1\times Y_2}\big[k(Y_1,Y_2)\big].
	$$
	Here we observe that
    \begin{align*}
    \frac{1}{n_1}\mathbb E_{X_1\times X_2}[k_\lambda(X_1,X_2)]
    &=\frac1{n_1}\int\int p_X(x_1)p_X(x_2)k_\lambda(x_1,x_2)\,dx_1dx_2\\
    &\leq \frac{\|p_X\|_\infty}{n_1}\int p_X(x_1)\left(\int k_\lambda(x_1,x_2)\,dx_2\right)dx_1\\
    &=\frac{\|p_X\|_\infty}{n_1}\int p_X(x_1)\,dx_1\\
    &\leq \frac{M_2}{n_1}.
    \end{align*}
	and similarly we have ${n_2}^{-1}\mathbb E_{Y_1\times Y_2}\big[k(Y_1,Y_2)\big]\leq{M_2}{n_2}^{-1}.$
	Therefore, we conclude that
	\begin{equation}
		\begin{aligned}\label{EW'}
			\mathbb E[W']&\leq M_2\bigg(\frac1{{n_1}}+\frac1{{n_2}}\bigg)\\
			&\leq \frac{C_{10}(M_2)}{n}
		\end{aligned}
	\end{equation}
	for some constant $C_{10}(M_2)>0$.
	
	\subsubsection*{Upper bound for $\sqrt{\frac4\beta \operatorname{Var}[W']}$}
	We note that the variance of $W'$ can be upper bounded as
	\begin{align*}
		\operatorname{Var}[W']&=\operatorname{Var}\bigg[\frac{1}{{n_1^2}({n_1}-1)}\sum_{1\leq i\neq j\leq {n_1}}\hat k(X_i,X_j)+\frac{1}{{n_2^2}({n_2}-1)}\sum_{1\leq i\neq j\leq {n_2}}\hat k(Y_i,Y_j)\bigg]\\
		&\leq\frac{2}{{n_1^2}}\operatorname{Var}\bigg[\frac{1}{{n_1}({n_1}-1)}\sum_{1\leq i\neq j\leq {n_1}}\hat k(X_i,X_j)\bigg]+\frac{2}{{n_2^2}}\operatorname{Var}\bigg[\frac{1}{{n_2}({n_2}-1)}\sum_{1\leq i\neq j\leq {n_2}}\hat k(Y_i,Y_j)\bigg].
	\end{align*}
	Moreover, recall $\hat k(x,y):= R^{-1} \sum_{r=1}^R\langle \psi_{\omega_r}(x),\psi_{\omega_r}(y)\rangle
		=\langle \boldsymbol{\psi}_{\boldsymbol{\omega}_R}(x),\boldsymbol{\psi}_{\boldsymbol{\omega}_R}(y)\rangle.$ And then, for some positive constant $C_{11}>0$, we also have
	{\allowdisplaybreaks
	\begin{align*}
		\operatorname{Var}\bigg[\frac{1}{{n_1}({n_1}-1)}\sum_{1\leq i\neq j\leq {n_1}}\hat k(X_i,X_j)\bigg]&=\mathbb E_{X\times Y}\Bigg[\operatorname{Var}_\omega\bigg[\frac{1}{{n_1}({n_1}-1)}\sum_{1\leq i\neq j\leq {n_1}}\hat k(X_i,X_j)\,\bigg|\,\mathcal X_{n_1}\bigg]\Bigg]\\
		&\quad+\operatorname{Var}_{X\times Y}\Bigg[\mathbb E_\omega\bigg[\frac{1}{{n_1}({n_1}-1)}\sum_{1\leq i\neq j\leq {n_1}}\hat k(X_i,X_j)\,\bigg|\,\mathcal X_{n_1}\bigg]\Bigg]\\
		&=\mathbb E_{X\times Y}\Bigg[\frac1R\operatorname{Var}_\omega\bigg[\frac{1}{{n_1}({n_1}-1)}\sum_{1\leq i\neq j\leq {n_1}}\langle \psi_{\omega}(X_i),\psi_{\omega}(X_j)\rangle\,\bigg|\,\mathcal X_{n_1}\bigg]\Bigg]\\
		&\quad+\operatorname{Var}_{X\times Y}\Bigg[\mathbb E_\omega\bigg[\frac{1}{{n_1}({n_1}-1)}\sum_{1\leq i\neq j\leq {n_1}}\hat k(X_i,X_j)\,\bigg|\,\mathcal X_{n_1}\bigg]\Bigg]\\
		&\leq\frac1R\mathbb E_{X\times Y}\Bigg[\mathbb E_\omega\bigg[\bigg(\frac{1}{{n_1}({n_1}-1)}\sum_{1\leq i\neq j\leq {n_1}}\langle \psi_{\omega}(X_i),\psi_{\omega}(X_j)\rangle\bigg)^2\,\bigg|\,\mathcal X_{n_1}\bigg]\Bigg]\\
		&\quad+\operatorname{Var}_{X\times Y}\bigg[\frac{1}{{n_1}({n_1}-1)}\sum_{1\leq i\neq j\leq {n_1}}k(X_i,X_j)\bigg]\\
		&\leq\frac{\kappa_\lambda(0)^2}R+\operatorname{Var}_{X\times Y}\bigg[\frac{1}{{n_1}({n_1}-1)}\sum_{1\leq i\neq j\leq {n_1}}k(X_i,X_j)\bigg]\\
		&\leq\frac{\kappa_\lambda(0)^2}R+C_{11}\frac{\kappa_\lambda(0)}{n_1},
	\end{align*}
	}
	where the first inequality follows from $|\langle \psi_{\omega}(x),\psi_{\omega}(y)\rangle|\leq \kappa_\lambda(0)$ for all $x,y\in\mathbb R^d,$ and the last inequality follows from the result in \citet[Appendix E.11]{kim2023diff}. In a similar manner, we can get
	\begin{align*}
		\operatorname{Var}\bigg[\frac{1}{{n_2}({n_2}-1)}\sum_{1\leq i\neq j\leq {n_2}}\hat k(Y_i,Y_j)\bigg]\leq\frac{\kappa_\lambda(0)^2}R+C_{11}\frac{\kappa_\lambda(0)}{n_2}.
	\end{align*}
	Therefore, using $\sqrt{x+y}\leq \sqrt x +\sqrt y$ for all $x,y\geq0,$ we conclude that
	\begin{equation}\label{VarW'}
		\begin{aligned}
			&\quad\sqrt{\frac4\beta\operatorname{Var}[W']}\\
			&\leq\sqrt{\frac{8}{\beta {n_1^2}}\operatorname{Var}\bigg[\frac{1}{{n_1}({n_1}-1)}\sum_{1\leq i\neq j\leq {n_1}}\hat k(X_i,X_j)\bigg]}+\sqrt{\frac{8}{\beta {n_2^2}}\operatorname{Var}\bigg[\frac{1}{{n_2}({n_2}-1)}\sum_{1\leq i\neq j\leq {n_2}}\hat k(Y_i,Y_j)\bigg]}\\
			&\leq \frac{\sqrt 8 \kappa_\lambda(0)}{\sqrt{\beta R}}\bigg(\frac1{{n_1}}+\frac1{{n_2}}\bigg)+\sqrt{\frac{8C_{11} \kappa_\lambda(0)}{\beta}}\bigg(\frac1{{n_1^{3/2}}}+\frac1{{n_2^{3/2}}}\bigg)\\
			&\leq C_{12}(\beta)\frac{\kappa_\lambda(0)}{\sqrt{R}{n}}+C_{13}(\beta)\frac{\sqrt{\kappa_\lambda(0)}}{{n}^{3/2}},
		\end{aligned}
	\end{equation}
	for some positive constants $C_{12}(\beta),C_{13}(\beta)>0.$
	
	\subsubsection*{Sufficient condition for Equation \eqref{twomoment}}
	Recall that Equation \eqref{twomoment},
	\begin{align*}
		\mathbb E\left[U\right] \geq \sqrt{\frac4\beta\operatorname{Var}\left[U\right]}+\sqrt{\frac2{\alpha\beta}\mathbb E\big[\operatorname{Var}_\pi[U_\pi\,|\,\mathcal X_{n_1},\mathcal Y_{n_2},\boldsymbol{\omega}_R]\big]}+\mathbb E\left[W'\right]+\sqrt{\frac4\beta\operatorname{Var}\left[W'\right]}+C'_0\frac{\kappa_\lambda(0)}{{n}^2},
	\end{align*}
	is a sufficient condition for $\mathbb P(\mathcal B_{\beta/2})\geq 1-\beta/2.$ So far, in Equations \eqref{lbexp}, \eqref{ubvar}, \eqref{ubcrit}, \eqref{EW'} and \eqref{VarW'},  we derived a lower bound for the left-hand side of the inequality, and upper bounds for the terms in the right-hand side of the inequality as follows:
	\begin{equation*}
		\begin{aligned}
			\mathbb E\left[U\right]&\geq \frac {1-S^2}2\|\xi\|^2_2+\frac12 \|\xi\ast\kappa_\lambda\|^2_2-C_1'(M_1,d,s)\sum^d_{i=1}\lambda_i^{2s},\\
			\sqrt{\frac4\beta\operatorname{Var}\left[U\right]}&\leq C_2''(\beta)\sqrt{\frac{\kappa_{\lambda}(0)}{R{n}}}\|\xi\|_2+C_3''(\beta)\frac{\kappa_{\lambda}(0)}{\sqrt R{n}}+C_4'(M_3,\beta,d)\sqrt{\frac{\kappa_{\lambda}(0)}{R}}\|\xi\|^2_2\\
			&\quad+\frac12\|\xi\ast\kappa_\lambda\|^2_2+\frac{C_5'(M_2,\beta,d)}{n}+C_5'(M_2,\beta,d)\frac{\sqrt{\kappa_{\lambda}(0)}}{{n}}, \\
			\sqrt{\frac{2}{\alpha\beta}\mathbb E\big[\operatorname{Var}_\pi[U_\pi\,|\,\mathcal X_{n_1},\mathcal Y_{n_2},\boldsymbol{\omega}_R]\big]} &\leq C_8(\alpha,\beta)\frac{\kappa_{\lambda}(0)}{\sqrt R{n}}+C_9(\alpha,\beta,M_2)\frac{\sqrt{\kappa_{\lambda}(0)}}{{n}},\\
			\mathbb E[W']&\leq \frac{C_{10}(M_2)}{n},\\
			\sqrt{\frac4\beta\operatorname{Var}[W']}&\leq C_{12}(\beta)\frac{\kappa_\lambda(0)}{\sqrt{R}{n}}+C_{13}(\beta)\frac{\sqrt{\kappa_\lambda(0)}}{{n}^{3/2}}.
		\end{aligned}
	\end{equation*}
	
	Plugging these results into Equation \eqref{twomoment}, a sufficient condition for Equation \eqref{twomoment} is
	\begin{align*}
		\frac {1-S^2}2\|\xi\|^2_2-C_1'(M_1,d,s)\sum^d_{i=1}\lambda_i^{2s} &\geq C_4'(M_3,\beta,d)\sqrt{\frac{\kappa_{\lambda}(0)}{R}}\|\xi\|^2_2+C_2''(\beta)\sqrt{\frac{\kappa_{\lambda}(0)}{R{n}}}\|\xi\|_2\\
		&\quad +C_3''(\beta)\frac{\kappa_{\lambda}(0)}{\sqrt R{n}}+C_8(\alpha,\beta)\frac{\kappa_{\lambda}(0)}{\sqrt R{n}}+C_{12}(\beta)\frac{\kappa_\lambda(0)}{\sqrt{R}{n}}\\
		&\quad +C_5'(M_2,\beta,d)\frac{\sqrt{\kappa_{\lambda}(0)}}{{n}}+C_9(M_2,\alpha,\beta)\frac{\sqrt{\kappa_{\lambda}(0)}}{{n}}+C_{13}(\beta)\frac{\sqrt{\kappa_\lambda(0)}}{{n}^{3/2}}\\
		&\quad +\frac{C_5'(M_2,\beta,d)}{n}+\frac{C_{10}(M_2)}{n}+\frac{C'_0\kappa_\lambda(0)}{{n}^2}.
	\end{align*}
	
	Recall that $\kappa_\lambda(0)=(\lambda_1\cdots\lambda_d)^{-1},$ and suppose that $\lambda_1\cdots\lambda_d\leq 1.$ This assumption does not compromise our analysis, as ultimately $\lambda_1,\dots,\lambda_d$ we choose later satisfies this assumption. Now, observe that $\lambda_1\dots\lambda_d\leq1$ implies ${n}^{-1}\leq {n}^{-1}(\lambda_1\cdots\lambda_d)^{-1/2}.$ Also note that ${n}\leq {n}^{3/2}$ for ${n}\geq1$. Then, by grouping similar terms, a sufficient condition for the above inequality is
	\begin{align*}
		\frac {1-S^2}2\|\xi\|^2_2 &\geq \frac{C_4'(M_3,\beta,d)}{\sqrt{R\lambda_1\cdots\lambda_d}}\|\xi\|^2_2+\frac{C_2''(\beta)}{\sqrt{R{n}\lambda_1\cdots\lambda_d}}\|\xi\|_2\\
		&\quad +\frac{C_{14}(\alpha,\beta)}{\sqrt{R}{n}\lambda_1\cdots\lambda_d}+\frac{C_{15}(M_2,\alpha,\beta,d)}{{n}\sqrt{\lambda_1\cdots\lambda_d}}+\frac{C'_0}{{n}^2\lambda_1\cdots\lambda_d}+C_1'(M_1,d,s)\sum^d_{i=1}\lambda_i^{2s}.
	\end{align*}
	We observe that the simultaneous satisfaction of the following four inequalities is a sufficient condition for the above inequality:
	\begin{align*}
		\text{(i)}:\quad\|\xi\|^2_2 &\geq \frac{4C_4'(M_3,\beta,d)}{\sqrt{R\lambda_1\cdots\lambda_d}}\|\xi\|^2_2,\\
		\text{(ii)}:\quad\|\xi\|^2_2&\geq \frac{4C_2''(\beta)}{\sqrt{R{n}\lambda_1\cdots\lambda_d}}\|\xi\|_2, \\
		\text{(iii)}:\quad\|\xi\|^2_2&\geq \frac{4C_{14}(\alpha,\beta)}{\sqrt{R}{n}\lambda_1\cdots\lambda_d},\\
		\text{(iv)}:\quad\|\xi\|^2_2&\geq \frac{4C_{15}(M_2,\alpha,\beta,d)}{{n}\sqrt{\lambda_1\cdots\lambda_d}}+\frac{4C'_0}{{n}^2\lambda_1\cdots\lambda_d}+4C_{1}'(M_1,d,s)\sum^d_{i=1}\lambda_i^{2s}.
	\end{align*}
	Now, we simplify the above inequalities to facilitate our discussion. Note that the inequality (i) is equivalent to the inequality denoted as $(a)$:
	\begin{alignat*}{3}
		\text{(i)}:&\quad &\|\xi\|^2_2 &\geq \frac{4C_4'(M_3,\beta,d)}{\sqrt{R\lambda_1\cdots\lambda_d}}\|\xi\|^2_2\\
		\Longleftrightarrow\quad\text{(a)}:&\quad &R&\geq \frac{C_{16}(M_3,\beta,d)}{\lambda_1\cdots\lambda_d},
	\end{alignat*}
	where $C_{16}(M_3,\beta,d):=16C_4'(M_3,\beta,d)^2.$
	Also, observe that the inequality (ii) is equivalent to
	\begin{alignat*}{3}
		& &\|\xi\|^2_2&\geq \frac{4C_2''(\beta)}{\sqrt{R{n}\lambda_1\cdots\lambda_d}}\|\xi\|_2\\
		&\Longleftrightarrow\quad &\|\xi\|^2_2&\geq \frac{16C_2''(\beta)^2}{R{n}\lambda_1\cdots\lambda_d}.
	\end{alignat*}
	Since $R\geq1$ and $R^{-1/2}\geq R ^{-1}$, a sufficient condition, denoted as (b), for simultaneously satisfying the inequalities (ii) and (iii) is 
	\begin{align*}
		\text{(b)}:\quad\|\xi\|^2_2\geq \frac{C_{17}(\alpha,\beta)}{\sqrt R{n}\lambda_1\cdots\lambda_d}
	\end{align*}
	for $C_{17}(\alpha,\beta):=\max\{4C_{14}(\alpha,\beta),16C_2''(\beta)^2\}.$ For the inequality (iv), we note that we would like to assume $\lambda_1\cdots\lambda_d\geq {n}^{-2}.$ Again, this assumption can be made harmlessly, as we choose $\lambda_1,\dots,\lambda_d$ later to satisfy this assumption. Under this assumption, observe that the term ${n}^{-2}(\lambda_1\cdots\lambda_d)^{-1}$ is dominated by the term ${n}^{-1}(\lambda_1\cdots\lambda_d)^{-1/2}$. Therefore, the following inequality, denoting (c), is sufficient to show that the inequality (iv) holds:
	\begin{align}\label{L2final2}
		\text{(c)}:\quad\|\xi\|^2_2\geq C_{18}(M_1,M_2,\alpha,\beta,d,s)\bigg(\frac{1}{{n}\sqrt{\lambda_1\cdots\lambda_d}}+\sum^d_{i=1}\lambda_i^{2s}\bigg)
	\end{align}
	where $C_{18}(M_1,M_2,\alpha,\beta,d,s):=\max\{4C_{15}(M_2,\alpha,\beta,d)+4C'_0,4C_1'(M_1,d,s)\}.$
	
	In summary, a sufficient condition for satisfying the inequalities (i), (ii), (iii) and (iv) at once is the simultaneous satisfaction of the following three inequalities:
	\begin{alignat*}{3}
		\text{(a)}:&\quad &R&\geq \frac{C_{16}(M_3,\beta,d)}{\lambda_1\cdots\lambda_d},\\
		\text{(b)}:&\quad &\|\xi\|^2_2&\geq \frac{C_{17}(\alpha,\beta)}{\sqrt R{n}\lambda_1\cdots\lambda_d},\\
		\text{(c)}:&\quad &\|\xi\|^2_2&\geq C_{18}(M_1,M_2,\alpha,\beta,d,s)\bigg(\frac{1}{{n}\sqrt{\lambda_1\cdots\lambda_d}}+\sum^d_{i=1}\lambda_i^{2s}\bigg).
	\end{alignat*}
	Then by the definition of uniform separation rate, for both tests $\Delta=\Delta_{{n_1},{n_2},R}^{\alpha,\lambda}$ or $\Delta=\Delta_{{n_1},{n_2},R}^{\alpha,u,\lambda}$, we have
	\begin{align*}
		\rho\left(\Delta,~\beta,~\mathcal C_{L_2},~\delta_{L_2}\right)^2\leq C_{19}(M_1,M_2,\alpha,\beta,d,s)\max\bigg\{\frac{1}{\sqrt R{n}\lambda_1\cdots\lambda_d}, \frac{1}{{n}\sqrt{\lambda_1\cdots\lambda_d}}+\sum^d_{i=1}\lambda_i^{2s}\bigg\},
	\end{align*}
	for $C_{19}(M_1,M_2,\alpha,\beta,d,s):=\max\{C_{17}(\alpha,\beta),C_{18}(M_1,M_2,\alpha,\beta,d,s)\}$ and $R\geq C_{16}(M_3,\beta,d)(\lambda_1\cdots\lambda_d)^{-1}.$ For the smallest order of ${n}$ possible, we choose the bandwidth $\lambda^\star_i:={n}^{-2/(4s+d)}$ for $i=1,\ldots,d$, and in this case, the condition on $R$ becomes $R\geq C_{16}(M_3,\beta,d){n}^{2d/(4s+d)}.$ Plugging these values into the above inequality, we get
	\begin{align*}
		\rho\left(\Delta,~\beta,~\mathcal C_{L_2},~\delta_{L_2}\right)^2&\leq C_{19}(M_1,M_2,\alpha,\beta,d,s)\max\bigg\{\frac{1}{\sqrt R{n}\lambda^\star_1\cdots\lambda^\star_d}, \frac{1}{{n}\sqrt{\lambda^\star_1\cdots\lambda^\star_d}}+\sum^d_{i=1}{\lambda^\star_i}^{2s}\bigg\}\\
		&\leq C_{20}(M_1,M_2,M_3,\alpha,\beta,d,s){n}^{-{4s}/(4s+d)}
	\end{align*}
	for $C_{20}(M_1,M_2,M_3,\alpha,\beta,d,s):=C_{19}(M_1,M_2,\alpha,\beta,d,s)\max\big\{C_{16}(M_3,\beta,d)^{-1/2},d+1\big\}.$ We note that 
	our choice $\{\lambda^\star_i\}^d_{i=1}$ satisfies the condition $\lambda_1\cdots\lambda_d\leq 1$ for Equation \eqref{mmduvar} and the condition $\lambda_1\cdots\lambda_d\geq {n}^{-2}$ for \Cref{L2final2}.
	By letting $C_{L_2}(M_1,M_2,M_3,\alpha,\beta,d,s):=C_{20}(M_1,M_2,M_3,\alpha,\beta,d,s)^{1/2},$ we conclude that
	$$
	\rho\left(\Delta,\,\beta,\,\mathcal C_{L_2},\,\delta_{L_2}\right)\leq C_{L_2}(M_1,M_2,M_3,\alpha,\beta,d,s){n}^{{-2s}/{(4s+d)}}
	$$
	holds and this completes the proof.
	
	\subsection{Proof of Theorem \ref{MMDthm}} \label{Section: proof of MMDthm}
	
	We start by recalling the event defined in Equation \eqref{thm5prob_kappa}, equipped with the kernel $k$:
	\begin{align*}
		\mathcal B_{\beta}:=\bigg\{\mathbb E\left[U\right] \geq \sqrt{\frac2\beta\operatorname{Var}\left[U\right]}+q_{{n_1},{n_2},1-\alpha}^u+\kappa(0)\bigg(\sqrt{\frac{2}{ \beta}}+2\bigg)\bigg(\frac1{{n_1}}+\frac1{{n_2}}\bigg)\bigg\}.
	\end{align*}
	As shown in the proof of Theorem \ref{conthm}, to control the probability of type II error of both tests $\Delta_{{n_1},{n_2},R}^{\alpha}$ and $\Delta_{{n_1},{n_2},R}^{\alpha,u}$ simultaneously, it is sufficient to show that $\mathbb P(\mathcal B_\beta)=1.$ Similar to the proof of Theorem \ref{L2thm}, we analyze the terms in the event $\mathcal B_\beta$ and derive a sufficient condition for the event $\mathcal B_\beta$. To start with, note that we have
	\begin{equation*}
		\begin{aligned}
			\mathbb E\left[U\right]&=\mathrm{MMD}^2\left(P_X,P_Y;\mathcal H_{k}\right),\\
			q_{{n_1},{n_2},1-\alpha}^u&\leq C_1(\alpha)\kappa(0)\bigg(\frac1{{n_1}}+\frac1{{n_2}}\bigg),
		\end{aligned}
	\end{equation*}
	for some positive constant $C_1(\alpha)>0,$ from Equation \eqref{thm5lhs} and \eqref{crit.ub2}. This gives
	\begin{align*}
		q_{{n_1},{n_2},1-\alpha}^u+\kappa(0)\bigg(\sqrt{\frac{2}{ \beta}}+2\bigg)\bigg(\frac1{{n_1}}+\frac1{{n_2}}\bigg)&\leq \kappa(0)\bigg(C_1(\alpha)+\sqrt{\frac{2}{ \beta}}+2\bigg)\bigg(\frac1{{n_1}}+\frac1{{n_2}}\bigg)\\
		&\leq \frac{C_2(\alpha,\beta,K)}{n}
	\end{align*}
	for some positive constant $C_2(\alpha,\beta,K)>0$. Therefore, a sufficient condition for $\mathbb P(\mathcal B_\beta)=1$ is the following inequality:
	\begin{align}\label{mmdsepobj}
		\mathrm{MMD}^2\left(P_X,P_Y;\mathcal H_{k}\right) \geq \sqrt{\frac2\beta\operatorname{Var}\left[U\right]}+\frac{C_2(\alpha,\beta,K)}{n}.
	\end{align}
	
	\subsubsection*{Upper bound for $\sqrt{\frac2\beta\operatorname{Var}\left[U\right]}$}
	Now we analyze the square root of the variance term. Recall the decomposition of the variance of $U$ in Equation \eqref{uvardecom}:
	\begin{align*}
		\sqrt{\frac2\beta\operatorname{Var}\left[U\right]}&\leq \sqrt{\frac4{\beta R}\mathbb E_{\omega}\big[\operatorname{Var}_{X\times Y}[U_1\,|\, \omega ]\big]}+\sqrt{\frac4{\beta R}\mathbb E_{\omega}\big[\big(\mathbb E_{X\times Y}[U_1\,|\, \omega ]\big)^2\big]}\\
		&\quad +\sqrt{\frac4\beta\operatorname{Var}_{X\times Y}\big[\widehat{\mathrm{MMD}}_u^2(\mathcal X_{n_1}, \mathcal Y_{n_2};\mathcal H_k) \big]}.
	\end{align*}
	We aim to analyze each term on the right-hand side and derive an upper bound for them. First, recall the result in Equation \eqref{EVarU1} and the assumption that the kernel $k$ is uniformly bounded by $K.$ Then we have
	\begin{equation}\label{uevar2}
		\begin{aligned}
			\sqrt{\frac4{\beta R}\mathbb E_{\omega}\big[\operatorname{Var}_{X\times Y}[U_1\,|\, \omega ]\big]}&\leq \frac{C_3(\beta,K)}{\sqrt{R{n}}}\mathrm{MMD}(P_X,P_Y;\mathcal H_{k})+\frac{C_4(\beta,K)}{\sqrt R{n}},
		\end{aligned}
	\end{equation}
	for some positive constants $C_3(\beta,K), C_4(\beta,K)>0.$
	
	For the term $\mathbb E_{\omega}[(\mathbb E_{X\times Y}[U_1\,|\, \omega ])^2]$, recall the statistic $U_1$ defined in \Cref{U_1} and let us denote the statistic $V$ with a single random feature as $V_1$, i.e.,
	\begin{equation*}
		\begin{aligned}
			V_1(\mathcal X_{n_1},\mathcal Y_{n_2};\omega):=&\frac{1}{{n_1^2}} \sum_{1\leq i,j \leq {n_1}} \langle {\psi_{\omega}}(X_i),{\psi_{\omega}}(X_j)\rangle-\frac{2}{{n_1} {n_2}} \sum_{i=1}^{n_1} \sum_{j=1}^{n_2} \langle {\psi_{\omega}}(X_i),{\psi_{\omega}}(Y_j)\rangle\\
			&\quad+\frac{1}{{n_2^2}} \sum_{1\leq i,j \leq {n_2}}  \langle {\psi_{\omega}}(Y_i),{\psi_{\omega}}(Y_j)\rangle\\
			=&\frac{1}{{n_1^2}} \sum_{1\leq i,j \leq {n_1}} \kappa(0)\cos\big({\omega^{\top}\left(X_i-X_j\right)}\big)-\frac{2}{{n_1} {n_2}} \sum_{i=1}^{n_1} \sum_{j=1}^{n_2} \kappa(0)\cos\big({\omega^{\top}\left(X_i-Y_j\right)}\big)\\
			&\quad+\frac{1}{{n_2^2}} \sum_{1\leq i,j \leq {n_2}} \kappa(0)\cos\big({\omega^{\top}\left(Y_i-Y_j\right)}\big).
		\end{aligned}
	\end{equation*}
	Also, let $W_1$ be the difference between $V_1$ and $U_1.$ Then, observe
	{\allowdisplaybreaks\begin{align*}
		\mathbb E_{\omega}\big[\big(\mathbb E_{X\times Y}[U_1\,|\, \omega ]\big)^2\big]&\stackrel{(a)}{\leq} \mathbb E_{\omega}\big[2\big(\mathbb E_{X\times Y}[V_1\,|\, \omega ]\big)^2+2\big(\mathbb E_{X\times Y}[-W_1\,|\, \omega ]\big)^2\big] \\
		&\stackrel{(b)}{\leq}8\kappa(0)\mathbb E_{\omega}\big[\mathbb E_{X\times Y}[V_1\,|\, \omega ]\big]+2\mathbb E_{\omega}\big[\mathbb E_{X\times Y}[W_1^2\,|\, \omega ]\big]\\
		&\stackrel{(c)}{\leq}8\kappa(0)\mathbb E_{\omega}\big[\mathbb E_{X\times Y}[V_1\,|\, \omega ]\big]+8\kappa(0)^2\bigg(\frac1{{n_1}-1}+\frac1{{n_2}-1}\bigg)^2\\
		&\leq 8\kappa(0)\Big(\mathbb E_{\omega}\big[\mathbb E_{X\times Y}[U_1\,|\, \omega ]\big]+\mathbb E_{\omega}\big[\mathbb E_{X\times Y}[W_1\,|\, \omega ]\big]\Big)+8C_0^2\frac{\kappa(0)^2}{{n}^2}\\
		&\stackrel{(d)}{\leq} 8\kappa(0)\mathrm{MMD}^2(P_X,P_Y;\mathcal H_{k})+8C_0\frac{\kappa(0)^2}{{n}}+8C_0^2\frac{\kappa(0)^2}{{n}^2}\\    
		&\leq 8K\mathrm{MMD}^2(P_X,P_Y;\mathcal H_{k})+8C_5\frac{K^2}{{n}}
	\end{align*}}
	where (a) follows from the inequality $(x+y)^2\leq 2x^2+2y^2$ for all $x,y\in \mathbb R$, (b) follows from $0\leq V_1\leq 4\kappa(0)$, (c), (d) is according to $0\leq W_1\leq\kappa(0)\Big(\frac{1}{{n_1}-1}+\frac{1}{{n_2}-1}\Big),$ and the last inequality holds with a constant $C_5:=C_0+{C_0}^2.$ Using $\sqrt{x+y}\leq\sqrt x +\sqrt y$ for all $x,y \geq 0$, we conclude that
	\begin{align}\label{eeu1}
		\sqrt{\frac4{\beta R}\mathbb E_{\omega}\big[\big(\mathbb E_{X\times Y}[U_1\,|\, \omega ]\big)^2\big]}\leq \frac{C_6(\beta,K)}{\sqrt{R}}\mathrm{MMD}(P_X,P_Y;\mathcal H_{k})+\frac{C_7(\beta,K)}{\sqrt {R{n}}},
	\end{align}
	for some positive constants $C_6(\beta,K),C_7(\beta,K)>0.$
	
	For $\operatorname{Var}_{X\times Y}\big[\widehat{\mathrm{MMD}}_u^2(\mathcal X_{n_1}, \mathcal Y_{n_2};\mathcal H_k) \big],$ we follow a similar logic to Equation \eqref{uvardecom} through Equation \eqref{uevar}. The difference is that instead of using $h_\omega$, we use the following kernel for a two-sample U-statistic:
	$$
	h(x_1,x_2;y_1,y_2):=\langle \dot k(x_1),\dot k(x_2)\rangle+\langle \dot k(y_1),\dot k(y_2)\rangle-\langle \dot k(x_1),\dot k(y_2)\rangle-\langle \dot k(x_2),\dot k(y_1)\rangle,
	$$
	where $\dot k(x)(\cdot)=k(x,\cdot).$ Note that we have $|\langle \dot k(x),\dot k(y)\rangle|=|k(x,y)|\leq \kappa(0)$ for all $x,y\in \mathbb R^d$ as shown in Equation \eqref{kbound}. This fact corresponds to the condition for the inequality (b) in \Cref{sigma10}. Also note that $\big\| \mathbb E_X[\dot k(X)]-\mathbb E_{Y}[\dot k(Y)] \big\|^2 =\mathrm{MMD}^2(P_X,P_Y;\mathcal H_{k})$ holds. This is the condition for the inequality (a) in \Cref{EVarU1}; therefore, we can follow the same logic to the previous analysis with kernel $h$ and we have a similar result to the first line in \Cref{EVarU1}:
	\begin{equation*}
		\begin{aligned}
			\operatorname{Var}_{X\times Y}\big[\widehat {\mathrm{MMD}}_u^2( \mathcal{X}_{n_1}, \mathcal{Y}_{n_2};\mathcal H_k) \big] &\leq C_8\frac{\kappa(0)}{{n}}\mathrm{MMD}^2(P_X,P_Y;\mathcal H_{k})+C_9\frac{\kappa(0)^2}{{n}^2},
		\end{aligned}
	\end{equation*}
	for some positive constants $C_8,C_9>0.$ We apply $\sqrt{x+y}\leq\sqrt x +\sqrt y$ for all $x,y \geq 0$ here and get the following result:
	\begin{equation} \label{varMMD}
		\begin{aligned}
			\sqrt{\frac4\beta\operatorname{Var}_{X\times Y}\big[\widehat{\mathrm{MMD}}_u^2(\mathcal X_{n_1}, \mathcal Y_{n_2};\mathcal H_k) \big]} &\leq \frac{C_{10}(\beta,K)}{ \sqrt{{n}}}\mathrm{MMD}(P_X,P_Y;\mathcal H_{k})+\frac{C_{11}(\beta,K)}{{n}}
		\end{aligned}
	\end{equation}
	where $C_{10}(\beta,K),C_{11}(\beta,K)>0$ are some positive constants.
	In summary, given Equations \eqref{uevar2},\eqref{eeu1} and \eqref{varMMD}, a valid upper bound for $\sqrt{\frac2\beta\operatorname{Var}\left[U\right]}$ is
	\begin{equation*}
		\begin{aligned}
			\sqrt{\frac2\beta\operatorname{Var}\left[U\right]}&\leq \sqrt{\frac4{\beta R}\mathbb E_{\omega}\big[\operatorname{Var}_{X\times Y}[U_1\,|\, \omega ]\big]}+\sqrt{\frac4{\beta R}\mathbb E_{\omega}\big[\big(\mathbb E_{X\times Y}[U_1\,|\, \omega ]\big)^2\big]}\\
			&\quad +\sqrt{\frac4\beta\operatorname{Var}_{X\times Y}\big[\widehat{\mathrm{MMD}}_u^2(\mathcal X_{n_1}, \mathcal Y_{n_2};\mathcal H_k) \big]}\\
			&\leq \bigg(\frac{C_3(\beta,K)}{\sqrt{R{n}}}+\frac{C_6(\beta,K)}{\sqrt{R}}+\frac{C_{10}(\beta,K)}{ \sqrt{{n}}}\bigg)\mathrm{MMD}(P_X,P_Y;\mathcal H_{k})\\
			&\quad+\frac{C_4(\beta,K)}{\sqrt R{n}} +\frac{C_7(\beta,K)}{\sqrt {R{n}}}+\frac{C_{11}(\beta,K)}{{n}}.
		\end{aligned}
	\end{equation*}
	
	\subsubsection*{Sufficient condition for \Cref{mmdsepobj}}
	Recall that \Cref{mmdsepobj},
	\begin{align*}
		\mathrm{MMD}^2\left(P_X,P_Y;\mathcal H_{k}\right) \geq \sqrt{\frac2\beta\operatorname{Var}\left[U\right]}+\frac{C_2(\alpha,\beta,K)}{n}
	\end{align*}
	is a sufficient condition for $\mathbb P(\mathcal B_\beta)=1.$
	Utilizing an upper bound for the variance of $U$ we derived, a sufficient condition for \Cref{mmdsepobj} to hold is
	\begin{align*}
		\mathrm{MMD}^2\left(P_X,P_Y;\mathcal H_{k}\right)&\geq \bigg(\frac{C_3(\beta,K)}{\sqrt{R{n}}}+\frac{C_6(\beta,K)}{\sqrt{R}}+\frac{C_{10}(\beta,K)}{ \sqrt{{n}}}\bigg)\mathrm{MMD}(P_X,P_Y;\mathcal H_{k})\\
		&\quad+\frac{C_4(\beta,K)}{\sqrt R{n}} +\frac{C_7(\beta,K)}{\sqrt {R{n}}}+\frac{C_{11}(\beta,K)}{{n}}+\frac{C_2(\alpha,\beta,K)}{n}.
	\end{align*}
	Note that ${n}^{-1}\leq {n}^{-1/2}\leq 1$ for ${n}\geq 1.$ Then, by merging similar terms, a sufficient condition for the above inequality is
	\begin{align*}
		\mathrm{MMD}^2\left(P_X,P_Y;\mathcal H_{k}\right)&\geq \bigg(\frac{C_{12}(\beta,K)}{\sqrt{R}}+\frac{C_{10}(\beta,K)}{ \sqrt{{n}}}\bigg)\mathrm{MMD}(P_X,P_Y;\mathcal H_{k})\\
		&\quad+\frac{C_{13}(\beta,K)}{\sqrt {R{n}}}+\frac{C_{14}(\alpha,\beta,K)}{{n}},
	\end{align*}
	for some positive constants $C_{12}(\beta,K),C_{13}(\beta,K),C_{14}(\alpha,\beta,K)>0.$
	Now the satisfaction of the following four inequalities at once is a sufficient condition for the above inequality:
	\begin{align*}
		\text{(i)}:\quad\mathrm{MMD}^2\left(P_X,P_Y;\mathcal H_{k}\right)&\geq \frac{4C_{12}(\beta,K)}{\sqrt{R}}\mathrm{MMD}(P_X,P_Y;\mathcal H_{k}),\\
		\text{(ii)}:\quad\mathrm{MMD}^2\left(P_X,P_Y;\mathcal H_{k}\right)&\geq \frac{4C_{10}(\beta,K)}{ \sqrt{{n}}}\mathrm{MMD}(P_X,P_Y;\mathcal H_{k}),\\
		\text{(iii)}:\quad\mathrm{MMD}^2\left(P_X,P_Y;\mathcal H_{k}\right)&\geq \frac{4C_{13}(\beta,K)}{\sqrt {R{n}}},\\
		\text{(iv)}:\quad\mathrm{MMD}^2\left(P_X,P_Y;\mathcal H_{k}\right)&\geq \frac{4C_{14}(\alpha,\beta,K)}{{n}}.
	\end{align*}
	Here we note that the inequalities (i) and (ii) are equivalent to the following inequalities (a) and (b), respectively:
	\begin{align*}
		\text{(a)}:\quad\mathrm{MMD}^2\left(P_X,P_Y;\mathcal H_{k}\right)&\geq \frac{16C_{12}(\beta,K)^2}{R},\\
		\text{(b)}:\quad\mathrm{MMD}^2\left(P_X,P_Y;\mathcal H_{k}\right)&\geq \frac{16C_{10}(\beta,K)^2}{{n}}.
	\end{align*}
	Considering the inequalities (a),(b),(iii) and (iv), for both tests $\Delta=\Delta_{{n_1},{n_2},R}^{\alpha}$ or $\Delta=\Delta_{{n_1},{n_2},R}^{\alpha,u}$, we have
	\begin{align*}
		\rho\big(\Delta,\,\beta,\,\mathcal C,\,\delta_{\mathrm{MMD}} \big)^2\leq C_{15}(\alpha,\beta,K)\max\bigg\{\frac{1}{R},\frac{1}{{n}},\frac{1}{\sqrt{R{n}}}\bigg\},
	\end{align*}
	for some positive constant $C_{15}(\alpha,\beta,K)>0.$
	For the smallest order of ${n}$ possible, choose $R={n}$ and then we have
	\begin{align*}
		\rho\big(\Delta,\,\beta,\,\mathcal C,\,\delta_{\mathrm{MMD}} \big)^2\leq C_{15}(\alpha,\beta,K){n}^{-1}.
	\end{align*}
	By letting $C_{\mathrm{MMD}}(\alpha,\beta,K):=C_{15}(\alpha,\beta,K)^{1/2},$ we conclude that
	$$
	\rho\big(\Delta,\,\beta,\,\mathcal C,\,\delta_{\mathrm{MMD}} \big)\leq C_{\mathrm{MMD}}(\alpha,\beta,K){n}^{-1/2}
	$$
	holds and this completes the proof.

	\subsection{Proof of Proposition \ref{MMDproposition}} \label{Section: proof of MMDproposition}
	Recall the event defined in Equation \eqref{thm5prob_kappa}:
	\begin{align*}
		\mathcal B_{\beta}:=\bigg\{\mathbb E\left[U\right] \geq \sqrt{\frac2\beta\operatorname{Var}\left[U\right]}+q_{{n_1},{n_2},1-\alpha}^u+\kappa_\lambda(0)\bigg(\sqrt{\frac{2}{ \beta}}+2\bigg)\bigg(\frac1{{n_1}}+\frac1{{n_2}}\bigg)\bigg\}
	\end{align*}
	and note that when $\mathbb P(\mathcal B_\beta)=1$, both tests $\Delta_{{n_1},{n_2},R}^{\alpha,u,\lambda}$ and $\Delta_{{n_1},{n_2},R}^{\alpha,\lambda}$ uniformly control the probability of type II error. Also, with $\kappa_\lambda(0)=\pi^{-d/2}(\lambda_1\cdots\lambda_d)^{-1}$ in place, \Cref{mmdsepobj} implies that a sufficient condition for $\mathbb P(\mathcal B_\beta)=1$ is the following inequality:
	\begin{align}\label{mmdpropositionobj}
		\mathrm{MMD}^2\left(P_X,P_Y;\mathcal H_{k_\lambda}\right) \geq \sqrt{\frac2\beta\operatorname{Var}\left[U\right]}+\frac{C_1(\alpha,\beta,d,
			\lambda)}{n}
	\end{align}
	where $C_1(\alpha,\beta,d,\lambda)>0$ is some positive constant. Similar to the proof of Theorem \ref{MMDthm}, our objective is to find an upper bound for the right-hand side of the above inequality.
	
	\subsubsection*{Upper bound for $\sqrt{\frac2\beta\operatorname{Var}\left[U\right]}$}
	Recall the decomposition in Equation \eqref{uvardecom}
	\begin{align*}
		\sqrt{\frac2\beta\operatorname{Var}\left[U\right]}&\leq \sqrt{\frac4{\beta R}\mathbb E_{\omega}\big[\operatorname{Var}_{X\times Y}[U_1\,|\, \omega ]\big]}+\sqrt{\frac4{\beta R}\mathbb E_{\omega}\big[\big(\mathbb E_{X\times Y}[U_1\,|\, \omega ]\big)^2\big]}\\
		&\quad +\sqrt{\frac4\beta\operatorname{Var}_{X\times Y}\big[\widehat{\mathrm{MMD}}_u^2(\mathcal X_{n_1}, \mathcal Y_{n_2};\mathcal H_{k_\lambda}) \big]}.
	\end{align*}
	Also, as previously noted in \Cref{uevar2} and \eqref{varMMD}, we derived upper bounds for the first term and the last term on the right-hand side of the above inequality, respectively:
	\begin{equation*}
		\begin{aligned}
			\sqrt{\frac4{\beta R}\mathbb E_{\omega}\big[\operatorname{Var}_{X\times Y}[U_1\,|\, \omega ]\big]}&\leq C_2(\beta)\sqrt{\frac{\kappa_{\lambda}(0)}{R{n}}}\mathrm{MMD}(P_X,P_Y;\mathcal H_{k_\lambda})+C_3(\beta)\frac{\kappa_{\lambda}(0)}{\sqrt R{n}},\\
			\sqrt{\frac4\beta\operatorname{Var}_{X\times Y}\big[\widehat{\mathrm{MMD}}_u^2(\mathcal X_{n_1}, \mathcal Y_{n_2};\mathcal H_{k_\lambda}) \big]} &\leq C_{4}(\beta)\sqrt{\frac{\kappa_\lambda(0)}{ {n}}}\mathrm{MMD}(P_X,P_Y;\mathcal H_{k_\lambda})+C_{5}(\beta)\frac{\kappa_\lambda(0)}{{n}},
		\end{aligned}
	\end{equation*}
	We now analyze and derive an upper bound for the remaining term, $\mathbb E_{\omega}\big[(\mathbb E_{X\times Y}[U_1\,|\, \omega ])^2\big].$ We emphasize that, unlike the result in Equation \eqref{eeu1} we stated in the proof of \Cref{MMDthm}, a stronger upper bound can be established here since we assumed a smaller class of distribution pairs, specifically a class of Gaussian distributions with a common fixed covariance, $\mathcal C_{N,\Sigma}\subsetneq \mathcal C.$ This favorable setting allows us to explicitly calculate the term $\mathbb E_{\omega}\big[(\mathbb E_{X\times Y}[U_1\,|\, \omega ])^2\big]$ and upper bound it by a higher power of the population MMD. In detail, Lemma~\ref{lemma: moment bound for Gaussian} yields 
	\begin{equation*}
		\begin{aligned}
			\mathbb E_{\omega}\big[\big(\mathbb E_{X\times Y}[U_1\,|\, \omega ]\big)^2\big] &\leq C_6(d,\lambda,\Sigma)\big(\mathbb E_{\omega}\big[\mathbb E_{X\times Y}[U_1\,|\, \omega ]\big]\big)^2\\
			&= C_6(d,\lambda,\Sigma)\mathrm{MMD}^4(P_X,P_Y;\mathcal H_{k_\lambda}),
		\end{aligned}
	\end{equation*}
	for some positive constant $C_6(d,\lambda,\Sigma)>0$. Therefore, it holds that 
	$$
	\sqrt{\frac4{\beta R}\mathbb E_{\omega}\big[\big(\mathbb E_{X\times Y}[U_1\,|\, \omega ]\big)^2\big]}\leq \frac{C_7(\beta,d,\lambda,\Sigma)}{\sqrt R}\mathrm{MMD}^2(P_X,P_Y;\mathcal H_{k_\lambda})
	$$
	We point out that there are two improvements on this upper bound, compared to the previous bound in \Cref{eeu1} that induces quadratic computational cost. Firstly, the power of $\mathrm{MMD}(P_X,P_Y;\mathcal H_{k_\lambda})$ in this upper bound is two, whereas it is one in the previous. Also, note that there is no additional term in the current bound such as $(R{n})^{-1/2}$ that exists in the previous bound. 
	
	To sum up, a valid upper bound for the square root of the variance of $U$ satisfies
	\begin{equation*}
		\begin{aligned}
			\sqrt{\frac2\beta\operatorname{Var}\left[U\right]}&\leq \sqrt{\frac4{\beta R}\mathbb E_{\omega}\big[\operatorname{Var}_{X\times Y}[U_1\,|\, \omega ]\big]}+\sqrt{\frac4{\beta R}\mathbb E_{\omega}\big[\big(\mathbb E_{X\times Y}[U_1\,|\, \omega ]\big)^2\big]}\\
			&\quad +\sqrt{\frac4\beta\operatorname{Var}_{X\times Y}\big[\widehat{\mathrm{MMD}}_u^2(\mathcal X_{n_1}, \mathcal Y_{n_2};\mathcal H_{k_\lambda}) \big]}\\
			&\leq \frac{C_7(\beta,d,\lambda,\Sigma)}{\sqrt R}\mathrm{MMD}^2(P_X,P_Y;\mathcal H_{k_\lambda})\\
			&\quad+\bigg(C_2(\beta)\sqrt{\frac{\kappa_{\lambda}(0)}{R{n}}}+C_{4}(\beta)\sqrt{\frac{\kappa_\lambda(0)}{ {n}}}\bigg)\mathrm{MMD}(P_X,P_Y;\mathcal H_{k_\lambda})+C_3(\beta)\frac{\kappa_{\lambda}(0)}{\sqrt R{n}}+C_{5}(\beta)\frac{\kappa_\lambda(0)}{{n}}\\
			&\stackrel{(\dagger)}{\leq} \frac{C_7(\beta,d,\lambda,\Sigma)}{\sqrt R}\mathrm{MMD}^2(P_X,P_Y;\mathcal H_{k_\lambda})+\frac{C_8(\beta,d,\lambda)}{ \sqrt{{n}}}\mathrm{MMD}(P_X,P_Y;\mathcal H_{k_\lambda})+\frac{C_9(\beta,d,\lambda)}{{n}}
		\end{aligned}
	\end{equation*}
	for some positive constants $C_8(\beta,d,\lambda),C_9(\beta,d,\lambda)>0,$ where the inequality $(\dagger)$ follows from $R^{-1/2}\leq 1$ for $R\geq 1$, and $\kappa_\lambda(0)=\pi^{-d/2}(\lambda_1\cdots\lambda_d)^{-1}$.
	
	\subsubsection*{Sufficient condition for \Cref{mmdpropositionobj}}
	Note that our objective is to find a sufficient condition for \Cref{mmdpropositionobj},
	\begin{align*}
		\mathrm{MMD}^2\left(P_X,P_Y;\mathcal H_{k_\lambda}\right) \geq \sqrt{\frac2\beta\operatorname{Var}\left[U\right]}+\frac{C_1(\alpha,\beta,d,\lambda)}{n}.
	\end{align*}
	Plugging the upper bound we derive in the preceding section into the above inequality, observe that a sufficient condition for \Cref{mmdpropositionobj} to hold is 
	\begin{align*}
		\mathrm{MMD}^2\left(P_X,P_Y;\mathcal H_{k_\lambda}\right) &\geq \frac{C_7(\beta,d,\lambda,\Sigma)}{\sqrt R}\mathrm{MMD}^2(P_X,P_Y;\mathcal H_{k_\lambda})+\frac{C_8(\beta,d,\lambda)}{ \sqrt{{n}}}\mathrm{MMD}(P_X,P_Y;\mathcal H_{k_\lambda})\\
		&\quad+\frac{C_9(\beta,d,\lambda)}{{n}}+\frac{C_1(\alpha,\beta,d,\lambda)}{n}.
	\end{align*}
	Now the simultaneous satisfaction of the following three inequalities is a sufficient condition for the above inequality:
	\begin{align*}
		\text{(i)}:\quad\mathrm{MMD}^2\left(P_X,P_Y;\mathcal H_{k_\lambda}\right)&\geq \frac{3C_7(\beta,d,\lambda,\Sigma)}{\sqrt R}\mathrm{MMD}^2(P_X,P_Y;\mathcal H_{k_\lambda}),\\
		\text{(ii)}:\quad\mathrm{MMD}^2\left(P_X,P_Y;\mathcal H_{k_\lambda}\right)&\geq \frac{3C_8(\beta,d,\lambda)}{ \sqrt{{n}}}\mathrm{MMD}(P_X,P_Y;\mathcal H_{k_\lambda}),\\
		\text{(iii)}:\quad\mathrm{MMD}^2\left(P_X,P_Y;\mathcal H_{k_\lambda}\right)&\geq \frac{3C_{10}(\alpha,\beta,d,\lambda)}{{n}}
	\end{align*}
	where $C_{10}(\alpha,\beta,d,\lambda):=C_9(\beta,d,\lambda)+C_1(\alpha,\beta,d,\lambda).$
	Observe that the inequalities (i) and (ii) are equivalent to the following inequalities (a) and (b), respectively:
	\begin{align*}
		\text{(a)}&:\quad R\geq 9C_7(\beta,d,\lambda,\Sigma)^2,\\
		\text{(b)}&:\quad\mathrm{MMD}^2\left(P_X,P_Y;\mathcal H_{k_\lambda}\right)\geq \frac{9C_8(\beta,d,\lambda)^2}{ {n}}.
	\end{align*}
	Considering the inequalities (a), (b) and (iii) above, for both tests $\Delta=\Delta_{{n_1},{n_2},R}^{\alpha,\lambda}$ or $\Delta=\Delta_{{n_1},{n_2},R}^{\alpha,u,\lambda}$, we have
	\begin{align*}
		\rho\big(\Delta,\,\beta,\,\mathcal C_{N,\Sigma},\,\delta_{\mathrm{MMD}} \big)^2\leq \frac{C_{11}(\alpha,\beta,d,\lambda)}{{n}},
	\end{align*}
	for some positive constant $C_{11}(\alpha,\beta,d,\lambda)>0$ and $R\geq 9C_7(\beta,d,\lambda,\Sigma)^2.$
	Therefore, we conclude that
	$$
	\rho\big(\Delta,\,\beta,\,\mathcal C_{N,\Sigma},\,\delta_{\mathrm{MMD}} \big)\leq C_{\mathrm{MMD}}(\alpha,\beta,d,\lambda){n}^{-1/2}
	$$
	with $C_{\mathrm{MMD}}(\alpha,\beta,d,\lambda):=C_{11}(\alpha,\beta,d,\lambda)^{1/2}$ and $R\geq 9C_7(\beta,d,\lambda,\Sigma)^2$.

		\subsection{Proof of Lemma~\ref{mmdcalculate}} \label{Section: Proof of lemma: mmdcalculate}
	Recall that the statistic $U_1$ in \Cref{U_1} is defined as 
	\begin{equation*}
		\begin{aligned}
			U_1=&\frac{1}{{n_1}({n_1}-1)} \sum_{1\leq i\neq j \leq {n_1}} \kappa(0)\cos\big({\omega^{\top}\left(X_i-X_j\right)}\big)-\frac{2}{{n_1} {n_2}} \sum_{i=1}^{n_1} \sum_{j=1}^{n_2} \kappa(0)\cos\big({\omega^{\top}\left(X_i-Y_j\right)}\big)\\
			&\quad+\frac{1}{{n_2}({n_2}-1)} \sum_{1\leq i\neq j \leq {n_2}} \kappa(0)\cos\big({\omega^{\top}\left(Y_i-Y_j\right)}\big).
		\end{aligned}
	\end{equation*}
	By taking conditional expectation with respect to $X$ and $Y$ given $\omega,$ we get
	\begin{align*}
		\mathbb E_{X\times Y}[U_1\,|\, \omega ]&=\mathbb E\big[\kappa(0)\cos\big(\omega^{\top}(X_1-X_2)\big)\,\big|\, \omega\big]\\
		&\quad+\mathbb E\big[\kappa(0)\cos\big(\omega^{\top}(Y_1-Y_2)\big)\,\big|\, \omega\big]\\
		&\quad-2\mathbb E\big[\kappa(0)\cos\big(\omega^{\top}(X_1-Y_1)\big)\,\big|\, \omega\big].
	\end{align*}
	Our strategy is to decompose the term $\big(\mathbb E_{X\times Y}[U_1\,|\, \omega ]\big)^2$ explicitly and simplify it with trigonometric identities. First, let $X_1,X_2,X_3,X_4$ be the independent copies of $X$, and $Y_1,Y_2,Y_3,Y_4$ be the independent copies of $Y$. And, let us drop the subscripts with respect to $X_1,\ldots,X_4,Y_1,\ldots,Y_4$ on $\mathbb E$ if the context is clear. Now, observe that 
	the term $\big(\mathbb E_{X\times Y}[U_1\,|\, \omega ]\big)^2$ can be expressed as
	\begin{align*}
		\big(\mathbb E_{X\times Y}[U_1\,|\, \omega ]\big)^2=\text{(i)}+\text{(ii)}+4\text{(iii)}+2\text{(iv)}-4\text{(v)}-4\text{(vi)},
	\end{align*}
	where
	\begin{align*}
		\text{(i)}&=\mathbb E\big[\kappa(0)^2\cos\big(\omega^{\top}(X_1-X_2)\big)\cos\big(\omega^{\top}(X_3-X_4)\big)\,\big|\, \omega\big],\\
		\text{(ii)}&=\mathbb E\big[\kappa(0)^2\cos\big(\omega^{\top}(Y_1-Y_2)\big)\cos\big(\omega^{\top}(Y_3-Y_4)\big)\,\big|\, \omega\big],\\
		\text{(iii)}&=\mathbb E\big[\kappa(0)^2\cos\big(\omega^{\top}(X_1-Y_1)\big)\cos\big(\omega^{\top}(X_2-Y_2)\big)\,\big|\, \omega\big],\\
		\text{(iv)}&=\mathbb E\big[\kappa(0)^2\cos\big(\omega^{\top}(X_1-X_2)\big)\cos\big(\omega^{\top}(Y_1-Y_2)\big)\,\big|\, \omega\big],\\
		\text{(v)}&=\mathbb E\big[\kappa(0)^2\cos\big(\omega^{\top}(X_1-X_2)\big)\cos\big(\omega^{\top}(X_3-Y_1)\big)\,\big|\, \omega\big],\\
		\text{(vi)}&=\mathbb E\big[\kappa(0)^2\cos\big(\omega^{\top}(Y_1-Y_2)\big)\cos\big(\omega^{\top}(Y_3-X_1)\big)\,\big|\, \omega\big].
	\end{align*}
	Here, observe that 
	\begin{align*}
		\text{(i)}&=\mathbb E\big[\kappa(0)^2\cos\big(\omega^{\top}(X_1-X_2)\big)\cos\big(\omega^{\top}(X_3-X_4)\big)\,\big|\, \omega\big]\\
		&=\mathbb E\bigg[\frac12\kappa(0)^2\Big(\cos\big(\omega^{\top}(X_1-X_2+X_3-X_4)\big)+\cos\big(\omega^{\top}(X_1-X_2+X_4-X_3)\big)\Big)\,\bigg|\, \omega\bigg]\\
		&=\mathbb E\big[\kappa(0)^2\cos\big(\omega^{\top}(X_1-X_2+X_3-X_4)\big)\,\big|\, \omega\big],
	\end{align*}
	where the last equality follows from $X_1-X_2+X_3-X_4\stackrel{d}{=}X_1-X_2+X_4-X_3$. A similar calculation guarantees that the term $\big(\mathbb E_{X\times Y}[U_1\,|\, \omega ]\big)^2$ can be written as
	\begin{align*}
		\big(\mathbb E_{X\times Y}[U_1\,|\, \omega ]\big)^2&=\text{(i)}+\text{(ii)}+4\text{(iii)}+2\text{(iv)}-4\text{(v)}-4\text{(vi)}\\
		&=(a)+(b)+4\bigg(\frac12(c)+\frac12(d)\bigg)+2(d)-4(e)-4(f)\\
		&=(a)+(b)+2(c)+4(d)-4(e)-4(f)
	\end{align*}
	where
	\begin{align*}
		(a)&=\mathbb E\big[\kappa(0)^2\cos\big(\omega^{\top}(X_1-X_2+X_3-X_4)\big)\,\big|\, \omega\big],\\
		(b)&=\mathbb E\big[\kappa(0)^2\cos\big(\omega^{\top}(Y_1-Y_2+Y_3-Y_4)\big)\,\big|\, \omega\big],\\
		(c)&=\mathbb E\big[\kappa(0)^2\cos\big(\omega^{\top}(X_1+X_2-Y_1-Y_2)\big)\,\big|\, \omega\big],\\
		(d)&=\mathbb E\big[\kappa(0)^2\cos\big(\omega^{\top}(X_1-X_2+Y_1-Y_2)\big)\,\big|\, \omega\big],\\
		(e)&=\mathbb E\big[\kappa(0)^2\cos\big(\omega^{\top}(X_1-X_2-X_3+Y_1)\big)\,\big|\, \omega\big],\\
		(f)&=\mathbb E\big[\kappa(0)^2\cos\big(\omega^{\top}(Y_1-Y_2-Y_3+X_1)\big)\,\big|\, \omega\big].
	\end{align*}
	Now, note that the symmetry of the cosine function allows different representations of the above terms. For example, combined with the symmetry of $X_1-X_2,$ ($e$) can also be written as 
	$$(e)=\mathbb E\big[\kappa(0)^2\cos\big(\omega^{\top}(X_1-X_2-X_3+Y_1)\big)\,\big|\, \omega\big]=\mathbb E\big[\kappa(0)^2\cos\big(\omega^{\top}(X_1-X_2+X_3-Y_1)\big)\,\big|\, \omega\big].$$
	Then, observe that
	\begin{align*}
		(a)+(d)-2(e)&=\mathbb E\big[\kappa(0)^2\cos\big(\omega^{\top}([X_1-X_2]-[X_3-X_4])\big)\,\big|\, \omega\big]\\
		&\quad+\mathbb E\big[\kappa(0)^2\cos\big(\omega^{\top}([Y_1-X_1]-[Y_2-X_2])\big)\,\big|\, \omega\big]\\
		&\quad-2\mathbb E\big[\kappa(0)^2\cos\big(\omega^{\top}([X_1-X_2]+[X_3-Y_1])\big)\,\big|\, \omega\big],
	\end{align*}
	and therefore,
	\begin{align*}
		\mathbb E_{\omega}\big[(a)+(d)-2(e)\big]=\kappa(0)\mathrm{MMD}^2(P_{X-X'},P_{X''-Y};\mathcal H_{k}),
	\end{align*}
	where $X',X''$ are independent copies of $X$. Similarly, we can show that
	\begin{align*}
		\mathbb E_{\omega}\big[(b)+(d)-2(f)\big]&=\kappa(0)\mathrm{MMD}^2(P_{Y-Y'},P_{X-Y''};\mathcal H_{k}),\\
		\mathbb E_{\omega}\big[(a)+(b)-2(c)\big]&=\kappa(0)\mathrm{MMD}^2(P_{X+X'},P_{Y+Y'};\mathcal H_{k}),
	\end{align*}
	where $Y',Y''$ are independent copies of $Y$.
	Since
	\begin{align*}
		\big(\mathbb E_{X\times Y}[U_1\,|\, \omega ]\big)^2&=(a)+(b)+2(c)+4(d)-4(e)-4(f)\\
		&=2\big((a)+(d)-2(e)\big)+2\big((b)+(d)-2(f)\big)\\
		&\quad-\big((a)+(b)-2(c)\big),
	\end{align*}
	we can conclude that
	\begin{align*}
		\mathbb E_{\omega}\big[\big(\mathbb E_{X\times Y}[U_1\,|\, \omega ]\big)^2\big]&=2\kappa(0)\mathrm{MMD}^2(P_{X-X'},P_{X''-Y};\mathcal H_{k})+2\kappa(0)\mathrm{MMD}^2(P_{Y-Y'},P_{X-Y''};\mathcal H_{k})\\
		&\quad-\kappa(0)\mathrm{MMD}^2(P_{X+X'},P_{Y+Y'};\mathcal H_{k}).
	\end{align*}
	Additionally, note that the second statement in the lemma can be proven in a similar manner, thereby completing the proof.

	\subsection{Proof of Lemma~\ref{lemma: moment bound for Gaussian}}  \label{Section: Proof of lemma: moment bound for Gaussian}
	 Recall the class of Gaussian distributions with a common fixed covariance matrix $\Sigma \in \mathbb R^{d\times d}$:
	\begin{align*}
		\mathcal C_{N,\Sigma}:=\big\{(P_X,P_Y)\in \mathcal{P}_{\mathrm{conti}} \,\big|\, P_X = N(\mu_X,\Sigma),~ P_Y = N(\mu_Y,\Sigma) ~ \text{where $\mu_X, \mu_Y\in \mathbb{R}^d$}  \big\}.
	\end{align*}
	Here we claim that the following inequality
	\begin{equation}\label{mmdnormalobj}
		\begin{aligned}
			\mathbb E_{\omega}\big[\big(\mathbb E_{X\times Y}[U_1\,|\, \omega ]\big)^2\big] \leq C \big( \mathrm{MMD}^2(P_X,P_Y;\mathcal H_{k_\lambda})\big)^c
		\end{aligned}
	\end{equation}
	holds for any distribution pair $(P_X,P_Y) \in\mathcal  C_{N,\Sigma}$, with $c=2$ and some positive constant $C>0.$
	To prove the claim, one important observation is that the exact calculation of $\mathrm{MMD}^2(P_{X},P_{Y};\mathcal H_{k})$ is feasible when we use the Gaussian kernel. To be specific, consider a Gaussian kernel with bandwidth $\lambda=(\lambda_1,\ldots,\lambda_d)^\top \in (0,\infty)^d$,
	$$
	k_\lambda(x,y)=\kappa_\lambda(x-y)=\prod^{d}_{i=1}\frac{1}{\sqrt{\pi}\lambda_i}e^{-\frac{(x_i-y_i)^2}{\lambda_i^2}}.
	$$
	There have been several existing results on calculating MMD with a Gaussian kernel for Gaussian distributions. Among them, we leverage the result from \citet[Proposition 1]{ramdas2015decreasing}, which is displayed in Lemma \ref{gaussian MMD lemma}:
	\begin{equation*}
		\begin{aligned}
			\mathrm{MMD}^2(P_X,P_Y;\mathcal H_{k_\lambda}) &= 2\left(\frac{1}{4\pi}\right)^{d / 2} \frac{1-\exp \big\{-(\mu_X-\mu_Y)^{\top}\left(\Sigma+D(\lambda^2/4)\right)^{-1} (\mu_X-\mu_Y) / 4\big\}}{\left|\Sigma+D(\lambda^2/4)\right|^{1 / 2}}\\
			&= C_1(d,\lambda,\Sigma)\big(1-\exp \big\{-(\mu_X-\mu_Y)^{\top}\left(\Sigma+D(\lambda^2/4)\right)^{-1} (\mu_X-\mu_Y) / 4\big\}\big),
		\end{aligned}
	\end{equation*}
	for a constant $C_1(d,\lambda,\Sigma)=2\left(\frac{1}{4\pi}\right)^{d / 2}\left|\Sigma+D(\lambda^2/4)\right|^{-1 / 2}$ and $D(\lambda^2/4)=\mathrm{diag}(\lambda_1^2/4,\ldots,\lambda_d^2/4).$
	We are now ready to analyze the two terms in \Cref{mmdnormalobj}.
	
	\subsubsection*{Exact value of $\mathbb E_{\omega}\big[\big(\mathbb E_{X\times Y}[U_1\,|\, \omega ]\big)^2\big]$}
	Recall Lemma \ref{mmdcalculate} and observe that $\mathbb E_{\omega}\big[\big(\mathbb E_{X\times Y}[U_1\,|\, \omega ]\big)^2\big]$ can be expressed with several $\mathrm{MMD}^2$ terms:
	\begin{align*}
		\mathbb E_{\omega}\big[\big(\mathbb E_{X\times Y}[U_1\,|\, \omega ]\big)^2\big]&=2\kappa_\lambda(0)\mathrm{MMD}^2(P_{X-X'},P_{X''-Y};\mathcal H_{k_\lambda})+2\kappa_\lambda(0)\mathrm{MMD}^2(P_{Y-Y'},P_{X-Y''};\mathcal H_{k_\lambda})\\
		&\quad-\kappa_\lambda(0)\mathrm{MMD}^2(P_{X+X'},P_{Y+Y'};\mathcal H_{k_\lambda}).
	\end{align*}
	To simplify the above equation, let us define Gaussian random variables $Z_1,Z_2,Z_3,Z_4$ such that 
	$$
	Z_1\sim N(0,2\Sigma),\quad Z_2\sim N(\mu_X-\mu_Y,2\Sigma),\quad Z_3\sim N(2\mu_X,2\Sigma),\quad Z_4\sim N(2\mu_Y,2\Sigma).
	$$
	Then, observe that $X-X',~Y-Y'\stackrel d= Z_1,~X-Y\stackrel d= Z_2,~X+X'\stackrel d= Z_3$ and $Y+Y'\stackrel d= Z_4$, and these equivalences in distribution yield
	\begin{align*}
		\mathbb E_{\omega}\big[\big(\mathbb E_{X\times Y}[U_1\,|\, \omega ]\big)^2\big]&=4\kappa_\lambda(0)\mathrm{MMD}^2(P_{Z_1},P_{Z_{2}};\mathcal H_{k_\lambda})-\kappa_\lambda(0)\mathrm{MMD}^2(P_{Z_3},P_{Z_4};\mathcal H_{k_\lambda}).
	\end{align*}
	We apply the $\mathrm{MMD}$ calculation formula in Lemma \ref{gaussian MMD lemma} here and obtain
	\begin{align*}
		&\quad\mathbb E_{\omega}\big[\big(\mathbb E_{X\times Y}[U_1\,|\, \omega ]\big)^2\big]\\ &=4\kappa_\lambda(0)C_1(d,\lambda,2\Sigma)\big(1-\exp \big\{-(\mu_X-\mu_Y)^{\top}\left(2\Sigma+D(\lambda^2/4)\right)^{-1} (\mu_X-\mu_Y) / 4\big\}\big)\\
		&\quad-\kappa_\lambda(0)C_1(d,\lambda,2\Sigma)\big(1-\exp \big\{-(\mu_X-\mu_Y)^{\top}\left(2\Sigma+D(\lambda^2/4)\right)^{-1} (\mu_X-\mu_Y) \big\}\big)\\
		&=C_2(d,\lambda,\Sigma)\bigl(3-4\exp(-s_a)+\exp(-4s_a)\bigr),
	\end{align*}
	where we denote $s_a=(\mu_X-\mu_Y)^{\top}\left(2\Sigma+D(\lambda^2/4)\right)^{-1} (\mu_X-\mu_Y) / 4$, and $C_2(d,\lambda,\Sigma)=\kappa_\lambda(0)C_1(d,\lambda,2\Sigma).$
	\subsubsection*{Exact value of $\big(\mathrm{MMD}^2(P_X,P_Y;\mathcal H_{k_\lambda})\big)^2$}
	By squaring the formula in Lemma \ref{gaussian MMD lemma}, we have
	\begin{equation*}
		\begin{aligned}
			&\big(\mathrm{MMD}^2(P_{X},P_{Y};\mathcal H_{k_\lambda})\big)^2\\=& C_1(d,\lambda,\Sigma)^2\big(1-\exp \big\{-(\mu_X-\mu_Y)^{\top}\left(\Sigma+D(\lambda^2/4)\right)^{-1} (\mu_X-\mu_Y) / 4\big\}\big)^2\\
			=&C_3(d,\lambda,\Sigma)\bigl(1-2\exp(-s_b)+\exp(-2s_b)\bigr),
		\end{aligned}
	\end{equation*}
	where $s_b=(\mu_X-\mu_Y)^{\top}\left(\Sigma+D(\lambda^2/4)\right)^{-1} (\mu_X-\mu_Y) / 4,$ and $C_3(d,\lambda,\Sigma)=C_1(d,\lambda,\Sigma)^2.$
	\subsubsection*{Existence of constant $C$}
	
	Our goal now is to show the existence of $C$ such that
\begin{align*}
    \mathbb E_{\omega}\big[\big(\mathbb E_{X\times Y}[U_1\,|\, \omega ]\big)^2\big]
    \leq C \big(\mathrm{MMD}^2(P_X,P_Y;\mathcal H_{k_\lambda})\big)^2.
\end{align*}
If $\mu_X=\mu_Y$, then $\mathrm{MMD}^2(P_X,P_Y;\mathcal H_{k_\lambda})=0$ and $
    \mathbb E_{\omega}\big[\big(\mathbb E_{X\times Y}[U_1\,|\, \omega ]\big)^2\big]=0$, so the desired inequality is trivial. Hence, in the rest of the argument, assume that $\mu_X\neq\mu_Y$. Then, we aim to show that there exists some positive constant $C$ such that
\begin{align*}
	\frac{\mathbb E_{\omega}\big[\big(\mathbb E_{X\times Y}[U_1\,|\, \omega ]\big)^2\big]}{\big(\mathrm{MMD}^2(P_X,P_Y;\mathcal H_{k_\lambda})\big)^2}\leq C.
\end{align*}
	Plugging our previous results in the above equation, it is equivalent to
	$$
	\frac{C_2(d,
		\lambda,\Sigma)}{C_3(d,
		\lambda,\Sigma)}\frac{3-4\exp(-s_a)+\exp(-4s_a)}{1-2\exp(-s_b)+\exp(-2s_b)}\leq C.
	$$
	Note that the last term can be written as
	\begin{align*}
		\frac{3-4\exp(-s_a)+\exp(-4s_a)}{1-2\exp(-s_b)+\exp(-2s_b)}=\underbrace{\big(3+2\exp(-s_a)+\exp(-2s_a)\big)}_{:=f(s_a)} \underbrace{\frac{1-2\exp(-s_a)+\exp(-2s_a)}{1-2\exp(-s_b)+\exp(-2s_b)}}_{:=\frac{g(s_a)}{g(s_b)}}.
	\end{align*}
	Note that $\mu_X\neq\mu_Y$, thus we have $s_a, s_b>0$. Also, since $2\Sigma+D(\lambda^2/4)\succeq \Sigma+D(\lambda^2/4)$, we have
\begin{align*}
    0< s_a\leq s_b.
\end{align*}
	Since $f(0)=6$ and $f'(x)=-2\exp(-x)\big(1+\exp(-x)\big)\leq0$
	for $\forall x \in \mathbb R,$ we have $f(s_a)< 6$ for $s_a >0.$
	Also, observe that $g'(x)=2\exp(-x)\big(1-\exp(-x)\big)\geq 0$ for $\forall x\geq 0$, $g(0)=0$ and $s_a\leq s_b$ for all $(\mu_X-\mu_Y) \in \mathbb R^d,$ thus we get $0< g(s_a)/g(s_b)\leq 1.$ Therefore, we can derive 
	$$
	f(s_a)\frac{g(s_a)}{g(s_b)}\leq 6,
	$$
	and this implies that there exists some positive constant $C(d,\lambda,\Sigma)>0$ satisfying
	$$
	\frac{C_2(d,
		\lambda,\Sigma)}{C_3(d,
		\lambda,\Sigma)}f(s_a)\frac{g(s_a)}{g(s_b)}\leq C(d,\lambda,\Sigma).
	$$
	This completes the proof of Lemma~\ref{lemma: moment bound for Gaussian}.

\end{document}